%% file: main.tex
\pdfoutput=1

\documentclass[10pt,twocolumn,letterpaper]{article}

\usepackage[pagenumbers]{cvpr}

\input{preamble}

\definecolor{cvprblue}{rgb}{0.21,0.49,0.74}
\usepackage[pagebackref,breaklinks,colorlinks,allcolors=cvprblue]{hyperref}

\usepackage{placeins}
\renewcommand{\topfraction}{0.9}\renewcommand{\dbltopfraction}{0.9}\renewcommand{\textfraction}{0.07}
\title{Every Edit Its Own Budget: Per-Instruction Token Freezing for\\Training-Free Acceleration of Multi-Edit Image Editing}

\author{%
Jyotirmoy Nath\textsuperscript{1}\hspace{2.2em}Neeraj Kumar\textsuperscript{2}\hspace{2.2em}Brejesh Lall\textsuperscript{1}\\[0.55em]
{\normalsize\textsuperscript{1}Indian Institute of Technology Delhi, India\hspace{2.6em}\textsuperscript{2}Adobe Inc., India}\\[0.3em]
{\small\ttfamily \{jyotirmoy.nath, brejesh\}@ee.iitd.ac.in,\hspace{0.6em}neerajkr2k14@gmail.com}\\[0.3em]
}
\begin{document}
\maketitle
\input{sec/0_abstract}

\input{sec/1_intro}
\input{sec/2_related}
\input{sec/3_method}
\input{sec/4_theory}
\input{sec/5_experiments}
\input{sec/6_conclusion}
\input{sec/7_llm_statement}
{
    \small
    \bibliographystyle{ieeenat_fullname}
    \bibliography{main}
}

\input{sec/X_suppl}

\end{document}

%% file: preamble.tex
\usepackage{array}

\renewcommand{\topfraction}{0.9}
\renewcommand{\bottomfraction}{0.8}
\renewcommand{\textfraction}{0.07}
\renewcommand{\floatpagefraction}{0.85}
\renewcommand{\dbltopfraction}{0.9}
\renewcommand{\dblfloatpagefraction}{0.85}
\usepackage{iftex}
\ifXeTeX
  \usepackage{fontspec}
\fi

\usepackage{cuted}      %
\usepackage{currfile}   %
\usepackage{multirow}
\usepackage{amsthm}
\usepackage{bm}
\usepackage{algorithm}
\usepackage{algpseudocode}
\usepackage{tikz}
\usetikzlibrary{arrows.meta,positioning,fit,calc,backgrounds}
\usepackage{pifont}   %
\usepackage{pgfplots}
\pgfplotsset{compat=1.17}

\newcommand{\method}{TIDE\xspace}
\DeclareMathOperator*{\argmin}{arg\,min}

\theoremstyle{plain}
\newtheorem{theorem}{Theorem}
\newtheorem{proposition}{Proposition}

\theoremstyle{definition}

\theoremstyle{remark}

\newcommand{\tabsize}{\footnotesize}

%% file: sec/0_abstract.tex
\begin{strip}
\vspace*{-45pt}
\centering
\definecolor{tzhead}{gray}{0.91}\definecolor{tzband}{rgb}{0.93,0.94,0.98}
\definecolor{tzdrop}{rgb}{0.86,0.12,0.12}\definecolor{tzkeep}{rgb}{0.18,0.66,0.22}
\newlength{\tw}\setlength{\tw}{0.1655\textwidth}\newlength{\tg}\setlength{\tg}{2pt}
\newcommand{\thd}[2]{\tikz\node[fill=#1,rounded corners=3pt,minimum width=\tw,minimum height=4.6mm,inner xsep=0pt,inner ysep=1pt,font=\footnotesize]{#2};}
\newcommand{\timg}[2]{\begin{tikzpicture}\begin{scope}\clip[rounded corners=3pt](0,0)rectangle(\tw,\tw);
\node[inner sep=0,anchor=south west] at (0,0) {\includegraphics[width=\tw]{#1}};\end{scope}
\foreach \xa/\ya/\xb/\yb/\c in {#2} {\draw[\c,line width=1.4pt,rounded corners=1.5pt] (\xa\tw,\ya\tw) rectangle (\xb\tw,\yb\tw);}
\end{tikzpicture}}
\newcommand{\timgT}[2]{\begin{tikzpicture}\begin{scope}\clip[rounded corners=3pt](0,0)rectangle(\tw,\tw);
\node[inner sep=0,anchor=south west] at (0,0) {\includegraphics[width=\tw]{#1}};\end{scope}
\foreach \xa/\ya/\xb/\yb/\c in {#2} {\draw[\c,line width=1.4pt,rounded corners=1.5pt] (\xa\tw,\ya\tw) rectangle (\xb\tw,\yb\tw);}
\draw[tzkeep!75,line width=0.8pt,rounded corners=3pt] (0.4pt,0.4pt) rectangle (\tw-0.4pt,\tw-0.4pt);
\end{tikzpicture}}
\newcommand{\tband}[2]{\tikz\node[fill=tzband,rounded corners=3pt,text width=\dimexpr5\tw+4\tg-6pt\relax,inner sep=3pt,align=justify,font=\scriptsize]
  {\textbf{\color{cvprblue}#1}\ \textit{#2}};}
\newsavebox{\tbA}\newsavebox{\tbB}\begin{lrbox}{\tbA}\begin{minipage}[c]{\dimexpr5\tw+4\tg\relax}
\setlength{\tabcolsep}{0pt}\renewcommand{\arraystretch}{0}
\begin{tabular}{@{}c@{\hspace{\tg}}c@{\hspace{\tg}}c@{\hspace{\tg}}c@{\hspace{\tg}}c@{}}
\thd{tzhead}{\textbf{Input}} & \thd{tzhead}{Vanilla (1.00$\times$)} & \thd{tzhead}{Global Threshold\,(2.09$\times$)} &
\thd{cvprblue!22}{\textbf{\color{cvprblue}\method (Ours, 2.20$\times$)}} & \thd{tzhead}{\textbf{\method Regions}} \\[2pt]
\timg{}{} & \timg{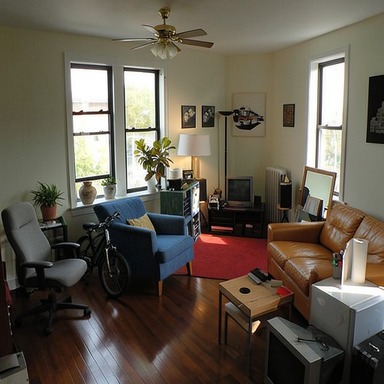}{} &
\timg{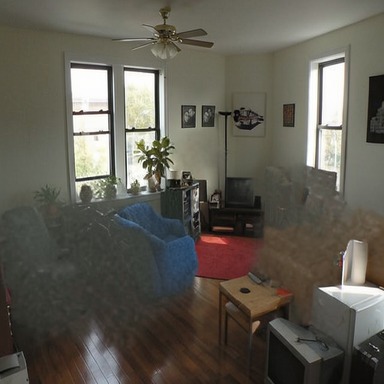}{0.01/0.15/0.20/0.48/tzdrop, 0.64/0.15/0.95/0.41/tzdrop} &
\timgT{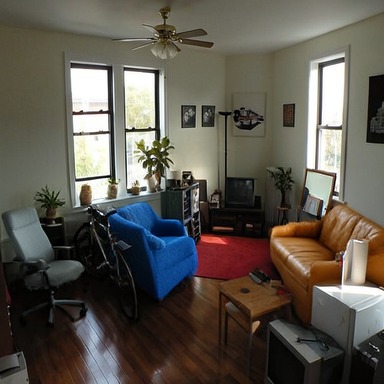}{0.01/0.15/0.20/0.48/tzkeep, 0.64/0.15/0.95/0.41/tzkeep} &
\timg{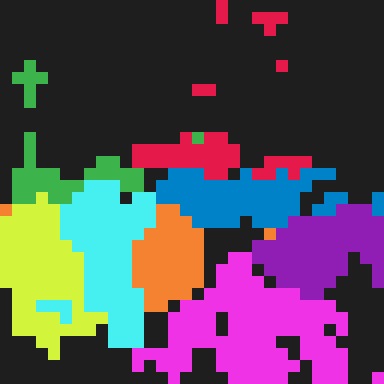}{} \\[2pt]
\multicolumn{5}{@{}c@{}}{\tband{(a) Living room }{Replace the red chair on the left with a gray office chair, the bicycle with a motorcycle, the green couch with a blue fabric armchair,
the dark couch on the right with a tan leather loveseat, the small plant on the window sill with a ceramic vase, the larger plant with a large lamp,
the TV on the stand with a vintage radio, and the TV on the floor with a low wooden coffee table.}} \\[3pt]
\timg{}{} & \timg{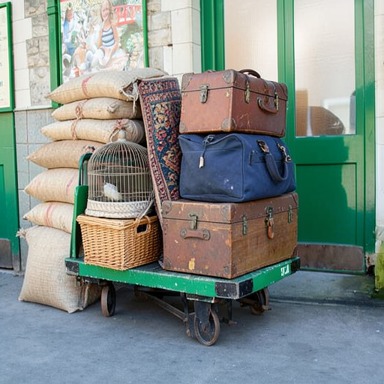}{} &
\timg{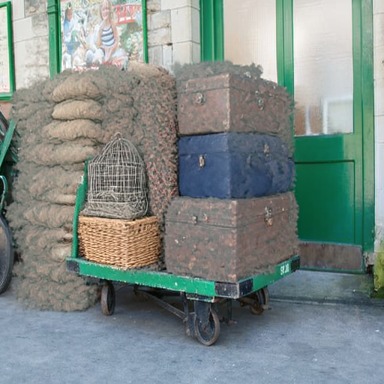}{0.29/0.36/0.47/0.83/tzdrop, 0.47/0.60/0.80/0.86/tzdrop} &
\timgT{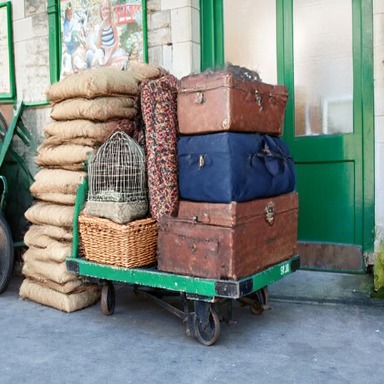}{0.29/0.36/0.47/0.83/tzkeep, 0.47/0.60/0.80/0.86/tzkeep} &
\timg{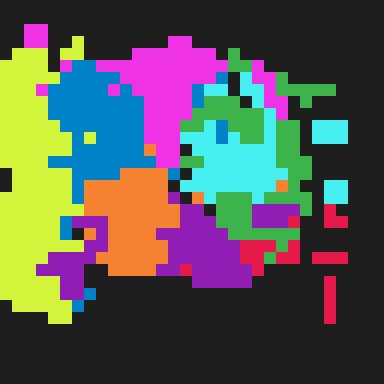}{} \\[2pt]
\multicolumn{5}{@{}c@{}}{\tband{(b) Luggage cart.}{``On the luggage cart, replace the tall vertical suitcase on the far left with a stack of heavy grain sacks, the vertical
one behind the left stack with a rolled-up Persian rug, and then swap the left stack from top to bottom with a cloth-covered bird cage, a wicker picnic basket, and a
rusty iron strongbox, while replacing the right stack from top to bottom with a vintage leather trunk, a navy blue canvas duffel bag, and a weathered wooden crate.''}}
\end{tabular}
\end{minipage}\end{lrbox}\begin{lrbox}{\tbB}\begin{tikzpicture}[font=\scriptsize,baseline=(current bounding box.center)]
\node[draw=black!22,rounded corners=4pt,fill=white,inner sep=4pt,text width=0.125\textwidth,align=left] {\raggedright
\textbf{Key}\\[3pt]
\tikz\draw[tzdrop,line width=1.4pt,rounded corners=1.5pt](0,0)rectangle(10pt,7pt);\ Edit dropped\\[0pt]
\hspace*{14pt}{\color{black!60}global threshold}\\[4pt]
\tikz\draw[tzkeep,line width=1.4pt,rounded corners=1.5pt](0,0)rectangle(10pt,7pt);\ Edit kept\\[0pt]
\hspace*{14pt}{\color{black!60}\method}\\[7pt]
\textbf{Regions}\\[2pt]
\foreach \cr/\cg/\cb in {230/25/75,60/180/75,0/130/200,245/130/48,145/30/180,70/240/240}
  {\tikz\fill[rounded corners=1pt,fill={rgb,255:red,\cr;green,\cg;blue,\cb}](0,0)rectangle(7pt,7pt);\hspace{0.6pt}}%
\\[1pt]
{\color{black!60}one color per part}\\[3pt]
\tikz\fill[rounded corners=1pt,fill={rgb,255:red,30;green,30;blue,30}](0,0)rectangle(7pt,7pt);\ background\\[4pt]
{\color{black!60}from the editor's own\\ attention, no masks}};
\end{tikzpicture}\end{lrbox}\par\noindent\usebox{\tbA}\hfill\usebox{\tbB}
\captionsetup{hypcap=false}

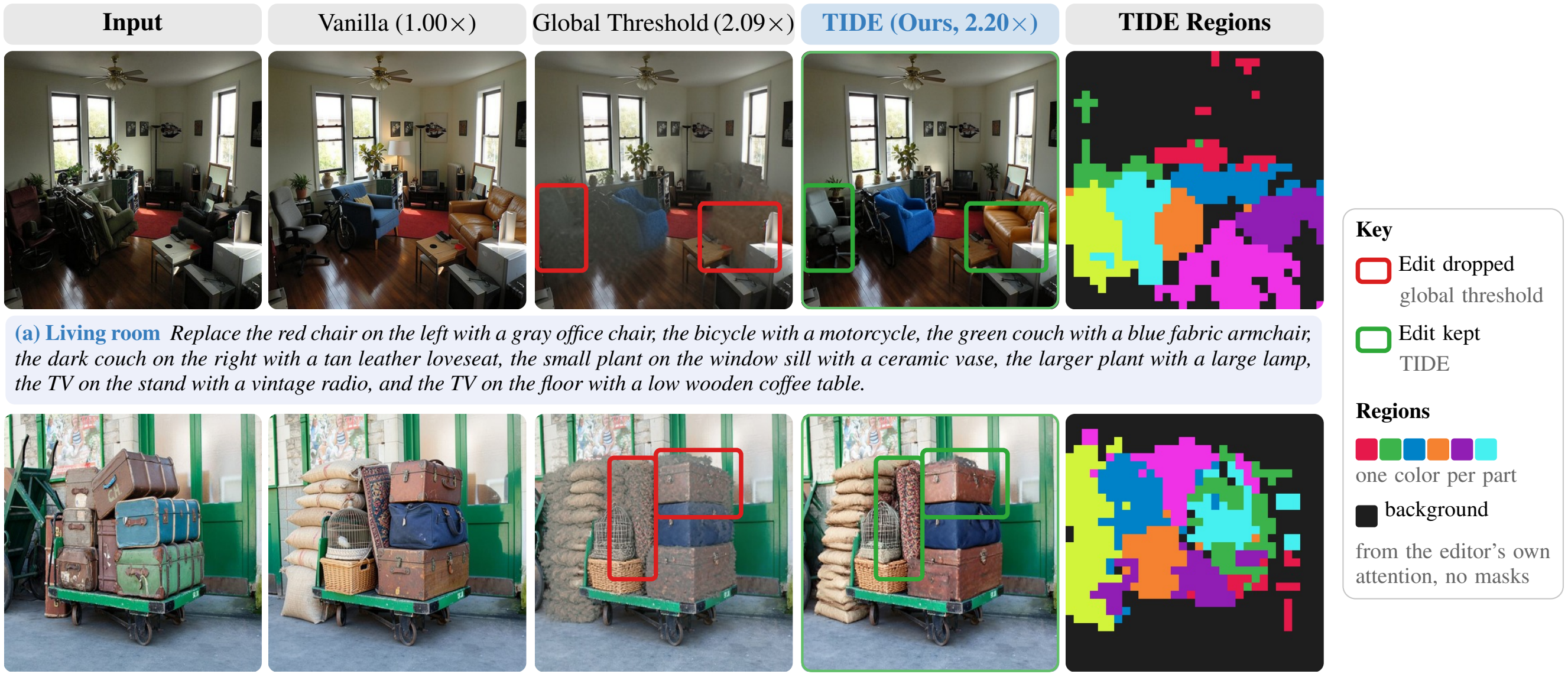
\captionof{figure}{\textbf{One global threshold drops edits; per-instruction budgets keep them.} A single skipping threshold freezes some edits before they are done (\textcolor{tzdrop}{red boxes}). \method finds each instruction part's region in the editor's own attention (right, no masks), freezes tokens relative to their own part, and keeps every edit at a higher speed (\textcolor{tzkeep}{green boxes}). FLUX.2 Klein 4B, 4 steps, MICE-Bench.}
\captionsetup{hypcap=true}
\label{fig:teaser}
\end{strip}

\begin{abstract}
Accelerating instruction-based image editing when one instruction requests several edits is challenging. Many training-free accelerators for flow-matching image editors skip tokens whose prediction has stopped changing, but decide this with a single threshold for the whole image. With several edits, such a threshold, tuned for speed, freezes small and slowly settling edits first, and requested edits disappear. We propose \method (Training-free Instruction-aware Drift-based Editing), which gives every instruction part its own compute budget. First, \method finds the region of each part in the editor's own attention by unbalanced optimal transport, without masks or external models. Second, it freezes a token once its drift is small relative to its own part and moves it along the straight path to its endpoint estimate. Third, a sparse engine computes only the remaining tokens. We prove that part-agnostic rules cannot guarantee per-part error budgets, characterize the optimal per-part threshold policy up to one tie class per part, and derive when our rule attains it. On FLUX.2 Klein, one global threshold breaks at least one edit's budget in 89\% of MICE-Bench images, and \method drops up to 77\% fewer edits at matched or higher speed (48\% on MICE-Bench), a gap human raters confirm. On the many-step Klein Base, \method is the fastest compared method on six benchmarks (4.81--4.89$\times$) and needs 5.3$\times$ fewer TFLOPs than vanilla, the largest saving we measure. With the same settings, it is also the fastest compared method on Qwen-Image-Edit-2511, a second model family (3.46--3.52$\times$), and drops the fewest edits on all six benchmarks.
Code will be made publicly available.
\end{abstract}

%% file: sec/1_intro.tex
\vspace{-10pt}
\section{Introduction}
\label{sec:intro}
Instruction-based image editing changes a source image to follow a text
instruction while leaving the rest intact~\cite{instructpix2pix,magicbrush}.
Flow-matching transformers~\cite{rectflow,flowmatch,sd3,dit} have made this
practical: in-context editors such as FLUX.1 Kontext~\cite{kontext},
FLUX.2~\cite{flux2}, and Qwen-Image-Edit~\cite{qwenimage} read the source image
and the instruction in a single joint attention and edit in 4 to 50 sampling
steps. Increasingly, one instruction asks for several changes at once:
MICE-Bench~\cite{mice} averages 8.0 edits per image, LoMOE-Bench~\cite{lomoe}
3.0, and MIRA-Bench~\cite{mirage} 5.0. Our goal is to accelerate these
multi-edit instructions without dropping requested edits and without retraining.

Training-free accelerators exploit the fact that most of an edited image
converges early, so its computation need not be repeated at every step. They
reuse or forecast features across whole steps~\cite{teacache,taylorseer,fora,deltadit},
skip tokens whose prediction has stopped changing~\cite{toca,ras,regione,spotedit},
or restrict computation to a given edit region~\cite{eedit,kvedit}. All of them
decide what has converged with \emph{one} threshold for the whole image. This is
adequate for a single edit, but a multi-edit instruction is several edits of
different size and speed: replacing a sofa moves many tokens far, while
recoloring a small cup moves a few tokens slightly. A single threshold tuned for
speed freezes the subtlest edit first, and that edit simply vanishes
(\cref{fig:teaser}). Giving each edit its own budget requires knowing where it lives. Multi-instance
editors obtain this from user masks~\cite{lomoe,mice} or an external
VLM~\cite{mirage}, but the editor already holds the answer: its own attention
ties every image token to the words of the instruction, and its own flow
predicts which tokens will move, so the information needed to budget each edit is
available at almost no extra cost. It cannot, however, be recovered from drift
alone---because a token's drift does not reveal which edit it serves, no rule
that sees only a token's own drift can separate a slowly settling subtle edit
from settled background (\cref{thm:agnostic}). The decision needs the instruction
part. Our contributions are:

\begin{itemize}
    \item  We cast training-free acceleration of multi-edit
    instructions as per-part budget allocation, and prove that per-part
    thresholds are optimal up to one tie class per part, that their gain over a
    single global threshold is exactly computable, and that no rule blind to the
    parts can guarantee heterogeneous budgets; we also give the condition under
    which our rule attains this optimum.

    \item We propose \method, which localizes every instruction
    part in the editor's own attention by unbalanced optimal transport (no masks,
    segmenter, or VLM), freezes a token once it has converged relative to its own
    part, and computes only the remaining tokens. We prove that transport keeps
    weakly attended parts that argmax drops and softmax shrinks.

    \item On real trajectories, one global threshold breaks at
    least one edit's budget in 89\% of MICE-Bench images. Across three editors
    from two families and six benchmarks, evaluated against seven training-free
    accelerators and multi-instance editors, \method drops up to 77\% fewer edits than a global threshold at matched or higher speed (48\% on MICE-Bench).
    On the many-step
    Klein Base it is the fastest method (4.81--4.89$\times$) with the largest
    compute saving (5.3$\times$ fewer TFLOPs) and drops fewer edits than the fastest baseline on five of six benchmarks; under the same settings it is also
    fastest on Qwen-Image-Edit-2511 (3.46--3.52$\times$).
\end{itemize}

%% file: sec/2_related.tex
\newcommand{\runin}[1]{\smallskip\noindent\textbf{#1}:}

\section{Prior Work}
\label{sec:related}

\runin{Multi-edit image editing} Instruction editors evolved from training-based models~\cite{instructpix2pix,magicbrush,emuedit} to
in-context flow transformers in which every image token attends to the source image and to the words of the
instruction~\cite{kontext,flux2,qwenimage,step1x}. With several edits, attribute leakage and missed edits grow~\cite{mice}. LoMOE~\cite{lomoe}
and MICE~\cite{mice} control each edit with a user mask, MIRAGE~\cite{mirage} grounds each part with an external VLM, and
ParallelEdits~\cite{paralleledits} groups attention by edit aspect. These methods improve quality, not speed, and most need masks or an
external model; \method needs neither.

\runin{Attention grounding and evaluation} Cross-attention maps localize words~\cite{p2p,daam,attendexcite}; argmax or thresholding turns them
into regions but lets salient words absorb the tokens of subtle ones. Unbalanced optimal transport~\cite{sinkhorn,peyre,uot} assigns mass under
marginal constraints, which keeps weakly attended parts. Multimodal judges check each requested element~\cite{tifa,dsg,viescore}, as our
per-part judge does (\cref{sec:setup}).

\runin{Temporal acceleration} A first family reuses computation across steps: block caches at fixed
intervals~\cite{fora,deltadit,deepcache,l2c}, step skipping when the timestep-modulated input changes little (TeaCache~\cite{teacache}) and
Taylor forecasts of features (TaylorSeer~\cite{taylorseer}). Each decision covers the whole image.

\runin{Token-level acceleration} A second family skips individual tokens: ToMeSD~\cite{tomesd} merges them, ToCa~\cite{toca} caches them by a
per-token score and RAS~\cite{ras} recomputes only the regions in focus. For editing, EEdit~\cite{eedit} and KV-Edit~\cite{kvedit} skip or
cache computation outside a given region or mask, SpotEdit~\cite{spotedit} and SpecEdit~\cite{specedit} recompute only tokens judged changed,
RegionCache~\cite{regioncache} reuses regions across turns, and FLUX.2 [klein] 9B-KV~\cite{flux2} caches the reference keys and values.
Closest to our work, RegionE~\cite{regione} lets the tokens of unedited regions finish early. All these accelerators decide where to skip
with one criterion for the whole image; none gives each instruction part its own budget, which protects a subtle edit next to a large one
(\cref{sec:theory,sec:claim}; positioning in the supplement).

%% file: sec/3_method.tex
\vspace{-7pt}
\section{Proposed Method}

\label{sec:method}
\begin{figure*}[t]
\centering
\resizebox{\textwidth}{!}{%
\begin{tikzpicture}[
  font=\sffamily\scriptsize, >={Stealth[length=4pt]},
  img/.style={inner sep=0pt, outer sep=0pt},
  blk/.style={draw=#1!60!black, fill=#1!12, rounded corners=3pt, align=center, inner sep=3pt, line width=0.5pt},
  blk/.default=gray,
  sub/.style={draw=#1!55!black, fill=#1!18, rounded corners=2pt, align=center, inner sep=2pt, font=\sffamily\tiny, minimum width=1.3cm},
  note/.style={draw=#1!60!black, fill=#1!8, rounded corners=2pt, align=center, inner sep=2.5pt, font=\sffamily\tiny},
  arr/.style={->, thick, black!65},
  lbl/.style={font=\sffamily\tiny, align=center, text=black!75},
  tok/.style={draw=black!50, minimum size=0.27cm, inner sep=0pt, font=\sffamily\tiny\bfseries},
  ]
\def\tpanel#1#2#3#4#5#6{%
  \fill[#5!4] [rounded corners=5pt] (#1,#2) rectangle (#3,#4);
  \begin{scope}\clip[rounded corners=5pt] (#1,#2) rectangle (#3,#4);
    \fill[#5!22] (#1,#2) rectangle (#3,#2-0.48);\end{scope}
  \draw[#5!55!black!70, line width=0.6pt, rounded corners=5pt] (#1,#2) rectangle (#3,#4);
  \node[anchor=west, font=\sffamily\small\bfseries, text=black!85] at (#1+0.15,#2-0.24) {#6};}
\tpanel{0}{0}{5.5}{-4.6}{blue}{\ding{172} Dense steps (few)}
\tpanel{5.7}{0}{15.65}{-4.6}{green}{\ding{173} Per-instruction freezing with wake-up (most steps)}
\tpanel{15.85}{0}{19.4}{-4.6}{blue}{\ding{174} Decode}
\node[img] (I) at (0.95,-1.55) {\includegraphics[width=1.3cm]{}};
\node[lbl, above=1pt of I] {source image $I$};
\node[blk=violet, font=\sffamily\tiny, inner sep=2.5pt, align=center] (P) at (0.95,-3.42) {\textbf{Instruction}\\(6 parts)\\[1pt]white mug,\\navy mug,\\{.\kern-0.03em.\kern-0.03em.}\\pink cupcake};
\node[img] (X0) at (2.75,-1.05) {\includegraphics[width=0.75cm]{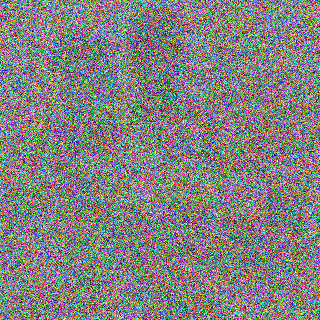}};
\node[lbl, left=1pt of X0] {$x_t$};
\node[blk=blue, minimum width=1.05cm, minimum height=1.45cm, font=\sffamily\small\bfseries] (D1) at (2.75,-2.6) {DiT\\[-1pt]{\tiny\mdseries all tokens}};
\draw[arr] (X0.south) -- (D1.north);
\draw[arr] (I.east) -- ++(0.2,0) |- ([yshift=0.35cm]D1.west);
\draw[arr] (P.east) -- ++(0.2,0) |- ([yshift=-0.35cm]D1.west);
\node[lbl] at (2.75,-3.6) {dense steps\\$0..s{-}1$};
\node[blk=gray, minimum width=1.55cm, minimum height=2.75cm, fill=white] (Rec) at (4.55,-2.55) {};
\node[font=\sffamily\tiny\bfseries] at (4.55,-1.38) {record};
\node[sub=orange] (r1) at (4.55,-1.85) {word attention\\$A_{ki}$};
\node[sub=orange] (r2) at (4.55,-2.62) {drift $\delta_i$};
\node[sub=orange] (r3) at (4.55,-3.3) {text/reference\\K,V};
\draw[arr] (D1.east) -- (Rec.west |- D1.east);
\node[blk=green, text width=1.5cm, font=\sffamily\scriptsize\bfseries] (R) at (6.85,-1.55) {Self-grounded regions\\[-1pt]{\tiny\mdseries unbalanced OT (a)}};
\draw[arr] (Rec.east |- R.west) -- (R.west);
\node[img] (Rimg) at (6.85,-3.1) {\includegraphics[width=1.05cm]{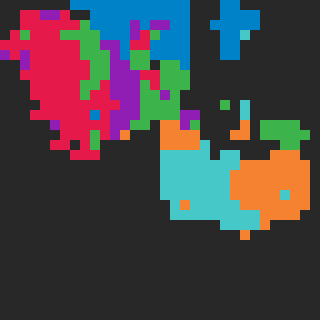}};
\node[lbl, below=1pt of Rimg] {part $k(i)$ per token};
\draw[arr] (R) -- (Rimg);
\node[blk=red, text width=1.85cm] (F) at (9.05,-1.55) {\textbf{Freeze test}\\[2pt]$\delta_{i}<\tau\, m_{k(i)}$\\[1pt]{\tiny one $\tau$, scale of own part}};
\draw[arr] (R.east) -- (F.west);
\node[img] (Fimg) at (9.05,-3.1) {\includegraphics[width=1.05cm]{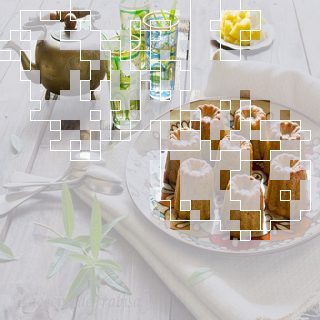}};
\node[lbl, below=1pt of Fimg] {frozen tokens (gray)};
\draw[arr] (F) -- (Fimg);
\node[blk=blue, text width=2.35cm] (D2) at (12.05,-1.55) {\textbf{active} tokens: sparse DiT\\[1pt]{\tiny gather $\to$ compute vs.\ all K/V $\to$ scatter (c)}};
\node[blk=gray, text width=2.35cm] (E) at (12.05,-2.8) {\textbf{frozen} tokens: not computed\\[2pt]$v_i=(x_{t,i}-\hat x^{\star}_{0,i})/t$};
\draw[arr] ([yshift=0.2cm]F.east) -- ++(0.35,0) |- (D2.west);
\draw[arr] ([yshift=-0.2cm]F.east) -- ++(0.35,0) |- (E.west);
\node[circle, draw=black!65, fill=white, inner sep=0.5pt, font=\sffamily\small\bfseries] (S2) at (14.0,-2.15) {$+$};
\draw[arr] (D2.east) -| (S2.north);
\draw[arr] (E.east) -| (S2.south);
\node[img] (Xm) at (14.95,-2.15) {\includegraphics[width=0.8cm]{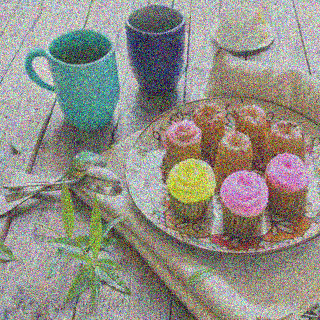}};
\node[lbl, below=1pt of Xm] {$x_{t_{j+1}}$};
\draw[arr] (S2) -- (Xm);
\draw[arr, black!45] (Xm.north) -- ++(0,1.1) -| (F.north);
\node[note=green, fill=white] at (12.05,-0.73) {next step: re-check every computed token};
\node[note=purple, text width=4.3cm] (W) at (12.3,-3.95) {\textbf{check step every $R$ steps} (adaptive, (b)): all tokens computed, caches refreshed; a frozen token whose drift rose again ($\tilde\delta_i\ge\tau m_{k(i)}$) \textbf{wakes up}};
\draw[->, purple!60, thick, dashed] (W.west) to[out=180, in=0] (Fimg.east);
\node[blk=gray, minimum height=0.9cm, minimum width=1.0cm, font=\sffamily\scriptsize\bfseries] (V) at (16.6,-2.15) {VAE\\dec.};
\node[img] (O) at (18.4,-2.15) {\includegraphics[width=1.6cm]{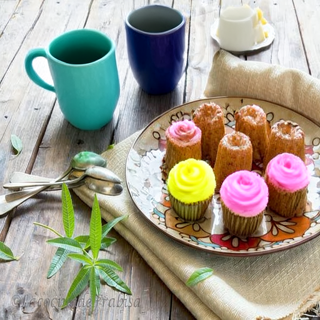}};
\node[lbl, above=1pt of O] {edited image};
\draw[arr] (Xm.east) -- (V.west);
\draw[arr] (V) -- (O);

\def\yb{-4.8}
\tpanel{0}{\yb}{5.95}{-8.05}{gray}{(a) Self-grounded regions}
\tpanel{6.15}{\yb}{12.7}{-8.05}{gray}{(b) Freeze-and-wake (adaptive re-check)}
\tpanel{12.9}{\yb}{19.4}{-8.05}{gray}{(c) Sparse step: queries vs.\ keys/values}
\node[img] (a1) at (0.65,\yb-0.95) {\includegraphics[width=0.78cm]{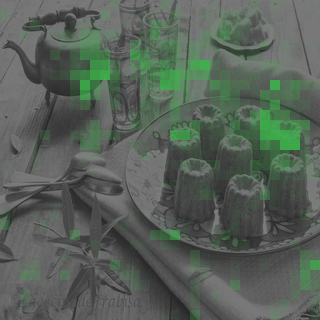}};
\node[img] (a2) at (0.65,\yb-1.83) {\includegraphics[width=0.78cm]{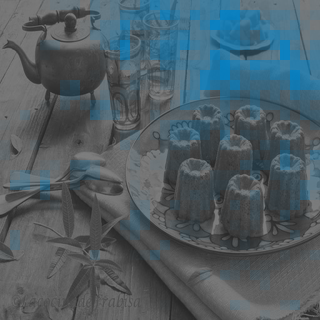}};
\node[img] (a3) at (0.65,\yb-2.71) {\includegraphics[width=0.78cm]{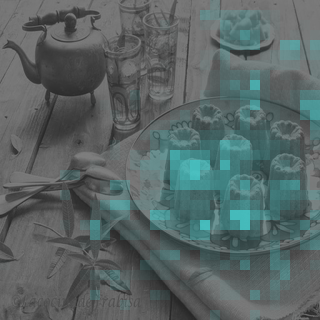}};
\node[lbl, anchor=west, align=left] at (1.1,\yb-0.95) {$A_{ki}$: ``navy\\mug''};
\node[lbl, anchor=west, align=left] at (1.1,\yb-1.83) {``turquoise\\mug''};
\node[lbl, anchor=west, align=left] at (1.1,\yb-2.71) {``pink\\cupcake''};
\node[blk=green, text width=1.6cm, font=\sffamily\tiny] (uot) at (3.55,\yb-1.83) {\textbf{unbalanced OT}\\[2pt]cost $-\log\tilde A_{ki}$\\[1pt]mass: edit demand $b_i$\\[1pt]+ TV smoothing};
\node[img] (reg) at (5.3,\yb-1.83) {\includegraphics[width=1.05cm]{fig/method_regions.png}};
\node[lbl, below=1pt of reg] {part $k(i)$};
\draw[arr] (2.15,\yb-0.95) -- (uot.north west);
\draw[arr] (2.15,\yb-1.83) -- (uot.west);
\draw[arr] (2.15,\yb-2.71) -- (uot.south west);
\draw[arr] (uot) -- (reg);
\node[lbl] at (3.55,\yb-3.02) {no mask, segmenter or VLM};
\def\cw{0.36}\def\ch{0.29}\def\gx{7.95}\def\gy{\yb-0.88}
\foreach \r/\c/\lab in {0/red!70!black/{navy mug}, 1/red!70!black/{navy mug}, 2/green!45!black/{pink cupcake}, 3/green!45!black/{pink cupcake}, 4/black!60/{background}, 5/black!60/{background}} {
  \node[lbl, anchor=east, text=\c] at (\gx-0.05,\gy-\r*\ch-\ch/2) {\lab};
}
\foreach \row [count=\ri from 0] in {{0,0,1,1,1,2,3,1,1,2,3,2},{0,0,1,2,2,2,3,2,2,2,3,2},{0,0,1,1,2,2,3,4,4,2,3,2},
                                     {0,0,1,2,2,2,3,2,2,2,3,2},{0,0,2,2,2,2,3,2,2,2,3,2},{0,0,1,2,2,2,3,2,2,2,3,2}} {
  \foreach \s [count=\ci from 0] in \row {
    \ifcase\s \def\fc{blue!45}\or \def\fc{blue!15}\or \def\fc{gray!30}\or \def\fc{magenta!35}\or \def\fc{orange!60}\fi
    \fill[\fc, draw=white, line width=0.6pt] (\gx+\ci*\cw,\gy-\ri*\ch) rectangle ++(\cw,-\ch);
  }
}
\node[lbl, anchor=east] at (\gx-0.05,\gy+0.13) {step $j$:};
\foreach \ci in {0,...,11} {\node[lbl] at (\gx+\ci*\cw+\cw/2,\gy+0.13) {\ci};}
\node[lbl, anchor=west] at (\gx,\gy-6*\ch-0.15) {(illustration)};
\foreach \fc/\t [count=\li from 0] in {blue!45/dense, blue!15/computed, gray!30/frozen, magenta!35/check, orange!60/woken} {
  \pgfmathsetmacro\lx{6.45+\li*1.22+(\li>1)*0.12}
  \fill[\fc, rounded corners=0.8pt] (\lx,\yb-2.98) rectangle ++(0.22,-0.17);
  \node[lbl, anchor=west] at (\lx+0.25,\yb-3.065) {\t};
}
\node[lbl] at (13.45,\yb-0.72) {queries};
\node[lbl] at (15.4,\yb-0.72) {keys / values (all tokens)};
\foreach \t/\c [count=\k from 0] in {T/yellow!45, R/cyan!25, A/blue!25, A/blue!25, F/gray!35, F/gray!35} {
  \node[tok, fill=\c] at (14.58+\k*0.33,\yb-1.0) {\t};
}
\foreach \k in {0,1} {\node[tok, fill=blue!25] at (13.45,\yb-1.35-\k*0.33) {A};}
\foreach \k in {0,...,5} {\foreach \q in {0,1} {
  \ifnum\k=2 \def\cc{0}\else\ifnum\k=3 \def\cc{0}\else \def\cc{1}\fi\fi
  \ifnum\cc=1
    \draw[red!70, fill=white, line width=0.7pt, rounded corners=0.8pt] (14.58+\k*0.33-0.13,\yb-1.35-\q*0.33+0.13) rectangle ++(0.26,-0.26);
  \else
    \draw[blue!50!black, fill=blue!15, rounded corners=0.8pt] (14.58+\k*0.33-0.13,\yb-1.35-\q*0.33+0.13) rectangle ++(0.26,-0.26);
  \fi
}}
\node[lbl, anchor=north west, text width=6.05cm, align=left] at (13.05,\yb-2.05) {%
\textcolor{blue!60}{$\blacksquare$} computed now: the active rows (with many steps also the text rows)\\
\textcolor{red!70}{$\square$} cached K/V: frozen tokens (last computed step) and the reference (refreshed at check steps)\\
every computed token still attends to \emph{all} tokens; with all rows active the step equals the dense forward};
\node[lbl, anchor=north west, align=left] at (16.9,\yb-0.85) {T\ text\\R\ reference\\A\ active\\F\ frozen};
\end{tikzpicture}}
\vspace{-0.4em}
\caption{\textbf{Overview of \method.} \ding{172}~Dense steps record keys/values, word attention and drift. \ding{173}~Unbalanced OT assigns every token to an instruction part (a); a token is frozen when its drift is below $\tau$ times its \emph{own} part's mean drift, and only active tokens are computed (c); every $R$ steps a check step computes all tokens and wakes frozen tokens that move again (b). \ding{174}~Decode.}
\label{fig:pipeline}
\end{figure*}

\subsection{Problem Setting and Method Overview}
\label{sec:setting}
Let $v_\theta$ be an in-context rectified-flow editor with frozen weights that
processes one joint sequence of $T$ text tokens, $N$ target tokens $x_t$ and $N$
reference tokens $r$ (the encoded source image), with Euler steps
$x_{t_{j+1}}=x_{t_j}-h_j\,v_\theta(x_{t_j},t_j)$ from $t_0{=}1$ to $t_n{=}0$,
$h_j=t_j-t_{j+1}$. Rectified flow moves along straight paths
$x_t=(1-t)x_0+t\,x_1$ from noise $x_1$ to data $x_0$ with velocity
$v=x_1-x_0$, so every step yields an \emph{endpoint estimate} and, per target
token $i$, a \emph{drift}:
\begin{equation}
\label{eq:endpoint}
\hat x_0^{(j)}=x_{t_j}-t_j\,v_\theta(x_{t_j},t_j),\qquad
\delta_{i,j}=\big\|\hat x_{0,i}^{(j)}-\hat x_{0,i}^{(j-1)}\big\| .
\end{equation}
A small drift means that the token's predicted result has stopped changing. The
instruction is split into parts $k=1,\dots,K$, one per requested edit (given by
the benchmark, or split at commas, semicolons and ``and''); part $0$ is the
background, and each image token $i$ is assigned a part $k(i)$. Skipping token
$i$ at step $j$ saves its compute $c_{i,j}$ and adds an error
$a_{i,j}\propto(h_j/t_j)\,\delta_{i,j}$ to its part (\cref{prop:endpoint}). The
objective is the largest saving with \emph{every part} within its own error
budget:
\begin{equation}
\label{eq:budget}
\max_{z\in\{0,1\}^{N\times n}}\ \sum_{i,j}c_{i,j}z_{i,j}
\quad\text{s.t.}\quad
\sum_{i:\,k(i)=k}\ \sum_j a_{i,j}z_{i,j}\le\varepsilon_k\ \ \forall k ,
\end{equation}
where $z_{i,j}=1$ skips token $i$ at step $j$; a global threshold applies one
skipping rule to all $K{+}1$ parts. For example, next to an object swap, a subtle
recolor whose tokens drift least is frozen whole by any single threshold that
saves anything, whereas one threshold per part saves most of the swap within both
budgets (supplement, Sec.~A.3).

Solving \eqref{eq:budget} needs the parts $k(i)$ and a skipping rule per part,
and the editor computes the signals for both in its first steps. \method
therefore has three stages (\cref{fig:pipeline}):
\textbf{(1)~Regions}: during the first $s$ steps, which are computed densely,
find the region of every instruction part from the editor's attention
(\cref{sec:regions}); \textbf{(2)~Freezing}: from then on, freeze tokens whose
drift is small relative to their own part (\cref{sec:freeze}); and
\textbf{(3)~Execution}: compute only the remaining tokens with a sparse engine
(\cref{sec:engine}). The editor's weights stay fixed and no mask, segmenter or
VLM is used, so the method is training-free.

\subsection{Self-grounded Regions}
\vspace{-4pt}
\label{sec:regions}
\runin{Word attention and edit demand} For part $k$ let $W_k$ be the
content words of the clause that names its target (words shared by two parts are
dropped). During the dense steps we accumulate the attention of image token $i$
on these words over all heads $\eta$ of all double-stream blocks $\ell$,
\[
A_{ki}=\sum_{j<s}\ \sum_{(\ell,\eta)}\ \sum_{w\in W_k}
\mathrm{softmax}_w\big(q^{\ell\eta}_{i}\cdot k^{\ell\eta}_{w}/\sqrt d\big),
\]
with the softmax taken over all keys of the joint sequence. These maps localize
each part but compete with each other (\cref{fig:pipeline}a): a subtle part is
out-attended by salient ones on most tokens. The flow gives a second, text-free
signal, the distance of a token's predicted result from the source,
$\omega_i=\|\hat x_{0,i}^{(s-1)}-r_i\|$ (target and reference tokens share one
grid); its robust normalization
$b_i\propto\mathrm{softplus}\big((\omega_i-\mathrm{med}\,\omega)/\mathrm{MAD}\,\omega\big)$
(MAD: median absolute deviation) is the mass each token must give to some part.

\runin{Unbalanced optimal transport} With cost
$C_{ki}=-\log\tilde A_{ki}$, $\tilde A_{ki}=A_{ki}/\sum_{i'}A_{ki'}$, and part
masses $\alpha_k\propto\sum_i\tilde A_{ki}b_i$, we solve
\begin{equation}
\label{eq:uot}
\begin{split}
P^\star=\argmin_{P\ge0}\ & \langle C,P\rangle-\zeta H(P) \\
& +\rho_1\mathrm{KL}(P\mathbf 1\,\|\,\alpha)
  +\rho_2\mathrm{KL}(P^{\!\top}\mathbf 1\,\|\,b)
\end{split}
\end{equation}
with entropy $H$, the generalized Kullback--Leibler divergence KL of non-negative
vectors, $\zeta{=}0.12$ and $\rho_1{=}\rho_2{=}1$, by the scaling iterations
of~\cite{uot}. The responsibilities $\pi_{ki}=P^\star_{ki}/\sum_lP^\star_{li}$ are
smoothed with total variation on the token grid~\cite{chambollepock}; token $i$
joins the part that dominates its responsibility if its demand mass passes a small
gate, and the background otherwise (gate values in the supplement).
Transport \eqref{eq:uot} keeps a part that is out-attended everywhere, which argmax drops
(\cref{thm:mass}); the solve takes 4.2\,ms.

\subsection{Per-instruction Freezing}
\label{sec:freeze}
\runin{A threshold per part} Relaxing \eqref{eq:budget} with one Lagrange multiplier per part gives a threshold on the error per unit
of compute $a_{i,j}/c_{i,j}$ that depends only on the part (\cref{thm:price}); at a given step this is a threshold on the drift, which we set
relative to the part's own motion. The budgets $\varepsilon_k$ are therefore implicit: instead of solving \eqref{eq:budget} for given
$\varepsilon_k$, we fix a price per part that scales with the part's drift. With $m_k=\mathrm{mean}_{i:\,k(i)=k}\,\delta_{i,s-1}$ the mean drift of part $k$ at the first
decision step $j{=}s$ (image-wide for the background), a token that is still computed is frozen at the start of step $j\ge s$ if
\begin{equation}
\label{eq:rule}
\delta_{i,j-1}<\tau\,m_{k(i)} ,
\end{equation}
with one $\tau$ for all parts, steps, images and benchmarks. A slowly moving part is thus judged against slow motion and a fast part against fast
motion, whereas one global threshold $\tau\bar\delta$ ($\bar\delta$: image-wide mean drift) compares every token with the same number. Steps $0,\dots,s{-}1$ compute every target token ($s{=}2$ on the 4-step editor); afterwards
\eqref{eq:rule} is re-checked at every step for every token still computed, and frozen tokens stay frozen (\emph{freeze-only}).

\runin{Endpoint-freeze} A frozen token is not computed. It keeps its last endpoint estimate $\hat x^\star_{0,i}$ and moves with
\begin{equation}
\label{eq:freeze}
v_i=\big(x_{t,i}-\hat x^\star_{0,i}\big)/t ,
\end{equation}
which keeps it exactly on the straight path to $\hat x^\star_{0,i}$ (\cref{prop:endpoint}) with no stored history; its error per step is exactly
$(h_j/t_j)$ times the endpoint drift it ignores, the quantity that \eqref{eq:rule} controls.

\runin{Adaptive wake-up for many-step editors} With tens of steps a token frozen early has far to go, so every $R$ steps a \emph{check
step} computes all tokens, re-measures $m_k$ and the drifts, rebuilds the frozen set with \eqref{eq:rule} (waking frozen tokens that move again) and
refreshes the cached text and reference keys and values. With guidance, the empty-prompt branch is computed every third sparse step and its
difference to the prompt branch reused in between; the last step is dense.

\subsection{Sparse Execution}
\label{sec:engine}
At a sparse step with active set $\mathcal A_j$, each block computes queries, keys and values only for the rows in $\mathcal A_j$ and writes them
into a per-block buffer that also holds the cached keys and values of the text, reference and frozen tokens. Active queries attend to the full buffer $\bar K=[K_{\mathrm{txt}};K_{\mathrm{ref}};K_{\mathcal F};K_{\mathcal A}]$ (and $\bar V$ alike),
\begin{equation}
\label{eq:attn}
o_{\mathcal A}=\mathrm{softmax}\big(Q_{\mathcal A}\bar K^{\top}/\sqrt d\big)\,\bar V ,
\end{equation}
so every computed token still sees every other token (frozen ones with the keys and values of their last computed step), while projections and MLPs cost $O(|\mathcal A_j|)$ rows instead of $O(T{+}2N)$. The engine
is plain PyTorch and reproduces the dense forward exactly when all rows, text and reference included, are computed (algorithm in the supplement). Only step 0 is a full forward; later, text and reference rows reuse the keys and values of the last full step (as FLUX.2 [klein] 9B-KV does for the reference), an approximation the supplement separates from freezing.

%% file: sec/4_theory.tex
\subsection{Analysis}
\label{sec:theory}
\begin{table}[t]
\centering
\tabsize
\renewcommand{\arraystretch}{0.88}
\begin{tabular}{@{}lrrrrrrr@{}}
\toprule
 & \multicolumn{3}{c}{One global threshold} & \multicolumn{3}{c}{TIDE (ours)} & \\
\cmidrule(lr){2-4}\cmidrule(lr){5-7}
Setting & Spd & Drop$\downarrow$ & All\%$\uparrow$ & Spd & Drop$\downarrow$ & All\%$\uparrow$ & $\Delta$Drop \\
\midrule
MICE, 8 steps & 2.56$\times$ & \textbf{151} & 35.1 & 2.61$\times$ & 154 & 32.4 & $+$2\% \\
& 3.46$\times$ & 384 & 25.2 & 3.51$\times$ & \textbf{200} & 31.7 & $-$48\% \\
& 3.88$\times$ & 269 & 28.2 & 4.14$\times$ & \textbf{169} & 33.6 & $-$37\% \\
& 4.27$\times$ & 157 & 34.4 & 4.29$\times$ & \textbf{153} & 33.6 & $-$3\% \\
\midrule
MICE, 4 steps & 1.93$\times$ & \textbf{162} & 32.1 & 1.97$\times$ & 166 & 31.3 & $+$2\% \\
& 2.09$\times$ & 237 & 29.8 & 2.20$\times$ & \textbf{155} & 32.1 & $-$35\% \\
\midrule
LoMOE, 8 steps & 3.78$\times$ & 26 & 51.6 & 4.04$\times$ & \textbf{6} & 65.6 & $-$77\% \\
MIRA, 8 steps & 3.84$\times$ & 74 & 15.0 & 4.14$\times$ & \textbf{42} & 23.0 & $-$43\% \\
InfoDet, 8 steps & 3.62$\times$ & 23 & 28.0 & 3.85$\times$ & \textbf{13} & 29.0 & $-$43\% \\
Crello, 8 steps & 3.81$\times$ & 57 & 33.0 & 4.06$\times$ & \textbf{33} & 46.0 & $-$42\% \\
\bottomrule
\end{tabular}
\caption{\textbf{Per-instruction vs.\ one global threshold at matched or higher speed}. Same engine and regions; only the freezing rule
differs. All: images with every edit done. $\Delta$Drop $\le-$35\%: $p\le0.031$, on MICE $p<10^{-12}$ (supplement, paired tests).}
\label{tab:claim}
\end{table}

\begin{table}[t]
\centering
\tabsize
\resizebox{\linewidth}{!}{\begin{tabular}{@{}lrrrrr@{}}
\toprule
Object area & $<$1\% & 1--2\% & 2--4\% & 4--8\% & $\ge$8\% \\
\midrule
Global threshold, 4 steps & 19.3 & 17.0 & 11.6 & 14.7 & 9.3 \\
\textbf{TIDE}, 4 steps & \textbf{14.7} & \textbf{12.4} & \textbf{6.3} & \textbf{6.9} & \textbf{5.6} \\
\midrule
Global threshold, 8 steps & 32.9 & 25.1 & 23.2 & 18.5 & 13.6 \\
\textbf{TIDE}, 8 steps & \textbf{17.1} & \textbf{17.1} & \textbf{8.7} & \textbf{10.1} & \textbf{6.2} \\
\bottomrule
\end{tabular}}
\caption{\textbf{Which edits are dropped} (MICE-Bench; \% of the edits vanilla completes that the method loses), by the edited object's mask
area. Matched speed: 2.09 vs.\ 2.20$\times$ (4 steps), 3.46 vs.\ 3.51$\times$ (8 steps).}
\label{tab:supp_size}
\end{table}

\begin{table*}[t]
\centering
\tabsize
\resizebox{\textwidth}{!}{\begin{tabular}{@{}lrrrrrrrrrrrrrrrrrr@{}}
\toprule
 & \multicolumn{3}{c}{MICE} & \multicolumn{3}{c}{LoMOE} & \multicolumn{3}{c}{MIRA} & \multicolumn{3}{c}{InfoDet} & \multicolumn{3}{c}{Crello} & \multicolumn{3}{c}{GEdit} \\
\cmidrule(lr){2-4}\cmidrule(lr){5-7}\cmidrule(lr){8-10}\cmidrule(lr){11-13}\cmidrule(lr){14-16}\cmidrule(lr){17-19}
Method & Spd$\uparrow$ & Drop$\downarrow$ & LP$\downarrow$ & Spd$\uparrow$ & Drop$\downarrow$ & LP$\downarrow$ & Spd$\uparrow$ & Drop$\downarrow$ & LP$\downarrow$ & Spd$\uparrow$ & Drop$\downarrow$ & LP$\downarrow$ & Spd$\uparrow$ & Drop$\downarrow$ & LP$\downarrow$ & Spd$\uparrow$ & Drop$\downarrow$ & LP$\downarrow$ \\
\midrule
Vanilla & 1.00$\times$ & -- & 0 & 1.00$\times$ & -- & 0 & 1.00$\times$ & -- & 0 & 1.00$\times$ & -- & 0 & 1.00$\times$ & -- & 0 & 1.00$\times$ & -- & 0 \\
\midrule
Step skip & 1.84$\times$ & 57 & 0.145 & 1.82$\times$ & 3 & 0.076 & 1.82$\times$ & 11 & 0.044 & 1.82$\times$ & 9 & 0.028 & 1.80$\times$ & 18 & 0.023 & 1.81$\times$ & 13 & 0.060 \\
RegionE~\cite{regione} & 1.30$\times$ & \textbf{56} & \textbf{0.113} & 1.35$\times$ & \textbf{1} & 0.065 & 1.40$\times$ & 15 & 0.045 & 1.51$\times$ & 14 & \textbf{0.028} & 1.52$\times$ & 21 & 0.030 & 1.38$\times$ & 19 & 0.047 \\
TaylorSeer~\cite{taylorseer} & 1.79$\times$ & 259 & 0.365 & 1.79$\times$ & 21 & 0.303 & 1.79$\times$ & 57 & 0.246 & 1.78$\times$ & 20 & 0.102 & 1.79$\times$ & 30 & 0.134 & 1.79$\times$ & 89 & 0.283 \\
TeaCache~\cite{teacache} & 1.77$\times$ & 91 & 0.180 & 1.80$\times$ & 3 & 0.101 & 1.81$\times$ & 23 & 0.061 & 1.82$\times$ & 19 & 0.030 & 1.82$\times$ & 25 & \textbf{0.027} & 1.78$\times$ & \textbf{17} & 0.076 \\
RAS~\cite{ras} & 1.65$\times$ & 127 & 0.216 & 1.65$\times$ & 5 & 0.135 & 1.66$\times$ & 41 & 0.098 & 1.67$\times$ & 16 & 0.056 & 1.64$\times$ & 30 & 0.062 & 1.67$\times$ & 27 & 0.109 \\
SpotEdit~\cite{spotedit} & 1.54$\times$ & 87 & 0.150 & 1.67$\times$ & 3 & 0.108 & 1.77$\times$ & 85 & 0.090 & 1.95$\times$ & 18 & 0.038 & 1.97$\times$ & 45 & 0.043 & 1.63$\times$ & 50 & 0.077 \\
MICE$^\dagger$~\cite{mice} & 0.47$\times$ & 263 & 0.330 & 0.74$\times$ & 15 & 0.199 & 0.61$\times$ & 116 & 0.150 & 0.61$\times$ & 23 & 0.056 & 0.74$\times$ & 31 & 0.064 & \multicolumn{3}{c}{n/a (no masks)} \\
MIRAGE$^\ddagger$~\cite{mirage} & 0.10$\times$ & 135 & 0.241 & 0.22$\times$ & 13 & 0.130 & 0.11$\times$ & 21 & 0.078 & 0.09$\times$ & 14 & 0.036 & 0.15$\times$ & \textbf{4} & 0.034 & 0.37$\times$ & 37 & 0.082 \\
\midrule
\textbf{TIDE-fast (ours)} & \textbf{2.20$\times$} & 155 & 0.222 & \textbf{2.22$\times$} & 7 & 0.046 & \textbf{2.24$\times$} & 9 & 0.037 & \textbf{2.20$\times$} & 8 & 0.053 & \textbf{2.24$\times$} & 28 & 0.061 & \textbf{2.21$\times$} & 25 & 0.042 \\
\textbf{TIDE-quality (ours)} & 1.79$\times$ & 121 & 0.204 & 1.89$\times$ & 8 & \textbf{0.034} & 1.92$\times$ & \textbf{8} & \textbf{0.034} & 1.93$\times$ & \textbf{7} & 0.058 & 1.92$\times$ & 31 & 0.061 & 1.85$\times$ & 24 & \textbf{0.041} \\
\bottomrule
\end{tabular}}
\caption{\textbf{All methods on six benchmarks} (FLUX.2 Klein 4B, 4 steps, 512$^2$). Spd: end-to-end speed-up over vanilla; Drop: edits
vanilla completes that the method loses (VLM judge, lower is better); LP: LPIPS to vanilla's output. Bold: best accelerator; step skip
(fewer steps) is a reference and not marked. $^\dagger$needs a mask per edit; $^\ddagger$includes VLM grounding time.}
\label{tab:main}
\end{table*}

\begin{table*}[t]
\centering
\tabsize
\begin{minipage}[b]{0.57\linewidth}\centering\resizebox{\linewidth}{!}{\begin{tabular}{@{}lrrrrrrr@{}}
\toprule
Method & Text & Ref.\ VAE & Transformer & VAE dec. & Total & $S_\mathrm{dec}$ & $S_\mathrm{e2e}$ \\
\midrule
Vanilla & 1.00 & 0.31 & 18.69 & 0.74 & 20.74 & 1.00$\times$ & 1.00$\times$ \\
Step skip & 1.00 & 0.29 & 9.31 & 0.67 & 11.27 & 2.01$\times$ & 1.84$\times$ \\
RegionE & 1.00 & 0.31 & 13.92 & 0.73 & 15.95 & 1.34$\times$ & 1.30$\times$ \\
TaylorSeer & 1.01 & 0.31 & 9.51 & 0.71 & 11.54 & 1.96$\times$ & 1.80$\times$ \\
TeaCache & 1.01 & 0.31 & 9.61 & 0.74 & 11.67 & 1.94$\times$ & 1.78$\times$ \\
RAS & 1.00 & 0.31 & 10.47 & 0.74 & 12.52 & 1.78$\times$ & 1.66$\times$ \\
SpotEdit & 0.99 & 0.29 & 11.42 & 0.66 & 13.36 & 1.64$\times$ & 1.55$\times$ \\
\midrule
\textbf{TIDE-fast} & 1.01 & 0.31 & \textbf{7.36} & 0.73 & \textbf{9.42} & \textbf{2.54}$\times$ & \textbf{2.20}$\times$ \\
\textbf{TIDE-quality} & 1.01 & 0.32 & 9.52 & 0.76 & 11.60 & 1.96$\times$ & 1.79$\times$ \\
\bottomrule
\end{tabular}}\end{minipage}\hfill
\begin{minipage}[b]{0.41\linewidth}\centering\resizebox{\linewidth}{!}{\begin{tabular}{@{}lrrrrr@{}}
\toprule
Benchmark & Parts & Found & Exact & IoU & OT ms \\
\midrule
MICE & 8.06 & 6.91 & 78\% & 0.227 & 4.2 \\
LoMOE & 3.00 & 2.47 & 80\% & 0.334 & 3.7 \\
MIRA & 5.00 & 5.00 & 100\% & 0.332 & 3.6 \\
InfoDet & 4.95 & 4.73 & 87\% & 0.142 & 3.8 \\
Crello & 2.94 & 2.81 & 93\% & 0.215 & 3.4 \\
\bottomrule
\end{tabular}}\end{minipage}
\caption{\textbf{Left: latency} (s per image, Klein 4B, MICE-Bench): encoders, transformer, VAE, total; $S_\mathrm{dec}$ / $S_\mathrm{e2e}$:
transformer / end-to-end speed-up from the mean stage times (can differ from \cref{tab:main} by 0.01$\times$). \textbf{Right: self-grounded regions}
of TIDE-fast vs.\ benchmark masks (used only here): parts per image, parts given a region, images with all parts found, IoU of each object mask
with its best-matching region (32$\times$32 token grid), OT time.}
\label{tab:supp_latency}
\end{table*}

In this section, we analyze the design of \cref{sec:regions,sec:freeze} under a linearized error model, in which the errors of frozen tokens add
up: the error of a frozen token is a known multiple of its drift (\cref{prop:endpoint}), budgets must be set per part
(\cref{thm:price,thm:agnostic}), and transport keeps weak parts that argmax loses (\cref{thm:mass}). Proofs are in the supplement, and
\cref{sec:premise} checks the premises on real trajectories. The results concern the class of per-part threshold policies; Rule~\eqref{eq:rule}
belongs to it, and Prop.~D (supplement) gives the condition under which it is the optimum of this class.
\begin{figure*}[t]
\centering
\pgfplotsset{pareto/.style={scale only axis,width=0.262\textwidth,height=2.5cm,grid=major,grid style={gray!15},
  tick label style={font=\scriptsize},label style={font=\scriptsize},title style={font=\footnotesize\bfseries,yshift=-5pt},
  xlabel={Speed-up over vanilla},xlabel style={yshift=4pt},ylabel style={yshift=-6pt},clip=false,
  axis line style={black!60}}}
{\scriptsize
\tikz[baseline=-0.6ex]\node[draw=cvprblue,fill=cvprblue,minimum size=4pt,inner sep=0pt]{};~\textbf{\method}\hspace{0.9em}
\tikz[baseline=-0.6ex]\node[draw=cvprblue,thick,minimum size=4pt,inner sep=0pt]{};~\method\ (2 steps / +TeaCache)\hspace{0.9em}
\tikz[baseline=-0.6ex]\node[draw=purple!70!black,minimum size=4pt,inner sep=0pt]{};~global threshold\hspace{0.9em}
\tikz[baseline=-0.6ex]\fill[orange!85!black](0,0.075)--(0.065,0)--(0,-0.075)--(-0.065,0)--cycle;~step skip\hspace{0.9em}
\tikz[baseline=-0.6ex]\fill[black!75](-0.065,-0.055)--(0.065,-0.055)--(0,0.065)--cycle;~TeaCache\hspace{0.9em}
\tikz[baseline=-0.6ex]\node[circle,fill=black!40,minimum size=3.5pt,inner sep=0pt]{};~other methods\hspace{0.9em}
\tikz[baseline=-0.6ex]\draw[gray!70,dashed,thick](0,0)--(0.45,0);~Pareto frontier}\\[1pt]
\begin{tikzpicture}
\begin{axis}[pareto,title={(a) Per part vs.\ one global threshold},ylabel={Edits dropped$\downarrow$},xmin=1.8,xmax=4.5,ymin=100,ymax=400,
  xlabel={Speed-up over vanilla, same steps}]
\addplot[purple!70!black,mark=square,mark size=1.4pt,mark options={solid}] coordinates {(2.56,151) (3.23,359) (3.46,384) (3.88,269) (4.27,157)};
\addplot[cvprblue,thick,mark=square*,mark size=1.7pt,mark options={solid}] coordinates {(2.61,154) (3.51,200) (4.14,169) (4.29,153)};
\addplot[purple!70!black,dashed,mark=square,mark size=1.4pt,mark options={solid}] coordinates {(1.93,162) (2.09,237) (2.24,218)};
\addplot[cvprblue,thick,dashed,mark=square*,mark size=1.7pt,mark options={solid}] coordinates {(1.97,166) (2.20,155) (2.29,140)};
\draw[->,thick,green!45!black] (axis cs:3.58,372) -- (axis cs:3.58,212);
\node[font=\scriptsize\bfseries,text=green!45!black,anchor=west,inner sep=1pt] at (axis cs:3.62,290) {$-$48\%};
\node[font=\scriptsize,text=purple!70!black,anchor=east,inner sep=2pt] at (axis cs:3.2,384) {global, 8 steps};
\node[font=\scriptsize,text=cvprblue,inner sep=1pt] at (axis cs:3.02,218) {\textbf{\method}, 8 steps};
\node[font=\scriptsize,text=purple!70!black,anchor=south,inner sep=1.5pt] at (axis cs:2.09,240) {global, 4};
\node[font=\scriptsize,text=cvprblue,anchor=north,inner sep=2pt] at (axis cs:2.13,134) {\textbf{\method}, 4};
\end{axis}
\end{tikzpicture}\hfill
\begin{tikzpicture}
\begin{axis}[pareto,title={(b) Klein 4B, all methods (4 steps)},xmin=0.9,xmax=3.3,ymin=0,ymax=290]
\fill[cvprblue!7] (axis cs:1.84,0) rectangle (axis cs:2.43,290);
\node[font=\tiny\itshape,text=cvprblue!80!black,anchor=north,align=center,inner sep=1pt] at (axis cs:2.135,288) {frontier\\held by \method};
\addplot[gray!70,dashed,thick] coordinates {(1.00,0) (1.00,23) (1.04,23) (1.04,49) (1.29,49) (1.29,56) (1.30,56) (1.30,57) (1.84,57) (1.84,130) (2.43,130) (2.43,238) (3.10,238)};
\addplot[orange!85!black,thick,dotted] coordinates {(1.84,57) (3.10,238)};
\addplot[orange!85!black,thick,mark=diamond*,mark size=1.9pt] coordinates {(1.00,0) (1.29,49) (1.84,57)};
\addplot[only marks,orange!85!black,mark=diamond*,mark size=1.9pt] coordinates {(3.10,238)};
\addplot[black!75,mark=triangle*,mark size=1.8pt] coordinates {(1.04,23) (1.29,57) (1.34,78) (1.77,91)};
\addplot[purple!70!black,mark=square,mark size=1.4pt,mark options={solid}] coordinates {(1.93,162) (2.09,237) (2.24,218)};
\addplot[cvprblue,thick,mark=square*,mark size=1.7pt] coordinates {(1.97,166) (2.20,155) (2.29,140)};
\addplot[only marks,cvprblue,mark=square*,mark size=1.7pt] coordinates {(1.79,121)};
\addplot[only marks,cvprblue,mark=square,mark size=1.7pt,mark options={line width=0.8pt}] coordinates {(2.43,130)};
\addplot[only marks,mark=*,black!40,mark size=1.4pt] coordinates {(1.30,56) (1.79,259) (1.65,127) (1.54,87)};
\node[font=\tiny,text=black!65,anchor=north west,inner sep=1pt] at (axis cs:1.33,44) {RegionE};
\node[font=\tiny,text=black!65,anchor=east,inner sep=2pt] at (axis cs:1.79,259) {TaylorSeer};
\node[font=\tiny,text=black!65,anchor=south,inner sep=2pt] at (axis cs:1.65,129) {RAS};
\node[font=\tiny,text=black!65,anchor=south east,inner sep=1pt] at (axis cs:1.53,89) {SpotEdit};
\node[font=\tiny,text=orange!70!black,anchor=north west,inner sep=1pt] at (axis cs:1.86,55) {2 steps};
\node[font=\tiny,text=orange!70!black,anchor=north,inner sep=2pt] at (axis cs:3.10,234) {1 step};
\node[font=\tiny,text=purple!70!black,anchor=west,inner sep=2pt] at (axis cs:2.28,228) {global};
\node[font=\tiny\bfseries,text=cvprblue,anchor=south west,inner sep=1pt] at (axis cs:2.22,158) {\method};
\node[font=\tiny\bfseries,text=cvprblue,anchor=north,inner sep=2pt] at (axis cs:1.79,117) {\method-q};
\node[font=\tiny,text=cvprblue,anchor=west,inner sep=2pt] at (axis cs:2.46,130) {\method, 2 steps};
\end{axis}
\end{tikzpicture}\hfill
\begin{tikzpicture}
\begin{axis}[pareto,title={(c) Klein Base, all methods (28 steps)},xmin=0.5,xmax=6.5,ymin=0,ymax=140]
\fill[cvprblue!7] (axis cs:4.5,0) rectangle (axis cs:6.5,140);
\node[font=\tiny\itshape,text=cvprblue!80!black,anchor=north,align=center,inner sep=1pt] at (axis cs:5.5,138) {beyond 4.5$\times$:\\only \method};
\addplot[gray!70,dashed,thick] coordinates {(1.00,0) (1.00,25) (2.00,25) (2.00,26) (2.76,26) (2.76,27) (3.86,27) (3.86,35) (4.50,35) (4.50,38) (4.94,38) (4.94,39) (5.23,39)};
\addplot[orange!85!black,thick,mark=diamond*,mark size=1.9pt] coordinates {(1.00,0) (2.00,25) (3.91,36)};
\addplot[black!75,mark=triangle*,mark size=1.8pt] coordinates {(2.76,26) (3.86,27) (4.50,35)};
\addplot[only marks,black!75,mark=triangle,mark size=1.8pt] coordinates {(2.68,35)};
\addplot[only marks,mark=*,black!40,mark size=1.4pt] coordinates {(2.01,132) (1.50,96) (3.66,68) (1.74,59) (0.60,77)};
\addplot[only marks,cvprblue,mark=square*,mark size=1.8pt] coordinates {(4.85,46) (5.23,39)};
\addplot[only marks,cvprblue,mark=square,mark size=1.7pt,mark options={line width=0.8pt}] coordinates {(4.82,39) (4.94,38) (4.93,47)};
\node[font=\tiny,text=black!65,anchor=west,inner sep=2pt] at (axis cs:2.01,132) {no guidance};
\node[font=\tiny,text=black!65,anchor=west,inner sep=2pt] at (axis cs:1.50,96) {RegionE};
\node[font=\tiny,text=black!65,anchor=west,inner sep=2pt] at (axis cs:3.66,68) {TaylorSeer};
\node[font=\tiny,text=black!65,anchor=west,inner sep=2pt] at (axis cs:1.74,59) {RAS};
\node[font=\tiny,text=black!65,anchor=west,inner sep=2pt] at (axis cs:0.60,77) {MIRAGE};
\node[font=\tiny,text=orange!70!black,anchor=north west,inner sep=1pt] at (axis cs:2.02,23) {14 steps};
\node[font=\tiny,text=orange!70!black,anchor=south,inner sep=2pt] at (axis cs:3.91,38) {7 steps};
\node[font=\tiny,text=black!70,anchor=south,inner sep=2pt] at (axis cs:2.6,37) {TeaCache, 14 steps};
\node[font=\tiny\bfseries,text=cvprblue,anchor=south,inner sep=2pt] at (axis cs:4.85,49) {\method};
\node[font=\tiny\bfseries,text=cvprblue,anchor=north,inner sep=2pt] at (axis cs:5.23,36) {14 steps};
\end{axis}
\end{tikzpicture}
\vspace{-0.6em}
\caption{\textbf{Speed vs.\ dropped edits on MICE-Bench} (right: faster, down: fewer drops). (a) Per part vs.\ one global threshold, one point per $\tau$
(Klein 4B). (b) Klein 4B, all methods. (c) Klein Base, all methods, first 100 images.}
\label{fig:pareto}
\end{figure*}
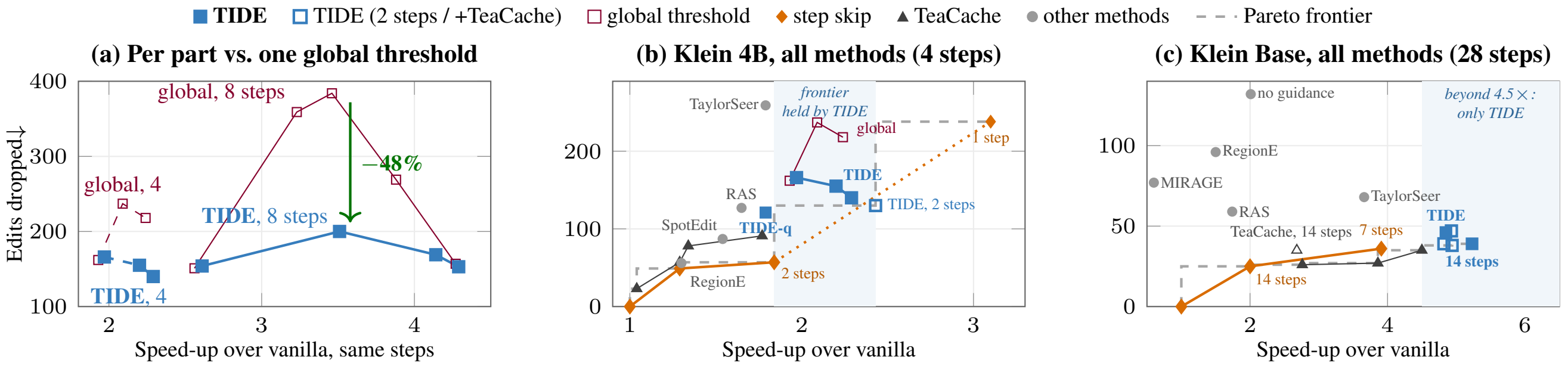

\begin{proposition}[Exact straight-path freezing]\leavevmode
\label{prop:endpoint}
Let token $i$ be frozen at time $t$ with endpoint $\hat x^\star_{0,i}$ and implied noise $\tilde x_{1,i}=(x_{t,i}-(1-t)\hat x^\star_{0,i})/t$.
Euler steps with $v_i=(x_{t,i}-\hat x^\star_{0,i})/t$ keep $x_{t',i}=(1-t')\hat x^\star_{0,i}+t'\tilde x_{1,i}$ for every later $t'$, and the
one-step deviation from the dense update is $(h/t)\,\|\hat x^\star_{0,i}-\hat x^{\theta}_{0,i}\|$, with $\hat x^{\theta}_{0,i}$ the endpoint the editor would
predict at that step.
\end{proposition}

\noindent\textbf{Budgets}: We restate \eqref{eq:budget} over \emph{items} $u=(i,j)\in\mathcal I$, token $i$ at sparse step $j$, with part
$\kappa(u)=k(i)\in\{0,\dots,K\}$. Freezing $u$ saves cost $c_u=c_{i,j}$ and adds error $a_u=a_{i,j}=w_i g_j\delta_{i,j}$ to its part, with a token
weight $w_i$ ($w_i{=}1$ in Rule~\eqref{eq:rule}) and $g_j=h_j/t_j$ up to the editor's amplification (supplement, Sec.~B). A policy
$z\in\{0,1\}^{\mathcal I}$ must keep every part within its budget, $E_k(z)=\sum_{\kappa(u)=k}a_u z_u\le\varepsilon_k$ ($\varepsilon_k\ge0$), and a part is
\emph{dropped}, the model counterpart of a lost edit, if all its items are frozen. With score $r_u=a_u/c_u$, a \emph{global threshold} policy freezes $r_u\le\theta$; a \emph{per-part}
policy freezes $r_u\le\theta_{\kappa(u)}$. Let $\beta_k$ be the smallest score at which part $k$ exceeds its budget (its \emph{breaking score})
and $S_k^-(\beta)$ the saving of its items with score below $\beta$.

\begin{theorem}[Exact price of a global threshold]\leavevmode
\label{thm:price}
The best feasible global threshold saves $S_{\mathrm{glob}}=\sum_k S_k^-(\beta_{\min})$ with $\beta_{\min}=\min_k\beta_k$; the best per-part
policy saves $S_{\mathrm{pc}}=\sum_k S_k^-(\beta_k)\ge S_{\mathrm{glob}}$, strictly iff some part $k$ has an item with
$\beta_{\min}\le r_u<\beta_k$. Per-part thresholds are within the cost of one tie class per part of the optimal 0-1 policy.
At a target saving $\rho$, the smallest global threshold $\theta_\rho$ reaching it violates part $k$ iff $\theta_\rho\ge\beta_k$ and drops it iff
$\theta_\rho\ge\max_{\kappa(u)=k}r_u$.
\end{theorem}
Thus a global policy is held to the part that breaks first; when that part is the smallest edit, a global threshold tuned for speed removes it. Rule~\eqref{eq:rule} is this per-part policy applied at each step (where $g_j$
is fixed), with uniform costs and a price per part calibrated to the part's own drift. These results diagnose which edits a global threshold drops (\cref{sec:claim}); they are not a guarantee for Rule~\eqref{eq:rule}. Its prices
rank the parts like their oracle breaking scores on the 4-step editor (Spearman 0.61), less so at 8 steps (0.28) and not on Klein Base
(0.06), where the parts' drift scales have not yet separated (supplement).

\begin{theorem}[Rules blind to the parts cannot guarantee heterogeneous budgets]\leavevmode
\label{thm:agnostic}
There are instances with two parts $A,B$ of $n_A,n_B$ items with identical features and cost $c$, on which a per-part policy
saves $n_A c$ within budget, while every deterministic rule that sees only token features saves nothing or violates the budget of $B$ (for
randomized rules, see the supplement).
\end{theorem}
Thus token-local criteria (drift, similarity, attention entropy) cannot replace knowing which part a token serves, and the parts must be found.

\begin{theorem}[Transport keeps weak parts]\leavevmode
\label{thm:mass}
(i) Unregularized balanced OT equals argmax with per-part offsets $\varphi$, $P^\star_{ki}>0\Rightarrow k\in\argmin_l(C_{li}-\varphi_l)$, and gives every part
its exact mass $\alpha_k$, whereas argmax assigns no token at all to a part that is out-attended on every token. (ii) Entropic unbalanced OT with row penalty
$\rho_1$ makes the part masses far less sensitive to attention competition: it enters the log mass ratio of two parts with weight
$1/(\rho_1+\zeta)$ instead of softmax's $1/\zeta$, about $9\times$ less with our constants (bound: supplement; accuracy: \cref{tab:supp_latency}).
\vspace{-4pt}
\end{theorem}

%% file: sec/5_experiments.tex
\section{Experiments and Results}
\label{sec:exp}

\begin{figure*}[t]
\centering
\setlength{\tabcolsep}{1pt}
\renewcommand{\arraystretch}{0.5}
\newlength{\qw}\setlength{\qw}{0.1075\textwidth}
\newcommand{\qhd}[2]{\tikz\node[fill=#1,rounded corners=2pt,minimum width=\qw,minimum height=3.6mm,inner xsep=0pt,inner ysep=0.5pt,font=\scriptsize]{#2};}
\newcommand{\qimg}[3]{\begin{tikzpicture}\begin{scope}\clip[rounded corners=2pt](0,0)rectangle(\qw,\qw);
\node[inner sep=0,anchor=south west] at (0,0) {\includegraphics[width=\qw]{fig/qual_156_#1.jpg}};\end{scope}
\foreach \xa/\ya/\xb/\yb/\c in {#2} {\draw[\c,line width=1pt,rounded corners=1pt] (\xa\qw,\ya\qw) rectangle (\xb\qw,\yb\qw);}#3
\end{tikzpicture}}
\newcommand{\qzm}[4]{\begin{tikzpicture}\begin{scope}\clip[rounded corners=2pt](0,0)rectangle(\qw,0.385\qw);
\node[inner sep=0,anchor=south west] at (0,0) {\includegraphics[width=\qw]{fig/qual_156_#1_zoom.jpg}};\end{scope}
\if\relax\detokenize{#2}\relax\else\node[anchor=south east,fill=white,fill opacity=0.85,text opacity=1,rounded corners=1.5pt,inner sep=1pt,
  font=\tiny\bfseries,text=#3] at (\qw-1.5pt,1.5pt) {#2};\fi#4
\end{tikzpicture}}
\begin{tabular}{@{}ccccccccc@{}}
\qhd{gray!18}{\textbf{Input}} & \qhd{gray!18}{Vanilla 1.00$\times$} & \qhd{gray!18}{Step skip 1.84$\times$} & \qhd{gray!18}{RegionE 1.30$\times$} &
\qhd{gray!18}{TeaCache 1.77$\times$} & \qhd{gray!18}{RAS 1.65$\times$} & \qhd{gray!18}{SpotEdit 1.54$\times$} & \qhd{gray!18}{TaylorSeer 1.79$\times$} &
\qhd{cvprblue!22}{\textbf{\color{cvprblue}\method\ 2.20$\times$}} \\[2pt]
\qimg{}{0.38/0.625/0.64/0.725/white}{} & \qimg{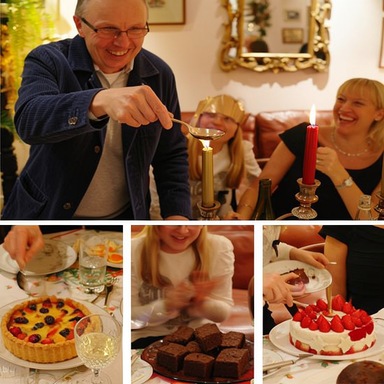}{0.38/0.625/0.64/0.725/red!85!black}{} & \qimg{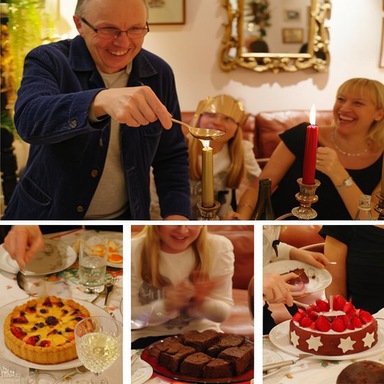}{0.38/0.625/0.64/0.725/red!85!black}{} &
\qimg{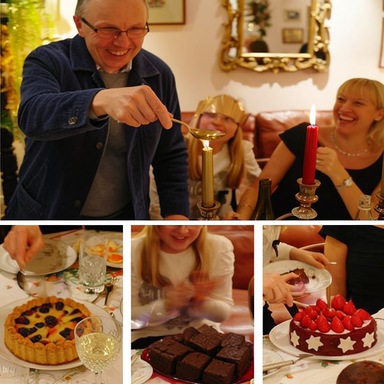}{0.38/0.625/0.64/0.725/red!85!black}{} & \qimg{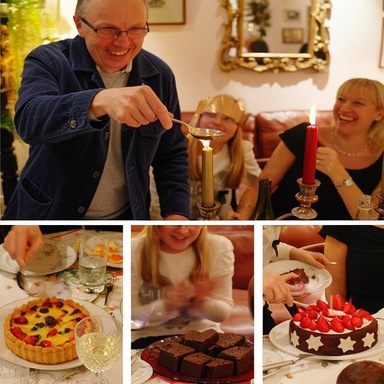}{0.38/0.625/0.64/0.725/red!85!black}{} & \qimg{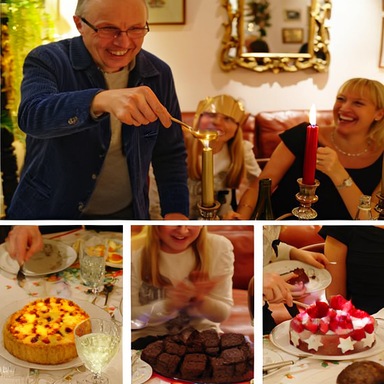}{0.38/0.625/0.64/0.725/red!85!black}{} &
\qimg{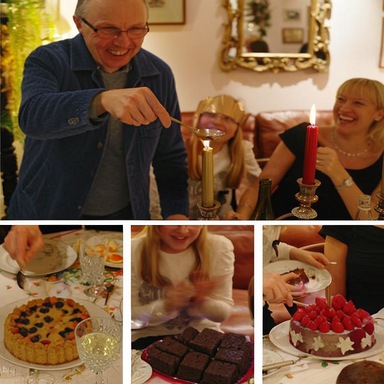}{0.38/0.625/0.64/0.725/red!85!black}{} & \qimg{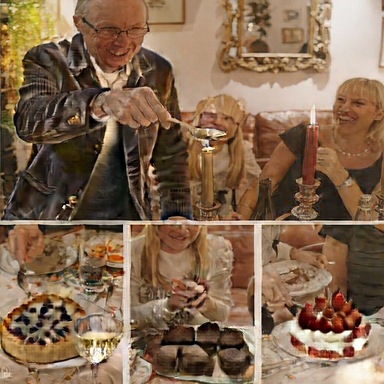}{0.38/0.625/0.64/0.725/red!85!black, 0.01/0.44/0.37/0.90/red!85!black}{} &
\qimg{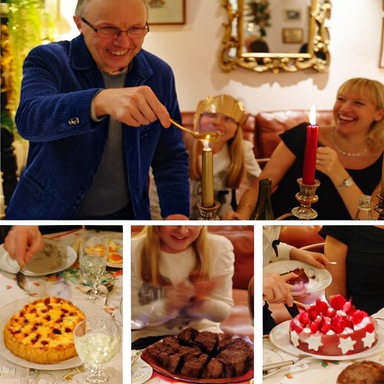}{0.38/0.625/0.64/0.725/green!55!black}{\draw[green!55!black!75,line width=0.8pt,rounded corners=2pt] (0.4pt,0.4pt) rectangle (\qw-0.4pt,\qw-0.4pt);} \\[1pt]
\qzm{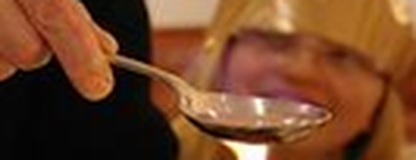}{}{}{} & \qzm{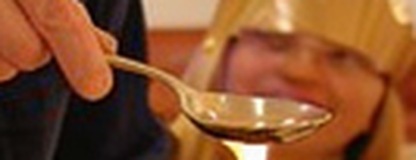}{5/6}{red!85!black}{} & \qzm{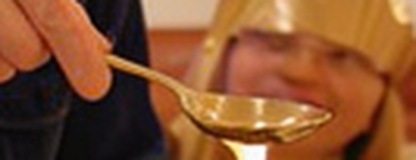}{5/6}{red!85!black}{} & \qzm{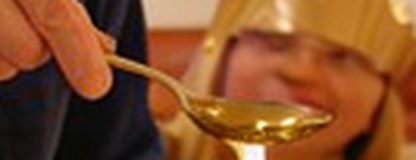}{5/6}{red!85!black}{} & \qzm{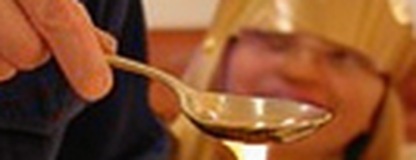}{5/6}{red!85!black}{} &
\qzm{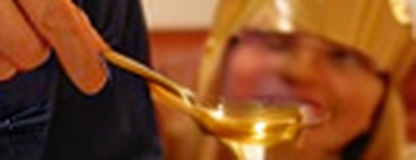}{5/6}{red!85!black}{} & \qzm{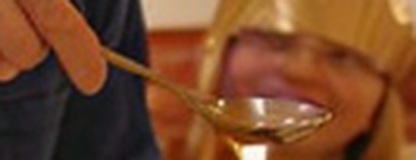}{5/6}{red!85!black}{} & \qzm{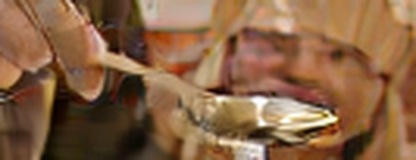}{4/6}{red!85!black}{} &
\qzm{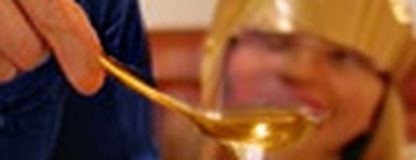}{6/6}{cvprblue}{\draw[green!55!black!75,line width=0.8pt,rounded corners=2pt] (0.4pt,0.4pt) rectangle (\qw-0.4pt,0.385\qw-0.4pt);} \\
\end{tabular}
\vspace{-0.6em}
\caption{\textbf{Qualitative comparison} (MICE-Bench \#156, Klein 4B, 4 steps). On this image only \method completes all six edits, at the highest
speed (\cref{tab:main}); every other method, vanilla included, loses the long golden soup spoon (red boxes, zoomed below; corner: edits done).}
\label{fig:qualitative}
\end{figure*}

\begin{table*}[t]
\centering
\tabsize
\renewcommand{\arraystretch}{0.9}
\resizebox{\linewidth}{!}{\begin{tabular}{@{}lrrrrrrrrrrrrrrrrrr@{}}
\toprule
 & \multicolumn{3}{c}{MICE} & \multicolumn{3}{c}{LoMOE} & \multicolumn{3}{c}{MIRA} & \multicolumn{3}{c}{InfoDet} & \multicolumn{3}{c}{Crello} & \multicolumn{3}{c}{GEdit} \\
\cmidrule(lr){2-4}\cmidrule(lr){5-7}\cmidrule(lr){8-10}\cmidrule(lr){11-13}\cmidrule(lr){14-16}\cmidrule(lr){17-19}
Method & Spd$\uparrow$ & Drop$\downarrow$ & LP$\downarrow$ & Spd$\uparrow$ & Drop$\downarrow$ & LP$\downarrow$ & Spd$\uparrow$ & Drop$\downarrow$ & LP$\downarrow$ & Spd$\uparrow$ & Drop$\downarrow$ & LP$\downarrow$ & Spd$\uparrow$ & Drop$\downarrow$ & LP$\downarrow$ & Spd$\uparrow$ & Drop$\downarrow$ & LP$\downarrow$ \\
\midrule
Vanilla & 1.00$\times$ & -- & 0 & 1.00$\times$ & -- & 0 & 1.00$\times$ & -- & 0 & 1.00$\times$ & -- & 0 & 1.00$\times$ & -- & 0 & 1.00$\times$ & -- & 0 \\
\midrule
\multicolumn{19}{@{}l}{\emph{FLUX.2 Klein Base 4B}} \\
\midrule
Step skip & 3.90$\times$ & 85 & 0.154 & 3.87$\times$ & 7 & 0.101 & 3.84$\times$ & 21 & 0.068 & 3.89$\times$ & 10 & 0.034 & 3.85$\times$ & 20 & 0.037 & 3.87$\times$ & 6 & 0.098 \\
RegionE~\cite{regione} & 1.48$\times$ & 235 & 0.214 & 1.61$\times$ & 23 & 0.166 & 1.66$\times$ & 64 & 0.160 & 1.91$\times$ & 14 & 0.052 & 1.90$\times$ & 37 & 0.052 & 1.57$\times$ & 8 & 0.139 \\
TaylorSeer~\cite{taylorseer} & 3.65$\times$ & 170 & 0.229 & 3.67$\times$ & 14 & 0.175 & 3.67$\times$ & 38 & 0.094 & 3.72$\times$ & 9 & 0.054 & 3.68$\times$ & 25 & 0.100 & 3.67$\times$ & 7 & 0.153 \\
TeaCache~\cite{teacache} & 4.50$\times$ & 100 & 0.202 & 4.44$\times$ & 9 & \textbf{0.124} & 4.46$\times$ & 28 & \textbf{0.087} & 4.48$\times$ & 8 & 0.034 & 4.47$\times$ & 25 & \textbf{0.039} & 4.46$\times$ & \textbf{6} & 0.115 \\
RAS~\cite{ras} & 1.76$\times$ & 147 & 0.216 & 1.76$\times$ & 12 & 0.147 & 1.76$\times$ & \textbf{22} & 0.102 & 1.76$\times$ & 7 & 0.049 & 1.77$\times$ & \textbf{21} & 0.055 & 1.75$\times$ & 8 & 0.131 \\
SpotEdit~\cite{spotedit} & 2.56$\times$ & 744 & 0.548 & 2.96$\times$ & 84 & 0.428 & 3.94$\times$ & 293 & 0.272 & 4.51$\times$ & 22 & 0.064 & 4.46$\times$ & 57 & 0.087 & 3.11$\times$ & 45 & 0.368 \\
MICE$^\dagger$~\cite{mice} & 0.71$\times$ & 626 & 0.529 & 0.91$\times$ & 51 & 0.395 & 0.82$\times$ & 147 & 0.319 & 0.82$\times$ & 56 & 0.261 & 0.91$\times$ & 83 & 0.295 & \multicolumn{3}{c}{n/a (no masks)} \\
MIRAGE$^\ddagger$~\cite{mirage} & 0.53$\times$ & 256 & 0.371 & 0.83$\times$ & 24 & 0.245 & 0.58$\times$ & 41 & 0.178 & 0.54$\times$ & 18 & 0.059 & 0.75$\times$ & 33 & 0.063 & 1.03$\times$ & 13 & 0.167 \\
\midrule
\textbf{TIDE (ours)} & \textbf{4.85$\times$} & \textbf{76} & \textbf{0.181} & \textbf{4.89$\times$} & \textbf{4} & 0.187 & \textbf{4.82$\times$} & 25 & 0.109 & \textbf{4.85$\times$} & \textbf{6} & \textbf{0.032} & \textbf{4.83$\times$} & 22 & 0.044 & \textbf{4.81$\times$} & 11 & \textbf{0.114} \\
\midrule
\multicolumn{19}{@{}l}{\emph{Qwen-Image-Edit-2511~\cite{qwenimage}}} \\
\midrule
Step skip & 3.20$\times$ & 63 & 0.141 & 3.17$\times$ & 3 & 0.104 & 3.22$\times$ & 5 & 0.028 & 3.19$\times$ & 16 & 0.039 & 3.18$\times$ & 18 & 0.025 & 3.19$\times$ & 4 & 0.115 \\
RegionE~\cite{regione} & 1.53$\times$ & 187 & 0.176 & 1.64$\times$ & 16 & 0.144 & 1.70$\times$ & 42 & 0.129 & 1.84$\times$ & 15 & 0.047 & 1.81$\times$ & 31 & 0.045 & 1.58$\times$ & 7 & 0.129 \\
TaylorSeer~\cite{taylorseer} & 3.08$\times$ & 142 & 0.198 & 3.04$\times$ & 11 & 0.157 & 3.11$\times$ & 27 & 0.086 & 3.09$\times$ & 13 & 0.049 & 3.03$\times$ & 22 & 0.084 & 3.06$\times$ & 6 & 0.139 \\
TeaCache~\cite{teacache} & 3.34$\times$ & 78 & 0.138 & 3.30$\times$ & 5 & \textbf{0.098} & 3.33$\times$ & 14 & 0.054 & 3.36$\times$ & 9 & \textbf{0.031} & 3.32$\times$ & 17 & 0.032 & 3.31$\times$ & 5 & 0.104 \\
RAS~\cite{ras} & 1.81$\times$ & 114 & 0.184 & 1.83$\times$ & 8 & 0.126 & 1.79$\times$ & 16 & 0.081 & 1.82$\times$ & 11 & 0.042 & 1.84$\times$ & 18 & 0.048 & 1.80$\times$ & 6 & 0.121 \\
SpotEdit~\cite{spotedit} & 2.28$\times$ & 546 & 0.482 & 2.65$\times$ & 64 & 0.375 & 2.84$\times$ & 188 & 0.238 & 3.18$\times$ & 21 & 0.058 & 3.14$\times$ & 47 & 0.076 & 2.76$\times$ & 36 & 0.324 \\
MICE$^\dagger$~\cite{mice} & 0.71$\times$ & 490 & 0.445 & 0.86$\times$ & 39 & 0.342 & 0.79$\times$ & 104 & 0.273 & 0.77$\times$ & 46 & 0.218 & 0.85$\times$ & 68 & 0.252 & \multicolumn{3}{c}{n/a (no masks)} \\
MIRAGE$^\ddagger$~\cite{mirage} & 0.74$\times$ & 215 & 0.312 & 1.01$\times$ & 18 & 0.208 & 0.78$\times$ & 29 & 0.146 & 0.73$\times$ & 14 & 0.053 & 0.94$\times$ & 25 & 0.054 & 1.18$\times$ & 10 & 0.151 \\
\midrule
\textbf{TIDE (ours)} & \textbf{3.48$\times$} & \textbf{59} & \textbf{0.112} & \textbf{3.46$\times$} & \textbf{2} & 0.118 & \textbf{3.52$\times$} & \textbf{4} & \textbf{0.042} & \textbf{3.50$\times$} & \textbf{8} & 0.044 & \textbf{3.47$\times$} & \textbf{16} & \textbf{0.030} & \textbf{3.49$\times$} & \textbf{3} & \textbf{0.103} \\
\bottomrule
\end{tabular}}
\caption{\textbf{Many-step editors:} Klein Base 4B (28 steps, guidance 4.0) and, bottom, Qwen-Image-Edit-2511 (50 steps, true CFG 4.0; vanilla as above; \method: Klein Base setting), 512$^2$, six benchmarks. Marks and bold as in \cref{tab:main}.}
\label{tab:base}
\end{table*}

\begin{table}[t]
\centering
\tabsize
\resizebox{\linewidth}{!}{\begin{tabular}{@{}lrrrr@{}}
\toprule
Variant & Spd$\uparrow$ & Done\%$\uparrow$ & Drop$\downarrow$ & LPIPS$\downarrow$ \\
\midrule
\multicolumn{5}{@{}l}{\emph{(a) Regions and scheduling rule (8 steps, $\tau{=}0.6$)}} \\
One global threshold ($\tau{=}0.72$) & 3.46$\times$ & 61.0 & 384 & 0.320 \\
Per-part, ground-truth masks & 3.41$\times$ & 69.0 & 226 & 0.299 \\
Per-part, argmax regions & 3.44$\times$ & 70.3 & 206 & 0.292 \\
\textbf{Per-part, OT regions (ours)} & \textbf{3.51$\times$} & \textbf{70.9} & \textbf{200} & \textbf{0.291} \\
\midrule
\multicolumn{5}{@{}l}{\emph{(b) When to decide (4 steps, $\tau{=}1.0$)}} \\
Decide once after step 1 (strict one-pass) & 2.10$\times$ & \textbf{72.8} & \textbf{146} & 0.226 \\
Background frozen to the source & 2.18$\times$ & 70.4 & 182 & 0.240 \\
\textbf{Re-check every step, freeze-only (ours)} & \textbf{2.20$\times$} & 72.1 & 155 & \textbf{0.222} \\

\bottomrule
\end{tabular}}
\caption{\textbf{Ablations} on MICE-Bench (262 images; VLM judge, Drop relative to vanilla with the same number of steps). Bold: best per block.}
\label{tab:ablation}
\end{table}

\begin{table}[t]
\centering
\tabsize
\resizebox{\linewidth}{!}{\begin{tabular}{@{}lrrrrrr@{}}
\toprule
Qwen-Image-Edit-2511 & MICE & LoMOE & MIRA & InfoDet & Crello & GEdit \\
\midrule
Vanilla, other seed (floor) & 28--32 & 4--6 & 9 & 12 & 19 & 4 \\
\midrule
Step skip & 25 & 3 & 5 & \textcolor{red!75!black}{16} & 18 & 4 \\
TeaCache & 31 & 5 & \textcolor{red!75!black}{14} & 9 & 17 & 5 \\
\textbf{\method} & \textbf{23} & \textbf{2} & \textbf{4} & \textbf{8} & \textbf{16} & \textbf{3} \\
\bottomrule
\end{tabular}}
\caption{\textbf{Seed noise floor} (Qwen-Image-Edit-2511; MICE: first 100 images). Floor: edits vanilla completes and loses when re-run
with another seed; below: each method's Drop. Red: above the floor. Only \method stays below it everywhere.}

\label{tab:floor}
\end{table}

\begin{table}[t]
\centering
\tabsize
\resizebox{\linewidth}{!}{\begin{tabular}{@{}lrrrrr@{}}
\toprule
 & \multicolumn{2}{c}{Klein 4B} & \multicolumn{2}{c}{Klein Base} & Qwen \\
\cmidrule(lr){2-3}\cmidrule(lr){4-5}\cmidrule(l){6-6}
Method & TFLOPs$\downarrow$ & A100$\uparrow$ & TFLOPs$\downarrow$ & A100$\uparrow$ & A100$\uparrow$ \\
\midrule
Vanilla & 71.0 & 1.00$\times$ & 993.7 & 1.00$\times$ & 1.00$\times$ \\
Step skip & 35.5 & 1.73$\times$ & 248.4 & 3.83$\times$ & 3.20$\times$ \\
TeaCache & 35.5 & 1.63$\times$ & 213.1 & 4.39$\times$ & 3.34$\times$ \\
\textbf{\method} & \textbf{27.1} & \textbf{1.74$\times$} & \textbf{186.4} & \textbf{4.69$\times$} & \textbf{3.48$\times$} \\
\bottomrule
\end{tabular}}
\caption{\textbf{Compute and a second GPU} (MICE-Bench). TFLOPs: transformer per image (PyTorch FLOP counter, first 5 images);
A100: end-to-end speed-up over vanilla on the same GPU. Klein 4B: \method-fast; step skip: 2 of 4 / 7 of 28 steps.}
\label{tab:compute}
\end{table}

\subsection{Experimental Setup}
\vspace{-4pt}
\label{sec:setup}

\runin{Editors and hardware} FLUX.2 Klein 4B~\cite{flux2}, a distilled in-context flow transformer, edits in 4 steps (its default;
the unaccelerated editor, \emph{vanilla}, takes about 21\,s per image); the undistilled Klein Base 4B uses 28 steps and guidance 4.0, and
Qwen-Image-Edit-2511~\cite{qwenimage}, a second model family, 50 steps and true CFG 4.0 (\cref{sec:base}). All runs use bf16, 512$\times$512
images and seed 0, Klein on GTX 1080 Ti GPUs (11\,GB, no bf16 hardware; A100: \cref{sec:latency}) and Qwen on one A100. Speed is the end-to-end
wall-clock ratio to vanilla timed in the same run (text encoding, all steps, region finding, VAE decoding).

\runin{Operating points} On Klein 4B, \textbf{\method-fast} uses $\tau{=}1.0$; \textbf{\method-quality} also computes the final step
in full, where $h/t{=}1$ and freezing errors weigh most, and freezes nothing for global edits (supplement). Both were fixed on 12 MICE-Bench and 22
GEdit images; $\tau{=}1.5$ is better on the test set (\cref{sec:main}), so \method-fast is conservative. On Klein Base, $s{=}4$, $R{=}8$,
$\tau{=}1.5$ were fixed on at most 20 MICE-Bench images and reused unchanged on Qwen.

\runin{Benchmarks} MICE-Bench~\cite{mice} (262 public images, 2{,}095 edits), LoMOE-Bench~\cite{lomoe} (64 images, 3 edits each),
MIRA-Bench~\cite{mirage} (100 images, 5 edits each), the InfoDet and Crello text-editing sets of InfoEdBench~\cite{infoedbench} (first 100 images
with at least two editable text boxes) and GEdit-Bench~\cite{step1x} (one edit each). Masks are used only for evaluation and mask-based baselines.

\runin{Metrics} \emph{Drop} counts edits that vanilla completes and the method loses: a VLM judge (Qwen3-VL-8B~\cite{qwen3vl})
answers one yes/no question per instruction part (done if $P(\text{yes})>0.5$), with the same prompts and vanilla reference for every method;
paired tests in the supplement. \emph{LPIPS}~\cite{lpips} to the vanilla output measures closeness to the editor.

\runin{Baselines} the accelerators TeaCache~\cite{teacache}, TaylorSeer~\cite{taylorseer}, RAS~\cite{ras}, RegionE~\cite{regione} and
SpotEdit~\cite{spotedit}, and the multi-instance editors MICE~\cite{mice} (benchmark masks) and MIRAGE~\cite{mirage} (time includes its VLM
grounding), on the same editor, images, seed, resolution and GPUs. On 4 steps they keep their decision rules with proportionally scaled
schedules; on Klein Base, their official settings. \emph{Step skip} runs fewer steps (2 of 4; 7 of 28 on Klein Base). Details: supplement.
\vspace{-7pt}
\subsection{Per-instruction versus One Global Threshold}
\label{sec:claim}
We compare \method with the same engine and regions but one global threshold on all tokens (as in RegionE-style accelerators), sweeping $\tau$
for both on all 262 MICE-Bench images at 4 and 8 steps (\cref{tab:claim}, \cref{fig:pareto}a). At matched speed, \method drops 48\% fewer edits
at 3.5$\times$ (200 vs.\ 384; paired 95\% CI of the difference $[137,233]$, McNemar $p<10^{-36}$) and 37\% fewer at about 4$\times$, while running
faster; at the default 4 steps it drops 35\% fewer at 2.2$\times$ (155 vs.\ 237; CI $[51,115]$, $p<10^{-12}$; net of the 120 drops of the shared
caches alone, 35 vs.\ 117), and fewer at two further seeds (183 vs.\ 273, 170 vs.\ 289, $p<10^{-13}$). On LoMOE, MIRA, InfoDet and Crello (8 steps)
it drops 42--77\% fewer at about 4$\times$ ($p{\le}0.031$). The rules coincide (within 3\%) at the extremes, where almost nothing or everything is
frozen, and differ most in between, where a global threshold already freezes the subtle parts but not yet the large edits (\cref{thm:price}).
Keeping edits does not cost image quality: at 3.5$\times$ \method is also closer to vanilla (LPIPS 0.291 vs.\ 0.320) and better on the
worst-edited object (CLIP 21.94 vs.\ 21.17; supplement).

\runin{Which edits are dropped}
At 8 steps it loses edits of small objects about twice as often as large ones (32.9 vs.\ 13.6\%, \cref{tab:supp_size}), as \cref{thm:price}
predicts; per-part budgets about halve these losses.

\runin{Premises on real data}\label{sec:premise}
Replaying vanilla trajectories with oracle breaking scores and budgets of 5\% of each edit's size (supplement, premise check), at a 60\% target saving
a global threshold breaks at least one edit in 89\% of MICE-Bench images at 4 steps, while per-part thresholds save 9.6 points more without breaking any.

\runin{Human check}
At 2.15$\times$ on Klein 4B the judge finds 108 edits lost only by one global threshold and 26 only
by \method (supplement, paired tests). On the 103 and 25 whose masks match their parts, plus 30 controls, 33 raters compared both
outputs side by side, blind, with a zoom on the edited object (16--17 votes per edit; supplement). For the rater majority the global threshold loses 49 of these edits and \method 15 (one rule only: 47 vs.\ 13,
$p{=}1.2{\times}10^{-5}$), and 11 vs.\ 4 of the objects under 1\%, too few for a separate test (Fleiss' $\kappa$ 0.74 on the global threshold's outputs, 0.58 on \method's; supplement). Raters confirm a
third of the judge's ``not done'' answers, so Drop overcounts absolute losses, but the gap between the rules holds.

\subsection{Comparison with Accelerators}
\label{sec:main}

\Cref{tab:main} compares all methods on Klein 4B. \method-fast is the fastest on every benchmark (2.20--2.24$\times$; stage times in \cref{tab:supp_latency}).
At similar speed it loses far fewer edits than TaylorSeer on MICE, LoMOE and GEdit, whose two-step forecasts are poor. TeaCache, which skips whole steps, drops fewer
edits than \method-quality at similar speed (91 vs.\ 121, $p{=}0.004$), and step skip (1.84$\times$) drops only 57. Most of \method-fast's losses
come from reused keys and values (cached text and reference alone: 120 drops at 1.71$\times$), not from freezing (121 vs.\ 120 with refreshed
context); it pays for its speed in no-reference quality and text accuracy (supplement).

In \cref{fig:pareto}b, between step skip's 1.84$\times$ and a single step (3.10$\times$, 238 drops), token freezing on 2 steps holds the
frontier (2.43$\times$, 130 drops; $p{<}10^{-18}$ vs.\ one step), and on 4 steps per-part budgets keep far more edits than one global
threshold ($\tau{=}1.5$: 140 vs.\ 218 at 2.29 vs.\ 2.24$\times$, $p{<}10^{-11}$). Within the seed noise floor (96--101 drops) nothing beats step skip on MICE-Bench. \Cref{fig:qualitative} shows an example.

\subsection{Generalization to Many-step Editors}
\label{sec:base}
On the undistilled Klein Base (\cref{tab:base}; one \method setting with adaptive wake-up, \cref{sec:freeze}), \method is the fastest method on all six benchmarks (4.81--4.89$\times$, baselines at most 4.51$\times$), and drops fewer edits than TeaCache on five of six (all but GEdit);
on MICE-Bench it drops 76 against TeaCache's 100 and is also closer to vanilla in LPIPS (0.181 vs.\ 0.202). \method's losses stay below the seed noise floor (edits vanilla itself loses when re-run with another seed) on five of six benchmarks (MICE: 46 vs.\ 53--55, first 100 images; supplement);
with 14 steps it reaches 5.23$\times$ within this noise floor (39 vs.\ TeaCache's 35 at 4.50$\times$, $p{=}0.62$; \cref{fig:pareto}c). On
Qwen-Image-Edit-2511 (\cref{tab:base}, bottom), with the same setting, \method is again the fastest on every benchmark (3.46--3.52$\times$,
baselines at most 3.36$\times$) and drops the fewest edits on all six, significantly fewer than TeaCache on MICE (59 vs.\ 78, $p{=}0.011$) and
MIRA (4 vs.\ 14, $p{=}0.013$); over three seeds it drops 22.7$\pm$1.5 edits against 25.3$\pm$1.5 for step skip and 31.3$\pm$2.5 for TeaCache
(MICE, 100 images). It is the only accelerator below the seed noise floor on all six benchmarks (\cref{tab:floor}), and a second
judge agrees (14 vs.\ 16 and 19 drops; supplement, Sec.~F).

\subsection{Ablation Studies}
\label{sec:ablation}
We vary one component at a time on MICE-Bench (\cref{tab:ablation}). \emph{(a) Regions.} Our self-grounded regions match or beat ground-truth
masks and argmax regions (200 vs.\ 226 and 206 drops, faster); they recover the number of parts in 78\% of images, and with a poor region \method
still drops fewer edits than one global threshold (Prop.~C; supplement). \emph{(b) When to decide.} Re-checking at every step is faster than
deciding once (2.20 vs.\ 2.10$\times$) at nine more drops; at 4 steps only 1--7\% of first-frozen tokens would drift back (18\% at 8 steps, hence wake-up).

\subsection{Compute and Hardware}
\label{sec:latency}
Independent of the GPU, \method's transformer needs 2.6$\times$ (Klein 4B) and 5.3$\times$ (Klein Base) fewer TFLOPs than vanilla, the
largest saving we measure, and on an A100 it stays the fastest method on all three editors, on Klein 4B only just (\cref{tab:compute}).

\subsection{Judge Reliability and Image Quality}
\label{sec:human}
\emph{Judge and quality.} On 167 edits rated by 33 raters (supplement) the judge agrees with the majority on 88\% (AC1 0.86) and is the
stricter of the two; MOS is near the ceiling for all three groups (4.74--4.87, 12 different images each), so this small study does not rank their quality (supplement). A second judge (Idefics3-8B~\cite{idefics3}) ranks the rules the same way (96 vs.\ 79 drops,
$p{=}0.02$; VIEScore: supplement).

%% file: sec/6_conclusion.tex
\section{Conclusion}
\label{sec:conclusion}
Giving every part of the instruction its own compute budget, found from the editor's own attention, matters: one global threshold breaks at
least one edit in 89\% of multi-edit images, per-instruction budgets drop up to 77\% fewer edits at matched or higher speed on Klein 4B (48\% on MICE-Bench), a gap
raters confirm, and on both many-step editors (Klein Base, Qwen-Image-Edit) the method is the fastest compared method, with no masks or training. On the 4-step editor,
step reduction is the better accelerator up to 1.84$\times$, and \method costs no-reference image quality and text accuracy (supplement); it pays off on the many-step editors.

%% file: sec/7_llm_statement.tex
\section*{LLM Usage Statement}
\vspace{-4PT}
\emph{Research.} Qwen3-VL-8B-Instruct~\cite{qwen3vl} is the VLM judge for every edit-completion result (\cref{sec:setup}) and the VIEScore scorer;
Idefics3-8B~\cite{idefics3} is the second judge. In the supplement, Qwen3-VL-8B also splits instructions in one robustness experiment (text
only), writes candidate judge prompts in the prompt-refinement check, and serves as the localizer of the MIRAGE baseline (its official option).
\emph{Writing and code.} An AI assistant (Claude, Anthropic) was used to  debug experiment and analysis scripts and to edit the text for clarity and consistency. All experiments were run, and all results, claims and
text were checked, by the authors.

%% file: sec/X_suppl.tex
\clearpage
\maketitlesupplementary
\appendix
\raggedbottom %
\renewcommand{\topfraction}{0.95}\renewcommand{\dbltopfraction}{0.95}\renewcommand{\bottomfraction}{0.8}
\renewcommand{\textfraction}{0.05}\renewcommand{\floatpagefraction}{0.8}\renewcommand{\dblfloatpagefraction}{0.8}
\setcounter{topnumber}{5}\setcounter{dbltopnumber}{4}\setcounter{bottomnumber}{3}\setcounter{totalnumber}{10}
\makeatletter
\setlength\@fptop{0pt}\setlength\@fpsep{12pt plus 2pt}\setlength\@fpbot{0pt plus 1fil}
\setlength\@dblfptop{0pt}\setlength\@dblfpsep{12pt plus 2pt}\setlength\@dblfpbot{0pt plus 1fil}
\makeatother
\setlength{\textfloatsep}{10pt plus 2pt minus 3pt}\setlength{\dbltextfloatsep}{10pt plus 2pt minus 3pt}
\setlength{\floatsep}{8pt plus 2pt minus 2pt}\setlength{\dblfloatsep}{8pt plus 2pt minus 2pt}
\newcommand{\thmtitle}{}
\newtheorem*{restatedthm}{\thmtitle}
\newenvironment{restated}[1]{\renewcommand{\thmtitle}{#1}\begin{restatedthm}}{\end{restatedthm}}
\noindent This supplement contains the proofs of the results in the main paper (\cref{sec:proofs}), the adaptive-freezing view and its
one-pass special case (\cref{sec:adaptive}), method and implementation details (\cref{sec:details}), the evaluation protocol including the
VLM judge and the human study (\cref{sec:protocol}), the settings of every baseline (\cref{sec:baselines}), additional quantitative results
(\cref{sec:moreresults}), additional qualitative results and failure cases (\cref{sec:qual}), and a discussion of limitations and broader
impact (\cref{sec:impact}). References to the main paper are written as ``main paper, Theorem~1''.

\section{Proofs and theory details}
\label{sec:proofs}

\subsection{Endpoint-freeze (main paper, Proposition~1)}\label{sec:supp_endpoint}

\begin{restated}{Proposition 1 (Exact straight-path freezing)}
Let token $i$ be frozen at time $t$ with endpoint $\hat x^\star_{0,i}$ and implied noise
$\tilde x_{1,i}=\big(x_{t,i}-(1-t)\hat x^\star_{0,i}\big)/t$. Euler steps with
$v_i=(x_{t,i}-\hat x^\star_{0,i})/t$ keep $x_{t',i}=(1-t')\hat x^\star_{0,i}+t'\tilde x_{1,i}$ for every later $t'$,
and the one-step deviation from the dense update is $(h/t)\,\|\hat x^\star_{0,i}-\hat x^\theta_{0,i}\|$.
\end{restated}

\begin{proof}
By the definition of the implied noise, the frozen state is
\[
x_{t,i}=(1-t)\,\hat x^\star_{0,i}+t\,\tilde x_{1,i},
\]
so the frozen velocity is a constant:
\[
v_i=\frac{x_{t,i}-\hat x^\star_{0,i}}{t}=\tilde x_{1,i}-\hat x^\star_{0,i}.
\]
One Euler step of size $h$ gives
\begin{align*}
x_{t-h,i}&=x_{t,i}-h\,(\tilde x_{1,i}-\hat x^\star_{0,i})\\
         &=\big(1-(t-h)\big)\,\hat x^\star_{0,i}+(t-h)\,\tilde x_{1,i},
\end{align*}
which is the same form at $t'=t-h$; induction covers all later steps. The dense update uses
$v^\theta_i=(x_{t,i}-\hat x^\theta_{0,i})/t$, so the two updates differ by
\[
h\,\|v_i-v^\theta_i\|=\frac{h}{t}\,\|\hat x^\star_{0,i}-\hat x^\theta_{0,i}\|. \qedhere
\]
\end{proof}

\paragraph{Consequences.}
\begin{enumerate}[label=(\roman*),leftmargin=*,itemsep=1pt,topsep=2pt]
  \item \emph{No stored history.} A frozen token is fully described by its endpoint and the current state.
  \item \emph{Same as velocity reuse under Euler.} Endpoint-freeze equals reusing the last velocity, because $v_i$ is
        constant along the frozen path.
  \item \emph{The last step weighs most.} The per-step error is the endpoint drift weighted by $h_j/t_j$, and $h/t=1$ at
        the final step. This is why \method-quality recomputes the final step in full.
\end{enumerate}

\subsection{Setting of the budget results}

\paragraph{Items and policies.}
An item $u=(i,j)\in\mathcal I$ is token $i$ at step $j$. Freezing it saves compute $c_u>0$ and adds error $a_u\ge0$ to
its part $\kappa(u)\in\{0,\dots,K\}$ ($0$ = background); its efficiency score is $r_u=a_u/c_u$. A policy
$z\in\{0,1\}^{\mathcal I}$ freezes $u$ iff $z_u=1$. Its error on part $k$ and its saving are
\[
E_k(z)=\sum_{\kappa(u)=k}a_u z_u,\qquad S(z)=\sum_u c_u z_u .
\]
Part $k$ has a budget $\varepsilon_k\ge0$. It is \emph{satisfied} if $E_k(z)\le\varepsilon_k$, and \emph{dropped} if
$z_u=1$ for every $u$ with $\kappa(u)=k$.

\paragraph{Policy classes.}
We compare
\begin{align*}
z^\theta_u&=\mathbf 1[r_u\le\theta] && \text{(global threshold)},\\
z^\Theta_u&=\mathbf 1[r_u\le\theta_{\kappa(u)}] && \text{(per-part thresholds)}.
\end{align*}
A policy is \emph{part-agnostic} if its law depends only on item features $f_u$ that do not contain $\kappa(u)$, and is
invariant under every feature-preserving permutation of the items. A global threshold on drift or similarity is
part-agnostic, with $f_u=r_u$, when costs and weights are uniform.

\paragraph{Breaking score.}
Write $E_k(\theta)=E_k(z^\theta)$. The breaking score of part $k$ and its saving below a score $\beta$ are
\begin{align*}
\beta_k&=\min\{r_u:\ \kappa(u)=k,\ E_k(r_u)>\varepsilon_k\},\\
S_k^-(\beta)&=\sum_{\kappa(u)=k}c_u\,\mathbf 1[r_u<\beta],
\end{align*}
with $\min\emptyset=+\infty$. Since $\theta\mapsto E_k(\theta)$ is a non-decreasing, right-continuous step function that
jumps only at item scores,
\begin{equation}
E_k(\theta)\le\varepsilon_k\iff\theta<\beta_k .\tag{B}
\end{equation}

\subsection{Price of a global threshold (main paper, Theorem~1)}

\begin{restated}{Theorem 1 (Exact price of a global threshold, full statement)}
Let $\beta_{\min}=\min_k\beta_k$. The best feasible global threshold and the best feasible per-part policy save
\[
S_{\mathrm{glob}}=\sum_k S_k^-(\beta_{\min}),\qquad S_{\mathrm{pc}}=\sum_k S_k^-(\beta_k),
\]
so the gain of per-part thresholds is
\[
S_{\mathrm{pc}}-S_{\mathrm{glob}}=\sum_k\big[S_k^-(\beta_k)-S_k^-(\beta_{\min})\big]\ge0,
\]
strictly iff some part $k$ has an item with $\beta_{\min}\le r_u<\beta_k$. Per-part thresholds are within one tie
class per part of the optimal 0-1 policy: $\mathrm{OPT}-S_{\mathrm{pc}}\le\sum_k C_k$, where $C_k$ is the total cost of
the items of part $k$ with score exactly $\beta_k$. Finally, if $\theta_\rho$ is the smallest score reaching a target
saving $\rho$, the global threshold $z^{\theta_\rho}$ violates part $k$ iff $\theta_\rho\ge\beta_k$, and drops it iff
$\theta_\rho\ge\max_{\kappa(u)=k}r_u$.
\end{restated}

\begin{proof}
\emph{Global threshold.} By (B),
\[
z^\theta\ \text{feasible}\iff\theta<\beta_k\ \ \forall k\iff\theta<\beta_{\min}.
\]
Since $S(z^\theta)=\sum_k\sum_{\kappa(u)=k}c_u\mathbf 1[r_u\le\theta]$ is non-decreasing in $\theta$ and there are
finitely many item scores,
\[
S_{\mathrm{glob}}=\sup_{\theta<\beta_{\min}}S(z^\theta)=\sum_k S_k^-(\beta_{\min}).
\]

\emph{Per-part policy.} The constraints separate by part, $z^\Theta$ feasible $\iff\theta_k<\beta_k\ \forall k$, so
\[
S_{\mathrm{pc}}=\sum_k\ \sup_{\theta_k<\beta_k}\ \sum_{\kappa(u)=k}c_u\mathbf 1[r_u\le\theta_k]=\sum_k S_k^-(\beta_k).
\]

\emph{Gap.} For every $k$, since $\beta_{\min}\le\beta_k$,
\[
S_k^-(\beta_k)-S_k^-(\beta_{\min})=\sum_{\kappa(u)=k}c_u\,\mathbf 1[\beta_{\min}\le r_u<\beta_k]\ \ge0,
\]
with equality iff no such item exists, as $c_u>0$.

\emph{Distance to the optimum.} The constraints separate by part, so $\mathrm{OPT}=\sum_k\mathrm{OPT}_k$ with
\[
\mathrm{OPT}_k=\max_{z}\Big\{\sum_{\kappa(u)=k}c_u z_u\ :\ \sum_{\kappa(u)=k}a_u z_u\le\varepsilon_k\Big\}.
\]
The LP relaxation of this knapsack is solved greedily in increasing $r_u$~\cite{dantzig}. By (B) all items with
$r_u<\beta_k$ fit, and by the definition of $\beta_k$ the tie class $\{r_u=\beta_k\}$ does not fit entirely, so
\[
\mathrm{OPT}_k\le\mathrm{LP}_k\le S_k^-(\beta_k)+C_k .
\]
Summing over $k$ gives $\mathrm{OPT}-S_{\mathrm{pc}}\le\sum_k C_k$.

\emph{Violated and dropped parts.} By (B) at $\theta=\theta_\rho$, part $k$ is violated iff $\theta_\rho\ge\beta_k$, and
\[
\begin{aligned}
k\ \text{dropped}&\iff z^{\theta_\rho}_u=1\ \ \forall u:\kappa(u)=k\\
&\iff\max_{\kappa(u)=k}r_u\le\theta_\rho.
\end{aligned}\qedhere
\]
\end{proof}
\paragraph{Worked example.}
Let $c_u=1$ and $a_u=\delta_u$. Part $A$ (an object swap) has 90 items with $\delta=0.02$ and 10 with $\delta=0.3$, and
budget $\varepsilon_A=2$. Part $B$ (a subtle recolor) has 10 items with $\delta=0.015$ and budget $\varepsilon_B=0.1$.
Then
\begin{align*}
E_A(0.02)&=1.8\le2<4.8=E_A(0.3),\\
E_B(0.015)&=0.15>0.1,
\end{align*}
so $\beta_A=0.3$ and $\beta_B=\beta_{\min}=0.015$. By Theorem~1,
\begin{align*}
S_{\mathrm{glob}}&=S_A^-(0.015)+S_B^-(0.015)=0,\\
S_{\mathrm{pc}}&=S_A^-(0.3)+S_B^-(0.015)=90 .
\end{align*}
Every global threshold that saves anything has $\theta\ge0.015=\max_{u\in B}r_u$ and drops $B$ entirely, while
per-part thresholds save 82\% of the items and keep both parts.

\paragraph{Per-part drift thresholds as the per-part policy.}
With uniform costs and $a_u=g_j\delta_{i,j}$, the per-part policy freezes $u$ iff $\delta_{i,j}\le\theta_k/g_j$. Within
one step $j$, where $g_j$ is a constant, the rule of the main paper (Eq.~4),
\[
\delta_{i,j}<\tau\,m_k ,
\]
is therefore this policy with $\theta_k=\tau m_k g_j$: a price per part that scales with the part's own drift. It is a
calibrated instance of the Lagrangian solution of the per-part problem, not an estimate of the exact breaking scores,
which would require the dense trajectory. \Cref{tab:supp_e1} measures how far a global threshold is from the per-part
optimum on real trajectories.

\begin{restated}{Proposition D (When Rule (4) is the per-part optimum)}
Fix a sparse step $j$ with uniform costs, so $a_u=g_j\delta_i$ with $\delta_i=\delta_{i,j}$. Let part $k$ have $n_k$
tokens with mean drift $m_k>0$, write $\delta_i=m_k\xi_i$, and let every budget be the same fraction $c\in[0,1)$ of
the part's own motion:
\begin{align*}
G_k(x)&=\frac{1}{n_k}\sum_{k(i)=k}\xi_i\,\mathbf 1[\xi_i<x],\\
\varepsilon_k&=c\,g_j\!\sum_{k(i)=k}\delta_i .
\end{align*}
(i)~If $G_k=G$ for every $k$, the largest feasible per-part drift thresholds are $\theta_k=\tau m_k$ with the same
$\tau=\max\{x:G(x)\le c\}$ for every part; this is Rule (4).
(ii)~If $\sup_x|G_k(x)-G(x)|\le\eta_k$, Rule (4) with this $\tau$ satisfies
\[
E_k(\tau m_k)\le\varepsilon_k+\eta_k\,g_j\!\sum_{k(i)=k}\delta_i .
\]
\end{restated}

\begin{proof}
Since $\delta_i=m_k\xi_i$ and $\sum_{k(i)=k}\delta_i=m_kn_k$,
\begin{align*}
E_k(\theta)&=g_j\!\sum_{k(i)=k}\!\delta_i\,\mathbf 1[\delta_i<\theta]=g_jm_kn_k\,G_k(\theta/m_k),\\
\varepsilon_k&=c\,g_jm_kn_k,
\end{align*}
hence
\begin{equation*}
\theta\ \text{feasible for part }k\iff G_k(\theta/m_k)\le c. \tag{$*$}
\end{equation*}

\emph{Part (i).} $G$ is non-decreasing and left-continuous, and $G(x)=1>c$ for $x>\max_i\xi_i$, so
\[
\{x:G(x)\le c\}=(-\infty,\tau],\qquad\tau<\infty .
\]
With $G_k=G$, ($*$) gives $\theta_k=\tau m_k$ for every $k$.

\emph{Part (ii).} Using $G(\tau)\le c$,
\begin{align*}
E_k(\tau m_k)&=g_jm_kn_k\,G_k(\tau)\\
             &\le g_jm_kn_k\big(G(\tau)+\eta_k\big)\\
             &\le\varepsilon_k+\eta_k\,g_jm_kn_k. \qedhere
\end{align*}
\end{proof}

Rule (4) thus attains the per-part optimum under two assumptions: the parts' drifts share one shape up to scale, and the
scales $m_k$ measured at the first decision stay representative later (on Klein Base, the check steps re-measure them).
\Cref{tab:supp_rule} shows how far this holds: the price ranks the parts like their oracle breaking scores on the
4-step editor, less so at 8 steps, and not on Klein Base, where the parts' drift scales have not yet separated at the
first decision.

\subsection{Part-agnostic rules (main paper, Theorem~2)}

\begin{restated}{Theorem 2 (Rules blind to the parts cannot guarantee heterogeneous budgets)}
There are instances with two parts $A,B$ of $n_A,n_B$ items with identical features and cost $c$, on which a per-part
policy saves $n_Ac$ within budget, while every deterministic part-agnostic rule saves nothing or violates the budget
of $B$. A randomized part-agnostic rule that violates the budget of $B$ with probability at most $\gamma$ saves at most
$c\gamma N^2/n_B$ in expectation, with $N=n_A+n_B$.
\end{restated}

\begin{proof}
\emph{Instance.} Let all items share the same feature, cost $c$ and error $a>0$, with
\[
\varepsilon_A=n_A a,\qquad \varepsilon_B=a/2 .
\]
Freezing any item of $B$ gives $E_B\ge a>\varepsilon_B$. The per-part policy $z_u=\mathbf 1[\kappa(u)=A]$ gives
\[
E_A=n_Aa=\varepsilon_A,\qquad E_B=0,\qquad S=n_Ac .
\]

\emph{Deterministic rules.} All items have the same feature, so invariance under every permutation forces
$z_u\equiv z_0$:
\[
z\equiv0\ \Rightarrow\ S=0,\qquad z\equiv1\ \Rightarrow\ E_B=n_Ba>\varepsilon_B .
\]

\emph{Randomized rules.} Let $M$ be the number of frozen items. Given $M=m\ge1$, the frozen set is uniform over
$m$-subsets, and $B$ is satisfied iff this subset lies in $A$, so
\[
\Pr[B\text{ satisfied}\mid M{=}m]=\frac{\binom{n_A}{m}}{\binom{N}{m}}\le\Big(\frac{n_A}{N}\Big)^{m}\le\frac{n_A}{N}.
\]
Averaging over $M$,
\begin{align*}
\Pr[B\text{ satisfied}]&\le\Pr[M{=}0]+\tfrac{n_A}{N}\Pr[M{\ge}1]\\
                       &=1-\tfrac{n_B}{N}\Pr[M{\ge}1].
\end{align*}
Hence $\Pr[B\text{ violated}]\le\gamma\Rightarrow\Pr[M{\ge}1]\le\gamma N/n_B$, and
\[
\mathbb E[S]=c\,\mathbb E[M]\le cN\Pr[M{\ge}1]\le\frac{c\gamma N^2}{n_B}. \qedhere
\]
\end{proof}

\paragraph{Interpretation.}
Token-local criteria (drift, similarity, attention entropy) cannot replace knowing which instruction part a token
serves when parts have different tolerances. This is why \method infers the parts first.

\subsection{Robustness to attribution errors (Proposition~C)}

\begin{restated}{Proposition C (Robustness to attribution errors)}
If a per-part policy with inferred parts $\hat\kappa$ satisfies
$\hat E_k(z)=\sum_{\hat\kappa(u)=k}a_uz_u\le\varepsilon_k$, then for every true part
\[
E_k(z)\le\varepsilon_k+\Lambda_k,\qquad \Lambda_k=\sum_{\kappa(u)=k,\ \hat\kappa(u)\ne k}a_uz_u .
\]
\end{restated}

\begin{proof}
Split the items of part $k$ by their inferred part. Since $a_uz_u\ge0$,
\begin{align*}
E_k(z)&=\sum_{\kappa(u)=k,\ \hat\kappa(u)=k}a_uz_u+\Lambda_k\\
      &\le\hat E_k(z)+\Lambda_k\ \le\ \varepsilon_k+\Lambda_k. \qedhere
\end{align*}
\end{proof}

Mis-assigned tokens hurt only through the error they carry while frozen. Coarse regions are therefore tolerable as long
as the tokens of a subtle edit are not assigned to a part with a much larger drift scale. \Cref{tab:supp_regstrat}
(drop rate by region quality) and \cref{tab:supp_clip} (ground-truth vs.\ self-grounded regions) show this empirically.

\subsection{Transport keeps weak parts (main paper, Theorem~3)}

\paragraph{Setting.}
Tokens $i$ carry demand $b_i>0$, parts carry target mass $\alpha_k>0$, and $C\in\mathbb R^{K\times N}$ is the
part--token cost, with entropic regulariser $\zeta>0$ and Gibbs kernel $\Gamma=e^{-C/\zeta}$ (entrywise). Argmax assigns
token $i$ to $\argmin_k C_{ki}$; softmax gives responsibilities and masses
\[
\sigma_{ki}=\frac{e^{-C_{ki}/\zeta}}{\sum_l e^{-C_{li}/\zeta}},\qquad
\hat\alpha^{\mathrm{sm}}_k=\sum_i b_i\sigma_{ki}.
\]
Part $k$ is \emph{dominated with gap $g$} if
\[
\forall i\ \ \exists\,l\ne k:\quad C_{ki}\ge C_{li}+g .
\]

\begin{restated}{Theorem 3 (Transport keeps weak parts)}
(i)~Unregularized balanced OT equals argmax with per-part offsets $\varphi$,
\[
P^\star_{ki}>0\ \Rightarrow\ k\in\argmin_l\,(C_{li}-\varphi_l),
\]
and gives every part its exact mass $\alpha_k$, whereas argmax assigns no token to a dominated part.
(ii)~With row penalty $\rho_1$ and $\lambda=\rho_1/(\rho_1+\zeta)$, the masses $\hat\alpha$ of entropic unbalanced OT
satisfy
\[
\Big|\log\frac{\hat\alpha_k}{\hat\alpha_l}-\lambda\log\frac{\alpha_k}{\alpha_l}\Big|
\le\frac{\|C_k-C_l\|_\infty}{\rho_1+\zeta},
\]
whereas softmax gives a part dominated with gap $g$ a mass $\hat\alpha^{\mathrm{sm}}_k\le e^{-g/\zeta}\sum_i b_i$.
\end{restated}

\begin{proof}
\emph{Part (i).} With $\sum_k\alpha_k=\sum_i b_i$, the primal and dual LPs
\begin{align*}
&\min_{P\ge0}\ \langle C,P\rangle\ \ \text{s.t.}\ \ P\mathbf 1=\alpha,\ P^\top\mathbf 1=b,\\
&\max_{\varphi,\psi}\ \textstyle\sum_k\alpha_k\varphi_k+\sum_i b_i\psi_i\ \ \text{s.t.}\ \ \varphi_k+\psi_i\le C_{ki}
\end{align*}
are feasible and bounded, so strong duality holds. As $b_i>0$, a dual optimum has
$\psi_i=\min_l(C_{li}-\varphi_l)$, and complementary slackness gives
\[
P^\star_{ki}>0\ \Rightarrow\ C_{ki}-\varphi_k=\min_l\,(C_{li}-\varphi_l).
\]
Feasibility gives $\sum_iP^\star_{ki}=\alpha_k$ with $P^\star_{ki}\le b_i$, so part $k$ is supported on at least
$\lceil\alpha_k/\max_i b_i\rceil$ tokens. Under domination, $k\notin\argmin_l C_{li}$ for every $i$, so argmax gives
part $k$ no token.

\emph{Part (ii).} The scaling iterations keep $P=\mathrm{diag}(u)\,\Gamma\,\mathrm{diag}(v)$ with the row update
\[
u_k\leftarrow\big(\alpha_k/(\Gamma v)_k\big)^{\lambda},
\]
so after every row update (and at the fixed point)
\[
\hat\alpha_k=u_k(\Gamma v)_k=\alpha_k^{\lambda}\,(\Gamma v)_k^{1-\lambda},
\]
hence
\[
\log\frac{\hat\alpha_k}{\hat\alpha_l}-\lambda\log\frac{\alpha_k}{\alpha_l}
=(1-\lambda)\log\frac{(\Gamma v)_k}{(\Gamma v)_l}.
\]
Since $v>0$, $(\Gamma v)_k/(\Gamma v)_l$ is a weighted average of $e^{-(C_{ki}-C_{li})/\zeta}$ (weights
$\propto\Gamma_{li}v_i$), so
\[
\Big|\log\frac{(\Gamma v)_k}{(\Gamma v)_l}\Big|\le\frac{\|C_k-C_l\|_\infty}{\zeta},
\]
and $1-\lambda=\zeta/(\rho_1+\zeta)$ gives the bound. For softmax, domination gives for every $i$ some $l(i)\ne k$
with
\[
\sigma_{ki}=\sigma_{l(i)i}\,e^{-(C_{ki}-C_{l(i)i})/\zeta}\le e^{-g/\zeta},
\]
as $\sigma_{l(i)i}\le1$; weighting by $b_i$ and summing gives the claim.
\end{proof}

\paragraph{Two parts: comparative advantage.}
For $K=2$, let $\Delta_i=C_{Bi}-C_{Ai}$ and $s=\varphi_B-\varphi_A$, fixed by the mass of $B$. By part (i),
\[
\Delta_i<s\ \Rightarrow\ i\mapsto B,\qquad \Delta_i>s\ \Rightarrow\ i\mapsto A .
\]
A part thus receives the tokens where it is \emph{relatively} least out-attended, even if it is out-attended
everywhere: exactly the situation of a subtle edit next to a salient one.

\paragraph{Numbers.}
With $\zeta=0.12$ and $\rho_1=1$, competition enters the masses with sensitivity $1/(\rho_1+\zeta)\approx0.89$ instead
of $1/\zeta\approx8.3$ for softmax, a factor of about 9. On a synthetic instance where one part is dominated with gap
$g=0.3$ and has target mass $0.2$, it receives:

\begin{center}
\small
\begin{tabular}{@{}lcccc@{}}
\toprule
 & Argmax & Softmax & Unbal.\ OT & Bal.\ OT \\
\midrule
Mass & 0 & 0.029 & 0.164 & 0.200 \\
\bottomrule
\end{tabular}
\end{center}

These results concern the masses and supports of the parts. Whether a support lands on the right object depends on the
attention cost; it is measured (the region table of the main paper).
\section{Adaptive freezing and the one-pass special case}
\label{sec:adaptive}

\paragraph{Step weights.}
By Proposition~1, freezing a token at step $j$ changes its update by $h_j/t_j$ times its endpoint drift. How much of this
change reaches the output depends on how the editor's later steps respond to it, so the error weight is $g_j=h_j/t_j$ only
up to an editor-dependent amplification. The budget results use $g_j$ only within one step, where it is the same for all
tokens, and the scheme below uses $G_l=\sum_{l'\ge l}g_{l'}$.

\paragraph{Adaptive freezing.}
A more general scheme decides, for every token and step, whether to keep computing it, freeze it, or wake a frozen
token again. Consider a token still computed at step $j$, let $\hat\Delta_i^{\,l}$ forecast its drift at step $l$, and
let $\mu_{k}$ be the price of part $k$. Freezing it after step $j'\ge j$ costs $c_i(j'-j)$ compute and, by the triangle
inequality on future endpoint gaps, at most $\sum_{l>j'}\hat\Delta_i^{\,l}G_l$ error, so its total cost is
\begin{align*}
J_i(j')&=c_i\,(j'-j)+\mu_{k(i)}\sum_{l>j'}\hat\Delta_i^{\,l}\,G_l,\\
G_l&=\sum_{l'\ge l}g_{l'} .
\end{align*}
The token is frozen now iff $j\in\argmin_{j'\ge j}J_i(j')$. Since
\[
J_i(j'+1)-J_i(j')=c_i-\mu_{k(i)}\,\hat\Delta_i^{\,j'+1}G_{j'+1}
\]
and $G_l$ is non-increasing in $l$, non-increasing forecasts make $J_i$ discretely convex, and one-step look-ahead is
exact:
\[
\text{freeze at } j\iff\mu_{k(i)}\,\hat\Delta_i^{\,j+1}G_{j+1}\le c_i .
\]
A frozen token is woken when the forecast error of staying frozen exceeds the cost of computing it again (with
hysteresis), and a per-part ledger keeps the committed error. Waking needs no stored history thanks to endpoint-freeze:
a frozen token sits on a valid straight path and resumes from its current state.

\paragraph{One pass as a special case.}
Switch off waking, forecast revision and contagion between tokens, and take uniform costs $c_i=c$, a geometric
forecast of rate $\bar\rho_j$ and a per-step calibrated price:
\begin{align*}
\hat\Delta_i^{\,l}&=\delta_i\,\bar\rho_j^{\,l-j},\\
\hat G_j(\rho)&=\sum_{l>j}\rho^{\,l-j-1}G_l,\\
\mu_{k,j}&=\frac{c\,(n-1-j)}{\tau\,m_k\,\bar\rho_j\,\hat G_j(\bar\rho_j)} .
\end{align*}
Comparing freezing now ($j'=j$) with never freezing ($j'=n-1$),
\begin{align*}
J_i(j)<J_i(n-1)
&\iff\mu_{k,j}\,\delta_i\,\bar\rho_j\,\hat G_j(\bar\rho_j)<c\,(n-1-j)\\
&\iff\delta_i<\tau\,m_k ,
\end{align*}
which is exactly Rule (4) of the main paper at every step. Deciding only at the first decision step is the strict
one-pass special case. As sets of frozen tokens,
\[
\text{one-pass}\ \subseteq\ \text{freeze-only re-check}\ \subseteq\ \text{adaptive},
\]
so on fixed forecasts the adaptive scheme cannot be worse; whether waking pays for its overheads is empirical.

\paragraph{Premise check.}
\Cref{tab:supp_af9} replays the rule on vanilla trajectories. Re-checking freezes far more tokens than deciding once.
At 4 steps only a few percent of the first-frozen tokens would drift back above their threshold, carrying a small share
of the remaining motion; at 8 steps both numbers grow. We therefore use freeze-only re-checking on the 4-step editor
and leave waking to many-step editors.

\begin{table}[t]
\centering
\tabsize
\setlength{\tabcolsep}{3.5pt}
\begin{tabular}{@{}llrrrrr@{}}
\toprule
Bench. & Steps & $\tau$ & One-pass & Re-check & Re-drift & Movement \\
\midrule
MICE  & 4 & 0.6 & 16.4\% & +45.7\% &  5.8\% &  0.7\% \\
      &   & 1.0 & 64.1\% & +25.3\% &  2.4\% &  2.8\% \\
\addlinespace[2pt]
      & 8 & 0.6 & 14.7\% & +82.6\% & 17.6\% &  2.3\% \\
      &   & 1.0 & 64.4\% & +35.4\% & 11.3\% & 10.8\% \\
\midrule
LoMOE & 4 & 0.6 & 10.5\% & +61.7\% &  6.5\% &  0.5\% \\
      &   & 1.0 & 67.3\% & +24.1\% &  1.4\% &  2.0\% \\
\addlinespace[2pt]
      & 8 & 0.6 &  8.8\% & +88.0\% & 15.6\% &  1.1\% \\
      &   & 1.0 & 67.2\% & +32.4\% &  5.5\% &  7.2\% \\
\bottomrule
\end{tabular}
\caption{\textbf{Premise of adaptive freezing} (vanilla trajectories, benchmark regions). \emph{One-pass}: share of
tokens frozen after the first drift. \emph{Re-check}: additional share frozen at later steps. \emph{Re-drift}: share of
first-frozen tokens whose drift later exceeds the threshold again. \emph{Movement}: their share of the motion after the
first decision.}
\label{tab:supp_af9}
\end{table}
\section{Method and implementation details}
\label{sec:impl}
\begin{figure*}[t]
\centering
\begin{tikzpicture}[
  x=\linewidth, y=1cm,
  img/.style={inner sep=0pt, outer sep=1.5pt, anchor=north},
  cap/.style={font=\scriptsize, align=center, text width=#1, inner sep=1pt, anchor=north, text depth=0pt},
  cap/.default=0.145\linewidth,
  stage/.style={draw=#1!55, fill=#1!5, rounded corners=3pt, inner xsep=0pt, inner ysep=4pt},
  title/.style={font=\footnotesize\bfseries, anchor=south west, inner sep=1.5pt, xshift=2pt},
  arr/.style={-{Stealth[length=4.5pt]}, thick, black!55}]
\pgfmathsetlengthmacro{\imgH}{0.118\linewidth}
\node[img] (in)  at (0.077,0) {\includegraphics[height=\imgH]{fig/method_input.png}};
\node[img] (a1)  at (0.310,0) {\includegraphics[height=\imgH]{fig/method_attn_2.png}};
\node[img] (a2)  at (0.452,0) {\includegraphics[height=\imgH]{fig/method_attn_3.png}};
\node[img] (a3)  at (0.594,0) {\includegraphics[height=\imgH]{fig/method_attn_6.png}};
\node[img] (reg) at (0.912,0) {\includegraphics[height=\imgH]{fig/method_regions.png}};
\node[cap] (inc) at (in.south)  {source (6 parts)};
\node[cap]       at (a1.south)  {$A_{ki}$: ``navy mug''};
\node[cap]       at (a2.south)  {$A_{ki}$: ``turquoise mug''};
\node[cap]       at (a3.south)  {$A_{ki}$: ``pink cupcake''};
\node[cap] (rgc) at (reg.south) {regions $k(i)$};
\draw[arr] (in.east) -- (a1.west);
\draw[arr] (a3.east) -- node[above, font=\scriptsize]{unbalanced OT} node[below, font=\scriptsize]{with demand $b_i$} (reg.west);
\coordinate (l1) at (0.004,0); \coordinate (r1) at (0.996,0);
\begin{scope}[on background layer]
\node[stage=blue, fit=(in)(inc)(reg)(rgc)(l1)(r1)] (s1) {};
\end{scope}
\node[title] at (s1.north west) {\textcolor{blue!60!black}{\ding{172}}\ Self-grounded regions\ \ \normalfont\scriptsize(dense steps)};
\def\yy{-3.35}
\tikzset{plot/.style={inner sep=0pt, outer sep=1.5pt, anchor=center, yshift=-0.5*\imgH},
         pcap/.style={cap=#1, yshift=-\imgH-3pt}, pcap/.default=0.145\linewidth}
\node[img]  (dr)  at (0.077,\yy) {\includegraphics[height=\imgH]{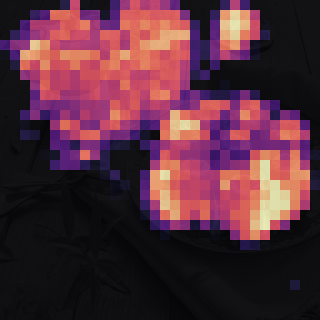}};
\node[plot] (th)  at (0.310,\yy) {\includegraphics[width=0.25\linewidth,height=\imgH,keepaspectratio]{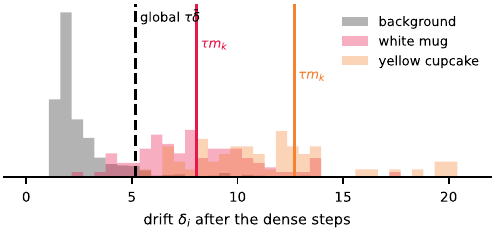}};
\node[img]  (fz)  at (0.535,\yy) {\includegraphics[height=\imgH]{fig/method_frozen_tide.png}};
\node[plot] (cs)  at (0.735,\yy) {\includegraphics[width=0.14\linewidth,height=\imgH,keepaspectratio]{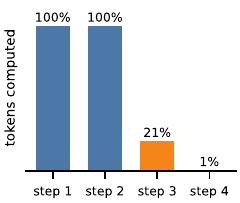}};
\node[img]  (out) at (0.912,\yy) {\includegraphics[height=\imgH]{fig/method_output.png}};
\node[pcap]                 (drc) at (0.077,\yy) {drift $\delta_i$};
\node[pcap=0.25\linewidth]  (thc) at (0.310,\yy) {freeze $i$ if $\delta_i<\tau\,m_{k(i)}$};
\node[pcap]                 (fzc) at (0.535,\yy) {frozen (gray), step 3};
\node[pcap]                 (csc) at (0.735,\yy) {tokens computed};
\node[pcap]                 (otc) at (0.912,\yy) {output, 2.20$\times$};
\draw[arr] (dr.east) -- (dr.east -| th.west);
\draw[arr] (fz.west -| th.east) -- (fz.west);
\draw[arr] (fz.east) -- (fz.east -| cs.west);
\coordinate (l2) at (0.004,\yy); \coordinate (m2) at (0.617,\yy);
\coordinate (m3) at (0.637,\yy); \coordinate (r3) at (0.996,\yy);
\begin{scope}[on background layer]
\node[stage=green,  fit=(dr)(drc)(th)(thc)(fz)(fzc)(l2)(m2)] (s2) {};
\node[stage=orange, fit=(cs)(csc)(out)(otc)(m3)(r3)] (s3) {};
\end{scope}
\node[title] at (s2.north west) {\textcolor{green!45!black}{\ding{173}}\ Per-instruction freezing\ \ \normalfont\scriptsize(later steps)};
\node[title] at (s3.north west) {\textcolor{orange!70!black}{\ding{174}}\ Sparse steps};
\end{tikzpicture}
\caption{\textbf{\method on a real example} (MICE-Bench \#103, FLUX.2 Klein 4B, 4 steps).
\ding{172}~Word attention and edit demand give one region per part. \ding{173}~A token freezes when its drift is small
relative to its own part's mean drift $m_k$; one global threshold (dashed) treats the parts unevenly. \ding{174}~Later
steps compute only active tokens.}
\label{fig:overview}
\end{figure*}

\label{sec:details}
\Cref{fig:overview} walks through \method on one real image, \cref{alg:tide} gives the 4-step version (freeze-only
re-checking), and \cref{tab:supp_settings} lists all its settings.

\paragraph{Instruction parts.}
Benchmarks with per-edit instructions or target phrases (MICE-Bench, MIRA-Bench, InfoEdBench) are split accordingly;
other instructions are split at commas, semicolons, full stops and ``and''. The words $W_k$ of part $k$ are its target
words plus the other content words of their clause, where content words exclude a fixed list of function words and
edit verbs (\eg ``the'', ``replace'', ``turn''; no tagger is used). A word shared by two parts is dropped from both
unless nothing else is left, so that two parts are not pulled onto one region.

\paragraph{Regions.}
During the dense steps we accumulate the softmax attention of every image query (target and reference rows) on each
part's words, over all heads of all double-stream blocks. The edit demand $\omega_i=\|\hat x_{0,i}-r_i\|$ at the last
dense step is normalized robustly (median and MAD) through a softplus. Unbalanced OT then uses
\[
C_{ki}=-\log\tilde A_{ki},\qquad \alpha_k\propto\sum_i\tilde A_{ki}\,b_i ,
\]
with $\tilde A$ the attention normalized per part, followed by total-variation smoothing on the 2D token grid
(Chambolle--Pock~\cite{chambollepock}). A token joins part $k$ when its responsibility exceeds the threshold and
dominates, and its demand mass exceeds the threshold; all other tokens are background.

\paragraph{Freezing.}
The drift scale $m_k$ is the mean drift of part $k$ at the last dense step (background: the image-wide mean). On the
4-step editor, decisions are made at the start of steps 3 and 4, and frozen tokens never wake. \method-quality also
computes the final step in full, where $h/t=1$ and freezing errors weigh most (\cref{sec:supp_endpoint}), and freezes
nothing on global edits such as style changes, detected when more than 70\% of the tokens have edit energy
$\|v-v_{\mathrm{ref}}\|/\|v\|>0.3$ at the region step.

\paragraph{Sparse engine and overheads.}
For each block, the queries, keys and values of the active target rows are written into a per-block buffer that also
holds the text and reference rows (cached at the dense steps) and the frozen target rows (from their last computed
step). Active queries use standard scaled dot-product attention over the full buffer, so every active token still sees
every other token; with all rows active the pass reproduces the dense forward (unit-tested), and no kernel is custom.
The OT solve and smoothing take 4--5\,ms per image and drift bookkeeping adds one subtraction and one norm per computed
token and step; all overheads are included in the reported end-to-end times.

\begin{algorithm}[t]
\caption{\method (training-free; one editor call per step).}
\label{alg:tide}
\small
\begin{algorithmic}[1]
\Require source $I$, instruction parts $1..K$, editor $v_\theta$, steps $n$, $\tau$
\State $r\gets\mathrm{VAE}(I)$; $x\sim\mathcal N(0,\mathbf I)$
\For{$j=0,\dots,s-1$} \Comment{dense steps}
  \State $v\gets v_\theta(x)$; cache text/ref K,V
  \State $\hat x_0\gets x-t_jv$; accumulate attention $A$
  \State $x\gets x-h_jv$
\EndFor
\State $k(i)\gets$ UnbalancedOT$(A,\ b(\|\hat x_0-r\|))$; scales $m_k$
\For{$j=s,\dots,n-1$}
  \State $\mathcal F\gets\mathcal F\cup\{i:\delta_{i}<\tau\,m_{k(i)}\}$ \Comment{freeze-only}
  \State $v_{\mathcal A}\gets v_\theta(x_{\mathcal A};\ \text{cached K,V})$ \Comment{active rows}
  \State $v_{\mathcal F}\gets(x_{\mathcal F}-\hat x_{0,\mathcal F})/t_j$ \Comment{endpoint-freeze}
  \State update $\hat x_{0,\mathcal A}$, $\delta_{\mathcal A}$; \ $x\gets x-h_jv$
\EndFor
\State \Return $\mathrm{VAE}^{-1}(x)$
\end{algorithmic}
\end{algorithm}

\begin{table}[t]
\centering
\tabsize
\begin{tabular}{@{}llr@{}}
\toprule
Stage & Setting & Value \\
\midrule
Parts    & function words / edit verbs removed & 41 \\
\addlinespace[2pt]
Regions  & dense steps $s$ & 2 \\
         & OT entropy $\zeta$; penalties $\rho_1,\rho_2$ & 0.12; 1, 1 \\
         & TV smoothing weight & 0.30 \\
         & responsibility / demand threshold & 0.35 / 0.18 \\
\addlinespace[2pt]
Freezing & $\tau$, 4 steps & 1.0 \\
         & $\tau$, 8-step curves & 0.3--1.5 \\
         & decisions (4 steps) & steps 3, 4 \\
\addlinespace[2pt]
Quality  & final step computed in full & step 4 \\
         & no freezing if share with energy $>0.3$ & $>70\%$ \\
\bottomrule
\end{tabular}
\caption{\textbf{Settings of \method on FLUX.2 Klein 4B.} One value for every part, step, image and benchmark.
Quality: \method-quality, fixed on 12 MICE-Bench and 22 GEdit images (\cref{tab:supp_quality_tune}).}
\label{tab:supp_settings}
\end{table}

\section{Evaluation protocol}
\label{sec:protocol}

\paragraph{Benchmarks.}
\Cref{tab:supp_bench} lists the six benchmarks; \emph{edits} are the instruction parts judged per image. On MICE (public
release, 262 images), the per-edit instructions give 2,030 parts, and the 3 images without them are split as in the main
paper (2,095 in all). \method's own parts are MICE's per-object target phrases (2,113; text only, the masks are used only
for evaluation): in 5 images one judged edit covers several objects (\eg ``the six donuts''), and \method gives each
object its own part while the judge asks one question. InfoDet and Crello are the first 100 images of InfoEdBench's sets
with at least two editable text boxes. On Klein Base and Qwen-Image-Edit, GEdit uses its first 100 images.

\begin{table}[t]
\centering
\tabsize
\begin{tabular}{@{}lrrrr@{}}
\toprule
Benchmark & Images & Edits & Edits/image & Range \\
\midrule
MICE    & 262 & 2095 & 8.0 & 5--36 \\
LoMOE   &  64 &  192 & 3.0 & 2--7 \\
MIRA    & 100 &  500 & 5.0 & 5 \\
InfoDet & 100 &  495 & 5.0 & 3--5 \\
Crello  & 100 &  294 & 2.9 & 2--5 \\
GEdit   & 606 &  606 & 1.0 & 1 \\
\bottomrule
\end{tabular}
\caption{\textbf{Benchmarks.} Edits: instruction parts judged per image.}
\label{tab:supp_bench}
\end{table}

\paragraph{Hardware and timing.}
NVIDIA GTX 1080 Ti (11\,GB), bf16, 512$\times$512, seed 0; the transformer and VAE share one GPU and the text encoder
runs on a second. Times are wall-clock, GPU-synchronised and end to end (text and reference encoding, all transformer
steps, region finding, VAE decoding), after one unrecorded warm-up image; vanilla is timed in the same run on the same
image and seed. The main paper's latency table splits this into prefill, transformer and decoding, and reports the
transformer speed-up $S_\mathrm{dec}$ next to the end-to-end speed-up $S_\mathrm{e2e}$.

\paragraph{Metrics.}
\emph{Done}: share of parts judged ``yes'' ($P(\text{yes})>0.5$); \emph{All}: share of images with every part done.
\emph{Drop}: parts done by vanilla (same steps, image and seed) but not by the method, so it counts only what the
accelerator loses. \emph{LPIPS}~\cite{lpips} (AlexNet) and \emph{PSNR}: method vs.\ vanilla output.
\emph{MAE$_\mathrm{B}$}: mean absolute pixel difference (0--255) to the input outside the edit masks (not on GEdit).
\emph{CLIP check}: an edit counts as applied if the CLIP~\cite{clip} similarity of its object crop to the edit text
rises from input to output.
\paragraph{VLM judge.}
Qwen3-VL-8B-Instruct~\cite{qwen3vl}, greedy, with $P(\text{yes})$ read from the answer logits. Both images are encoded
once and every part is asked on the cached prefix (GEdit without the cache, for all its methods); the same judge,
prompts and vanilla reference serve every method of a benchmark. The prompts are:

\smallskip
\noindent\fbox{\begin{minipage}{\dimexpr\linewidth-2\fboxsep-2\fboxrule\relax}\footnotesize
\textbf{Edit.} You are shown two images: first the ORIGINAL image, then the EDITED image. The full editing instruction
was: \{instruction\}. Question: in the EDITED image, was this part of the instruction carried out correctly: \{part\}?
Answer with yes or no.\\[2pt]
\textbf{Quality} (\cref{tab:supp_sweep8}). This is an edited image. Rate its visual quality (natural look, no artifacts,
no distortions) on a scale from 1 (very poor) to 5 (excellent). Answer with a single digit.
\end{minipage}}

\paragraph{Reliability of the judge.}
Beyond agreement with raters (as in VIEScore~\cite{viescore}), we test the judge on the 167 human-checked edits
(\cref{tab:supp_judge}). \emph{Reproducibility:} rescoring on CPU in fp32 reproduces every yes/no answer (GPU, bf16).
\emph{Wording:} 84.4\% of the edits get the same answer under the original and two rewordings (pairwise 89.6\%).
\emph{Self-improvement:} on a development half (18 images), the judge rewrites its prompt from its disagreements with the
raters (3 rounds, 2 candidates each); none of the six candidates beats the original (82.9--87.5\% vs.\ 87.5\%), which
agrees with the raters on 88.6\% of the held-out half (AC1 0.87). \emph{Leniency:} the second rewording agrees more
(94.6\%) only by saying ``done'' to 97\% of the edits and missing every edit the raters marked not done, so we keep the
stricter original.

\begin{table}[t]
\centering
\tabsize
\setlength{\tabcolsep}{4pt}
\begin{tabular}{@{}lrrrrr@{}}
\toprule
Prompt & Edits & Agree & AC1 & Done & Caught \\
\midrule
Original (paper)    & 167 & 88.0 & 0.86 & 89.2 & 1/4 \\
Rewording 1         & 167 & 84.4 & 0.82 & 85.6 & 1/4 \\
Rewording 2         & 167 & 94.6 & 0.94 & 97.0 & 0/4 \\
\midrule
Original, test half &  79 & 88.6 & 0.87 & 89.9 & 1/3 \\
\bottomrule
\end{tabular}
\caption{\textbf{Judge vs.\ raters' majority} (raters: 97.6\% done; Agree and Done in \%). \emph{Caught}: edits the
raters mark not done that the judge also marks not done. The best self-refined prompt is the original.}
\label{tab:supp_judge}
\end{table}

\paragraph{Human study.}
Raters see the input and one edited image with its edits, answer yes / no / unclear / can't tell per edit, and rate
damage (none / slight / clear) and quality (1--5) per image (\cref{fig:supp_ui}). The 36 images follow fixed rules that
ignore every output: at most 6 edits, every edited object at least 1\% of the image, no ambiguous references, and
vanilla completes every edit. Vanilla, \method and one global threshold (Klein 4B, 4 steps) give 12 images each, 167
edits in all, in random order with the method hidden. All 33 raters answered all 239 questions, never with ``unclear''
or ``can't tell''; labels are the raters' majority.
The judge agrees with the raters on 88.0\% of the edits (AC1 0.86, PABAK 0.76; Cohen's $\kappa$ 0.05 only because 98\%
of the labels are ``done''), and the raters confirm 98.0\% of its ``done'' answers. Between raters, Fleiss' $\kappa$ is
0.66 on edits (93\% unanimous) and Krippendorff's $\alpha$ 0.33 on quality, where nearly all scores are 4 or 5; judge
quality and rater MOS correlate at Spearman $-0.10$ (\cref{tab:supp_human}; main paper, Sec.~4.7).

\begin{table}[t]
\centering
\tabsize
\setlength{\tabcolsep}{3.2pt}
\begin{tabular}{@{}lrrrrrrrr@{}}
\toprule
 & & & \multicolumn{2}{c}{Edits done} & & \multicolumn{2}{c}{All done} & \\
\cmidrule(lr){4-5}\cmidrule(lr){7-8}
 & Imgs & Edits & Raters & Judge & Agree & Raters & Judge & Dmg. \\
\midrule
Vanilla     & 12 &  52 & 100.0 & 100.0 & 100.0 & 100 & 100 & 0 \\
\method     & 12 &  51 &  96.1 &  80.4 &  76.5 &  83 &  50 & 0 \\
Global thr. & 12 &  64 &  96.9 &  87.5 &  87.5 &  83 &  75 & 8 \\
\midrule
All         & 36 & 167 &  97.6 &  89.2 &  88.0 &  89 &  75 & 3 \\
\bottomrule
\end{tabular}

\vspace{0.5em}
\setlength{\tabcolsep}{3pt}
\begin{tabular}{@{}lccc@{}}
\toprule
Quality (1--5) & Vanilla & \method & Global thr. \\
\midrule
Raters & 4.74 {\scriptsize[4.52, 4.92]} & 4.75 {\scriptsize[4.55, 4.92]} & 4.87 {\scriptsize[4.81, 4.93]} \\
Judge  & 3.73 {\scriptsize[3.43, 3.95]} & 3.24 {\scriptsize[2.95, 3.53]} & 3.12 {\scriptsize[2.91, 3.35]} \\
\bottomrule
\end{tabular}
\caption{\textbf{Human study} (36 images, 33 raters). Top: edits done, images with all edits done, and images called
damaged (\%). Bottom: mean quality, 95\% bootstrap intervals over images.}
\label{tab:supp_human}
\end{table}

\begin{figure}[t]
\centering
\pgfplotsset{hc/.style={scale only axis, height=2.1cm, width=0.32\linewidth,
  tick label style={font=\scriptsize}, label style={font=\scriptsize},
  title style={font=\footnotesize, yshift=-3pt}, axis line style={gray!60}, tick style={gray!60},
  xmajorgrids, grid style={gray!15}, ytick style={draw=none}, axis y line*=left, axis x line*=bottom,
  xlabel style={yshift=3pt}}}
\begin{tikzpicture}
\begin{axis}[hc, name=pa, title={(a) Discordant edits lost},
  xbar=1pt, bar width=5.5pt, xmin=0, xmax=125, xtick={0,50,100}, xlabel={edits},
  ymin=0.4, ymax=2.6, ytick={1,2}, yticklabels={\method,Global},
  nodes near coords, nodes near coords style={font=\scriptsize, inner sep=1.5pt},
  legend to name=humleg, legend columns=2, legend style={font=\scriptsize, draw=none, column sep=8pt},
  legend image code/.code={\draw[#1] (0cm,-0.08cm) rectangle (0.3cm,0.08cm);}]
\addplot[fill=cvprblue, draw=cvprblue]   coordinates {(15,1) (49,2)};
\addlegendentry{Raters}
\addplot[fill=black!20, draw=black!50]   coordinates {(25,1) (103,2)};
\addlegendentry{Judge}
\end{axis}
\begin{axis}[hc, at={(pa.east)}, anchor=west, xshift=1.35cm, title={(b) Image quality},
  xmin=2.8, xmax=5.1, xtick={3,4,5}, xlabel={mean score (1--5)},
  ymin=0.4, ymax=3.6, ytick={1,2,3}, yticklabels={Global,\method,Vanilla}]
\addplot[only marks, mark=*, mark size=1.8pt, cvprblue,
  error bars/.cd, x dir=both, x explicit, error bar style={cvprblue, line width=0.8pt}]
  table[x index=0, y index=1, x error minus index=2, x error plus index=3]{
  4.738 3.15 0.214 0.179
  4.750 2.15 0.202 0.167
  4.869 1.15 0.060 0.060
};
\addplot[only marks, mark=*, mark size=1.8pt, mark options={fill=black!20, draw=black!60},
  error bars/.cd, x dir=both, x explicit, error bar style={black!55, line width=0.8pt}]
  table[x index=0, y index=1, x error minus index=2, x error plus index=3]{
  3.726 2.85 0.294 0.220
  3.237 1.85 0.291 0.289
  3.124 0.85 0.211 0.225
};
\end{axis}
\node[anchor=south] at ([yshift=-1pt]current bounding box.north) {\pgfplotslegendfromname{humleg}};
\end{tikzpicture}
\caption{\textbf{Human checks} (Klein 4B, 2.15$\times$; 33 raters). (a) Edits each rule loses among the 128 that exactly
one rule loses for the judge. (b) Mean image quality, 95\% bootstrap intervals.}
\label{fig:human}
\end{figure}
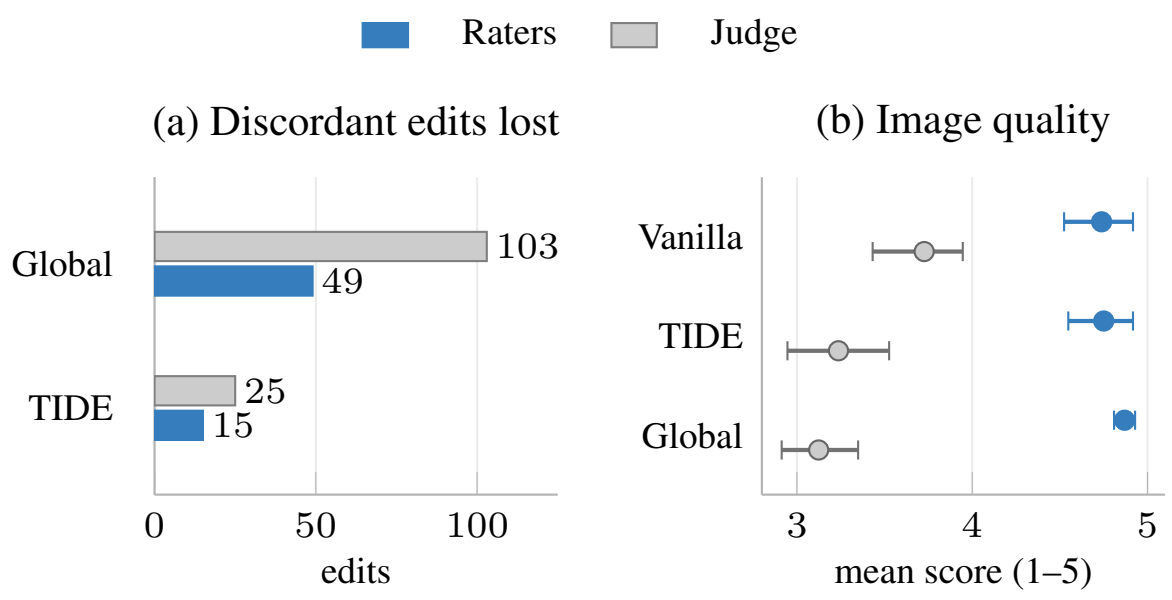

\paragraph{Paired check of the discordant edits.}
At matched speed on Klein 4B (2.15$\times$, $\tau{=}1.0$ for both rules), the judge finds 108 edits lost only by the
global threshold and 26 only by \method (\cref{tab:supp_stats}). We keep those on images whose masks match their parts
(103 and 25) and add 30 controls kept (20) or lost (10) by both: 158 edits on 92 images. Each screen shows the edit, the
input and the two outputs as ``A'' and ``B'' (random side, names hidden), with a dashed box and zoomed crops so that
objects under 1\% of the image (42 edits) can be judged. Two halves with the same mix are rated by 17 and 16 raters (16--17
votes per edit; majority label, ties unresolved).
The raters confirm 49 lost edits for the global threshold and 15 for \method (\cref{tab:supp_pairs}, \cref{fig:human}a):
a difference of 34, 95\% bootstrap interval [19, 50]; McNemar on edits lost by one rule only (47 vs.\ 13),
$p{=}1.2{\times}10^{-5}$. Raters agree with the majority on 93\% of control answers (lowest rater 87\%). Overall they
confirm 86\% of the judge's ``done'' answers (144/168) but only 34\% of its ``not done'' (50/148): the judge is strict,
so Drop overcounts absolute losses, but the gap between the rules holds.
\begin{table}[t]
\centering
\tabsize
\setlength{\tabcolsep}{3.5pt}
\begin{tabular}{@{}lrcccc@{}}
\toprule
 & & \multicolumn{2}{c}{Global threshold} & \multicolumn{2}{c}{\method} \\
\cmidrule(lr){3-4}\cmidrule(lr){5-6}
Judge says & Edits & Done & Not & Done & Not \\
\midrule
Lost by global only & 103 & 59 & \textbf{43} & \textbf{92} & 11 \\
Lost by \method only &  25 & 19 & 6 & 21 & \textbf{4} \\
Kept by both        &  20 & 15 & 5 & 18 & 2 \\
Lost by both        &  10 &  9 & 1 &  8 & 2 \\
\midrule
\multicolumn{6}{@{}l}{\emph{Raters, on the 128 discordant edits}} \\
Lost                     & & \multicolumn{2}{c}{49} & \multicolumn{2}{c}{\textbf{15}} \\
Lost by this rule only   & & \multicolumn{2}{c}{47} & \multicolumn{2}{c}{13} \\
Objects $<$1\% (35 edits) & & \multicolumn{2}{c}{11} & \multicolumn{2}{c}{4} \\
Fleiss' $\kappa$         & & \multicolumn{2}{c}{0.74} & \multicolumn{2}{c}{0.58} \\
\bottomrule
\end{tabular}
\caption{\textbf{Paired human check} (Klein 4B, MICE-Bench, 2.15$\times$). Top: the raters' verdict on each rule's
output, per judge group (one tie in the first row). Bottom: edits the raters confirm as lost.}
\label{tab:supp_pairs}
\end{table}

\paragraph{Human-subjects statement.}
Both studies (main paper, Secs.~4.2 and~4.7) used the same 33 raters: unpaid adult volunteers (researchers and graduate
students), about one hour each, who gave informed consent and could stop at any time. They saw only public benchmark
images and their edited versions; answers were stored under anonymous codes and no personal data were collected.
\begin{figure*}[t]
\centering
\pgfplotsset{hfig/.style={scale only axis,height=3.1cm,tick label style={font=\tiny},label style={font=\scriptsize},
  title style={font=\scriptsize,yshift=-3pt},axis line style={gray!70},tick style={gray!70}}}
\begin{tikzpicture}
\begin{axis}[hfig,width=3.1cm,title={(a) Edits: judge vs.\ raters},xmin=-0.5,xmax=1.5,ymin=-0.5,ymax=1.5,
  xtick={0,1},xticklabels={done,not done},ytick={0,1},yticklabels={not done,done},xlabel={VLM judge},ylabel={Raters (majority)},
  colormap={bl}{color=(white) color=(cvprblue!35) color=(cvprblue)},point meta min=0,point meta max=146,
  enlargelimits=false,axis on top]
\addplot[matrix plot*,mesh/cols=2,point meta=explicit] table[meta=c]{x y c
  0 1 146
  1 1 17
  0 0 3
  1 0 1
};
\node[font=\scriptsize\bfseries,text=white] at (axis cs:0,1) {146};
\node[font=\scriptsize\bfseries,text=black] at (axis cs:1,1) {17};
\node[font=\scriptsize\bfseries,text=black] at (axis cs:0,0) {3};
\node[font=\scriptsize\bfseries,text=black] at (axis cs:1,0) {1};
\end{axis}
\end{tikzpicture}\hspace{0.6em}%
\begin{tikzpicture}
\begin{axis}[hfig,width=4.2cm,title={(b) Calibration of the judge},xmin=0,xmax=100,ymin=0,ymax=105,
  xlabel={Judge P(done) (\%)},ylabel={Raters saying done (\%)},grid=major,grid style={gray!15}]
\addplot[gray!60,dashed,domain=0:100,samples=2] {x};
\addplot[only marks,mark=*,cvprblue,mark size=2.2pt,error bars/.cd,y dir=both,y explicit,error bar style={cvprblue,line width=0.6pt},
  error mark options={rotate=90,mark size=2pt}] table[x index=0,y index=1,y error minus index=2,y error plus index=3]{
  8.7 100.0 0.0 0.0 12
  31.0 100.0 0.0 0.0 3
  46.9 75.0 50.0 25.0 4
  73.8 100.0 0.0 0.0 8
  99.0 96.4 2.8 2.1 140
};
\addplot[only marks,mark=none,nodes near coords,point meta=explicit symbolic,every node near coord/.append style={font=\tiny,anchor=north,yshift=-2pt,text=black!70}]
  table[x index=0,y index=1,meta index=2]{
  x y l
  8.7 100.0 n=12
  31.0 100.0 n=3
  46.9 25.0 n=4
  73.8 100.0 n=8
  99.0 93.7 n=140
};
\end{axis}
\end{tikzpicture}\hspace{0.6em}%
\begin{tikzpicture}
\begin{axis}[hfig,width=4.2cm,title={(c) Image quality},xmin=1,xmax=5.2,ymin=0.4,ymax=3.6,ytick={1,2,3},yticklabels={Global thr.,\method,Vanilla},
  xlabel={Score (1--5)},xmajorgrids,grid style={gray!15},legend style={font=\tiny,draw=none,fill=none,at={(0.02,0.98)},anchor=north west},
  legend cell align=left]
\draw[gray!55,line width=1.2pt] (axis cs:3.726,3) -- (axis cs:4.738,3);
\draw[gray!55,line width=1.2pt] (axis cs:3.237,2) -- (axis cs:4.750,2);
\draw[gray!55,line width=1.2pt] (axis cs:3.124,1) -- (axis cs:4.869,1);
\addplot[only marks,mark=*,cvprblue,mark size=2.2pt,error bars/.cd,x dir=both,x explicit,error bar style={cvprblue}]
  table[x index=0,y index=1,x error minus index=2,x error plus index=3]{
  4.738 3 0.214 0.179
  4.750 2 0.202 0.167
  4.869 1 0.060 0.060
};
\addlegendentry{Raters (MOS)}
\addplot[only marks,mark=*,mark options={fill=white,draw=black!65},mark size=2.2pt,error bars/.cd,x dir=both,x explicit,error bar style={black!65}]
  table[x index=0,y index=1,x error minus index=2,x error plus index=3]{
  3.726 3 0.294 0.220
  3.237 2 0.291 0.289
  3.124 1 0.211 0.225
};
\addlegendentry{VLM judge}
\end{axis}
\end{tikzpicture}
\vspace{-0.5em}
\caption{\textbf{Human check vs.\ the VLM judge} (36 images, 167 edits, 33 raters). (a) Judge vs.\ raters' majority:
88\% agreement, 98\% of judge ``done'' confirmed; the judge is stricter. (b) Raters' ``done'' share by judge
confidence, 95\% bootstrap intervals; dashed: perfect calibration. (c) Mean quality per group (12 images each),
95\% intervals: all near the ceiling and overlapping, so the study does not rank quality.}
\label{fig:supp_human}
\end{figure*}
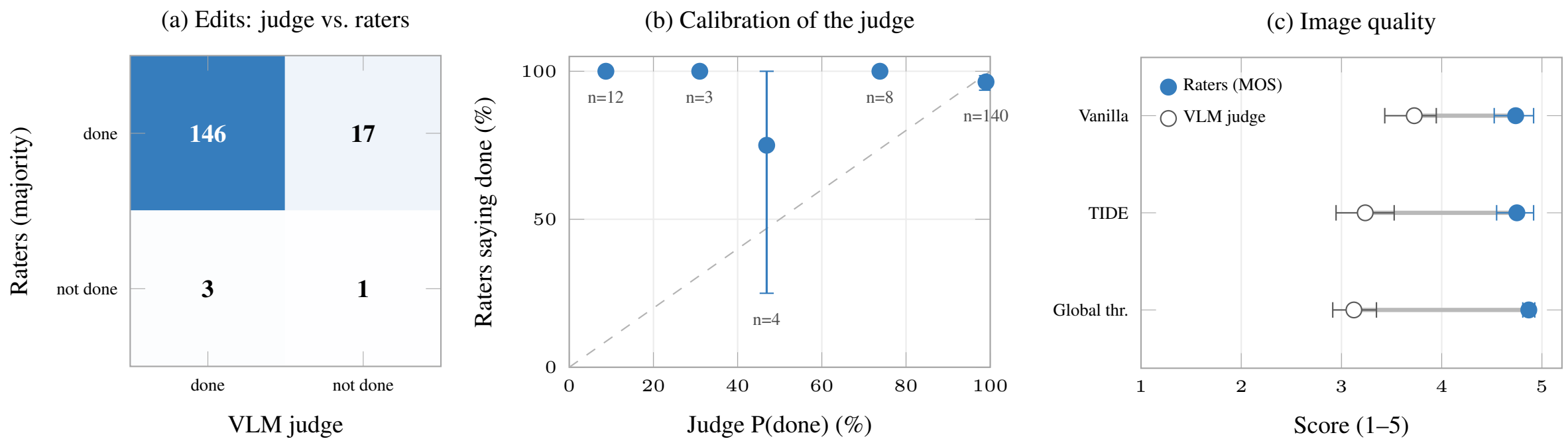

\begin{figure}[t]
\centering
\includegraphics[width=\linewidth]{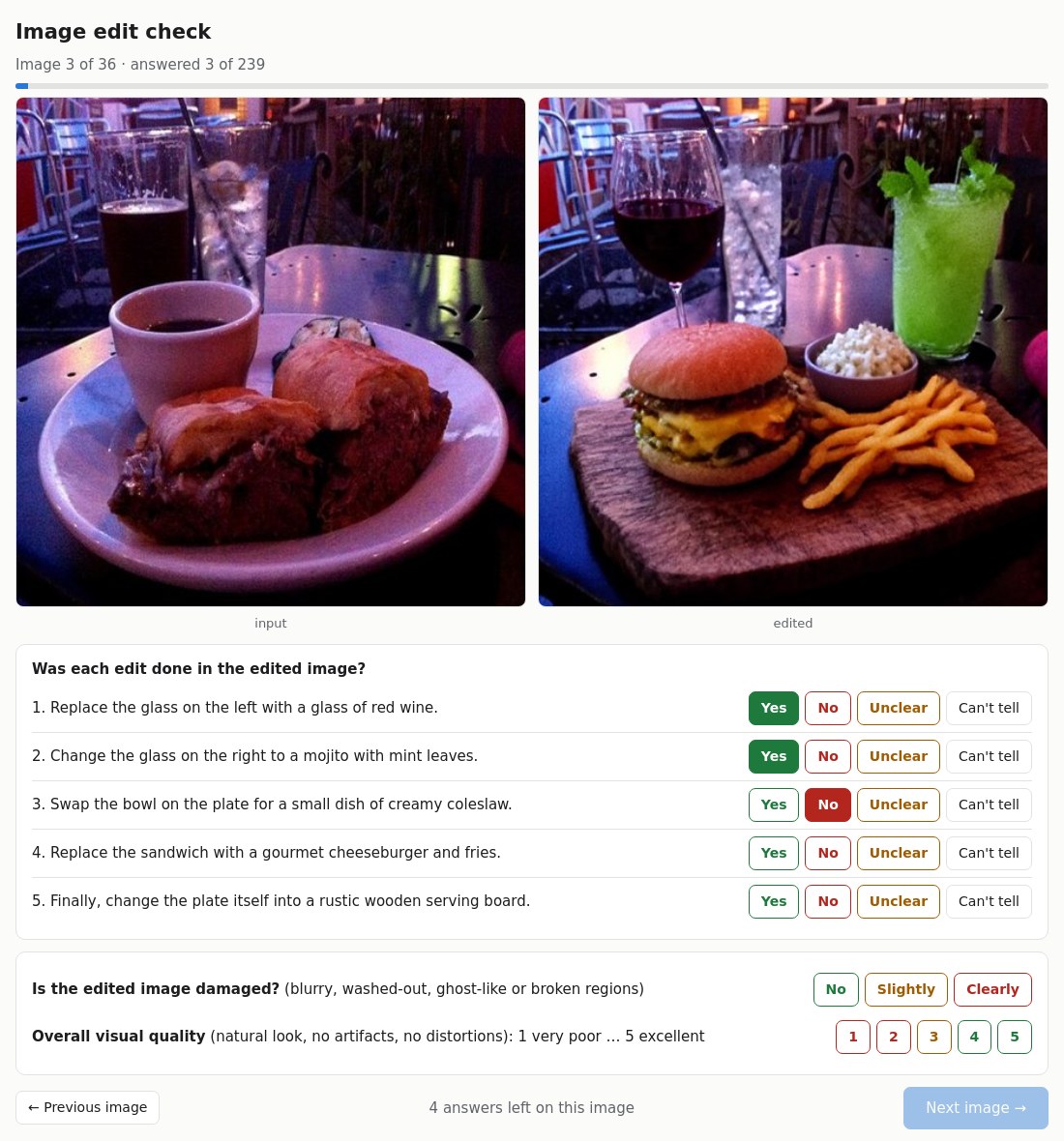}
\caption{\textbf{Human-study interface}}
\label{fig:supp_ui}
\end{figure}

\section{Baselines}
\label{sec:baselines}
All baselines use their official code for every decision and run on the same editor, images, seed, resolution and GPUs
as \method (\cref{tab:supp_baselines}); \cref{tab:supp_position} positions \method against the closest accelerators.
Methods designed for 28--50 steps keep their rules, with schedules scaled to 4 steps in the same proportions. On
11\,GB GPUs we needed exact, memory-only adaptations: query-chunked attention (exact up to rounding), the last
transformer blocks or the VLM split over two GPUs, and MICE's per-instance prompts encoded four at a time.

\begin{table*}[t]
\centering
\footnotesize
\setlength{\tabcolsep}{4pt}
\renewcommand{\arraystretch}{0.95}
\begin{tabular}{@{}>{\raggedright}p{0.10\textwidth}>{\raggedright}p{0.255\textwidth}>{\raggedright}p{0.33\textwidth}>{\raggedright\arraybackslash}p{0.25\textwidth}@{}}
\toprule
Method & Official setting & Klein 4B, 4 steps (main) & Klein Base, 28 steps \\
\midrule
RegionE~\cite{regione} & FLUX.1 Kontext, 28 steps; warm-up 6, refresh 16, final 2; cosine 0.93, erosion/dilation; velocity cache
  & official \texttt{token\_selector}, 0.93; warm-up 1, final 1; no refresh or velocity cache; reference and unedited K/V cached
  & warm-up 6, refresh 15, final 2; 0.93 \\
TeaCache~\cite{teacache} & FLUX.1 polynomial; thresholds 0.25--0.8
  & polynomial refitted for Klein (official procedure, 102 pairs); 0.8
  & refitted for Klein Base; 0.8 \\
TaylorSeer~\cite{taylorseer} & FLUX, 50 steps; first 3 full, every 6th full; order 1
  & first 2 full, interval 3, order 1
  & first 3 full, every 6th full; order 1 \\
RAS~\cite{ras} & SD3, 28 steps; ratio 0.5, start 4, resets 12/20, starvation 1, high ratio 0.3
  & official rule; start 1, no reset; reference tokens cached
  & official settings \\
SpotEdit~\cite{spotedit} & Klein code, 50 steps; LPIPS 0.4 from step 4; resets 13/22/31
  & judging from step 1 (from step 4 nothing is skipped); no resets
  & LPIPS 0.4 from step 4; resets 13/22 \\
MICE~\cite{mice} & Klein 4B, 4 steps, $\approx$1024$^2$, masks per instance
  & official (kernel 11, temp.\ 3.0, binding layers 0--5 / 0--20); 512$^2$
  & official, guidance 4.0 \\
MIRAGE~\cite{mirage} & Klein base 9B, 50 steps; Qwen3.5-9B localizer; patch 0.4
  & Qwen3-VL-8B localizer (official option); patch 0.4; VLM time included
  & official, guidance 4.0; patch 0.4 \\
\bottomrule
\end{tabular}
\caption{\textbf{Baseline settings.} Decision rules are always the official code.}
\label{tab:supp_baselines}
\end{table*}

\begin{table*}[t]
\centering
\footnotesize
\setlength{\tabcolsep}{5pt}
\renewcommand{\arraystretch}{0.95}
\begin{tabular}{@{}lllll@{}}
\toprule
Method & Skips & Decision criterion & Regions from & Skipped tokens \\
\midrule
TeaCache~\cite{teacache}                  & whole steps                   & one per image         & --                              & cached output reused \\
TaylorSeer~\cite{taylorseer}              & steps between full ones       & fixed schedule        & --                              & features forecast \\
RAS~\cite{ras}                            & tokens outside the focus      & one per-token score   & model's focus                   & previous output reused \\
RegionE~\cite{regione}                    & unedited region               & one per image         & edited vs.\ unedited split      & finished early \\
SpotEdit~\cite{spotedit}                  & unchanged tokens              & one per image         & unchanged-token test            & source features reused \\
EEdit~\cite{eedit}, KV-Edit~\cite{kvedit} & outside the edit region       & given region          & user mask                       & kept from the source \\
SpecEdit~\cite{specedit}                  & tokens unchanged in a draft   & one per image         & low-res draft vs.\ source       & kept at coarse resolution \\
RegionCache~\cite{regioncache}            & regions shared with last turn & one per turn          & cross-attention on shared words & cached states reused \\
\midrule
\textbf{\method} & \textbf{tokens} & \textbf{one per instruction part} & \textbf{editor's attention, via OT} & \textbf{endpoint-freeze, wake-up} \\
\bottomrule
\end{tabular}
\caption{\textbf{Positioning.} Only \method sets its skipping criterion per instruction part.}
\label{tab:supp_position}
\end{table*}

\paragraph{Klein Base 4B (28 steps, guidance 4.0).}
Step skip runs vanilla Klein Base with 7 or 14 steps, and \emph{no guidance} runs it with guidance 1.0 (one branch per
step). \method uses $s{=}4$ dense steps, a check every $R{=}8$ steps, $\tau{=}1.5$, the empty-prompt branch every third
sparse step and a dense last step, unchanged on all benchmarks; on 14 steps it uses $s{=}2$ and $R{=}4$ with the same
$\tau$ (\cref{tab:supp_bsteps}). These values were fixed on pilot runs on the first 3 and 20 MICE-Bench images and not
changed afterwards. On all 262 MICE-Bench images, the final setting runs at 4.85$\times$ with 76 drops, and Drop barely
depends on $\tau$ (70--76 for $\tau{=}0.6$--$3.0$, \cref{tab:supp_base_abl}).

\section{Additional quantitative results}
\label{sec:moreresults}

\paragraph{Significance.}
All Drop comparisons are paired: both methods are judged on the same edits (\cref{tab:supp_stats}). Per-part budgets
beat one global threshold significantly at medium speed on Klein 4B (\eg 2.15$\times$: 237 vs.\ 155, $p<10^{-12}$),
also on LoMOE, MIRA, InfoDet and Crello at 8 steps ($p\le0.031$), and tie at the ends of the range. On Klein Base, \method and TeaCache differ significantly only on MICE-Bench, where
\method drops fewer edits, and per-part thresholds drop fewer edits than a global one at every $\tau$ (70--76 vs.\
116--125, $p{<}0.003$). A different seed alone changes about 6\% of vanilla's completed edits, which sets the scale
for small differences.

\paragraph{Further results.}
\Cref{tab:supp_perbench} gives speed, edits done and fidelity for every method and benchmark, \cref{tab:supp_density}
splits Drop by edit density, \cref{tab:supp_gedit} by GEdit task, and \cref{tab:supp_robust} checks seeds, stacking
with 2 steps, 768$^2$ and PIE-Bench++.

\begin{table*}[t]
\centering
\footnotesize
\setlength{\tabcolsep}{2.4pt}
\renewcommand{\arraystretch}{0.95}
\begin{minipage}[t]{0.49\textwidth}
\vspace{0pt}
\begin{tabular*}{\linewidth}{@{\extracolsep{\fill}}lrrrrrr@{}}
\multicolumn{7}{@{}l}{\textbf{Klein 4B}, MICE-Bench (262 images) unless noted} \\
\toprule
A vs.\ B & A & B & A$-$B & 95\% CI & Only & $p$ \\
\midrule
\multicolumn{7}{@{}l}{\emph{Global vs.\ \method}} \\
4 st., 2.15$\times$ & 237 & 155 &  +82 & [+51, +115]  & 108/26 & $4{\times}10^{-13}$ \\
4 st., 1.95$\times$ & 162 & 166 &   -4 & [-19, +11]   &  25/29 & 0.68 \\
8 st., 2.6$\times$  & 151 & 154 &   -3 & [-20, +14]   &  28/31 & 0.79 \\
8 st., 3.5$\times$  & 384 & 200 & +184 & [+137, +233] & 210/26 & $6{\times}10^{-37}$ \\
8 st., 4.0$\times$  & 269 & 169 & +100 & [+67, +134]  & 123/23 & $1{\times}10^{-17}$ \\
8 st., 4.3$\times$  & 157 & 153 &   +4 & [-13, +22]   &  34/30 & 0.71 \\
\midrule
\multicolumn{7}{@{}l}{\emph{Global vs.\ \method, 8 st., $\tau{=}1.0$ (main paper, Tab.~1)}} \\
LoMOE (64)    & 26 &  6 & +20 & [+10, +31] & 21/1 & $1{\times}10^{-5}$ \\
MIRA (100)    & 74 & 42 & +32 & [+20, +45] & 38/6 & $9{\times}10^{-7}$ \\
InfoDet (100) & 23 & 13 & +10 & [+2, +19]  & 14/4 & 0.031 \\
Crello (100)  & 57 & 33 & +24 & [+10, +39] & 31/7 & $1{\times}10^{-4}$ \\
\midrule
\multicolumn{7}{@{}l}{\emph{vs.\ baselines} (q.: quality)} \\
\method-fast vs.\ step skip & 155 & 57 & +98 & [+72, +125] & 121/23 & $3{\times}10^{-17}$ \\
\method-q.\ vs.\ step skip  & 121 & 57 & +64 & [+41, +87]  &  89/25 & $1{\times}10^{-9}$ \\
\method-q.\ vs.\ TeaCache   & 121 & 91 & +30 & [+8, +52]   &  67/37 & 0.004 \\
\bottomrule
\end{tabular*}
\end{minipage}\hfill
\begin{minipage}[t]{0.49\textwidth}
\vspace{0pt}
\begin{tabular*}{\linewidth}{@{\extracolsep{\fill}}lrrrrrr@{}}
\multicolumn{7}{@{}l}{\textbf{Klein Base}, 28 steps} \\
\toprule
A vs.\ B & A & B & A$-$B & 95\% CI & Only & $p$ \\
\midrule
\multicolumn{7}{@{}l}{\emph{\method vs.\ TeaCache}} \\
MICE      & 76 & 100 & -24 & [-44, -4] & 39/63 & 0.02 \\
LoMOE     &  4 &   9 &  -5 & [-12, +2] &   3/8 & 0.23 \\
MIRA      & 25 &  28 &  -3 & [-11, +5] &  7/10 & 0.63 \\
InfoDet   &  6 &   8 &  -2 & [-6, +2]  &   1/3 & 0.62 \\
Crello    & 22 &  25 &  -3 & [-13, +7] & 11/14 & 0.69 \\
GEdit$^*$ & 11 &   6 &  +5 & [-1, +11] &   7/2 & 0.18 \\
\midrule
\multicolumn{7}{@{}l}{\emph{Per-part vs.\ global, MICE}} \\
$\tau{=}0.6$ & 70 & 116 & -46 & [-60, -32] & -- & $<$0.001$^\P$ \\
$\tau{=}1.5$ & 76 & 120 & -44 & [-64, -24] & -- & $<$0.003$^\P$ \\
$\tau{=}2.5$ & 74 & 125 & -51 & [-68, -34] & -- & $<$0.001$^\P$ \\
$\tau{=}3.0$ & 76 & 121 & -45 & [-60, -30] & -- & $<$0.002$^\P$ \\
\bottomrule
\end{tabular*}
\end{minipage}
\caption{\textbf{Significance of the Drop differences.} A, B: Drop against vanilla on the same images. CI: paired
bootstrap over images (10,000 resamples). Only: edits lost by only A / only B; $p$: exact two-sided McNemar on those.
Noise floor (vanilla, seed 1 or 2 vs.\ seed 0): 96--101 of 1,577 edits on Klein 4B, 53--55 of 501 on Klein Base (100 MICE
images). $^*$First 100 images. $^\P$Largest $p$ over all discordant splits consistent with the Drop totals.}
\label{tab:supp_stats}
\end{table*}

\begin{table*}[t]
\centering
\footnotesize
\setlength{\tabcolsep}{4pt}
\renewcommand{\arraystretch}{0.92}
\begin{tabular*}{\textwidth}{@{\extracolsep{\fill}}lrrrrrr@{}}
\toprule
Method & MICE & LoMOE & MIRA & InfoDet & Crello & GEdit \\
\midrule
\multicolumn{7}{@{}l}{\emph{(a) Speed-up / PSNR to vanilla}} \\
Step skip        & 1.84$\times$      & 1.82$\times$      & 1.82$\times$      & 1.82$\times$      & 1.80$\times$      & 1.81$\times$ \\
RegionE          & 1.30$\times$/20.9 & 1.35$\times$/23.7 & 1.40$\times$/26.0 & 1.51$\times$/21.6 & 1.52$\times$/25.2 & 1.38$\times$/26.9 \\
TaylorSeer       & 1.79$\times$/14.2 & 1.79$\times$/15.7 & 1.79$\times$/17.3 & 1.78$\times$/18.4 & 1.79$\times$/19.1 & 1.79$\times$/16.7 \\
TeaCache         & 1.77$\times$/18.6 & 1.80$\times$/21.4 & 1.81$\times$/24.2 & 1.82$\times$/21.9 & 1.82$\times$/25.4 & 1.78$\times$/25.0 \\
RAS              & 1.65$\times$/17.7 & 1.65$\times$/19.9 & 1.66$\times$/21.5 & 1.67$\times$/19.5 & 1.64$\times$/21.8 & 1.67$\times$/21.7 \\
SpotEdit         & 1.54$\times$/20.9 & 1.67$\times$/22.2 & 1.77$\times$/22.7 & 1.95$\times$/20.7 & 1.97$\times$/24.1 & 1.63$\times$/29.3 \\
MICE$^\dagger$   & 0.47$\times$/14.4 & 0.74$\times$/16.9 & 0.61$\times$/18.2 & 0.61$\times$/19.6 & 0.74$\times$/21.9 & n/a \\
MIRAGE$^\ddagger$& 0.10$\times$/17.2 & 0.22$\times$/20.9 & 0.11$\times$/23.3 & 0.09$\times$/22.2 & 0.15$\times$/25.9 & 0.37$\times$/30.6 \\
\textbf{\method-fast}    & 2.20$\times$/17.8 & 2.22$\times$/19.7 & 2.24$\times$/21.6 & 2.20$\times$/19.6 & 2.24$\times$/22.2 & 2.21$\times$/21.6 \\
\textbf{\method-quality} & 1.79$\times$/18.0 & 1.89$\times$/19.8 & 1.92$\times$/21.7 & 1.93$\times$/19.2 & 1.92$\times$/22.0 & 1.85$\times$/21.2 \\
\midrule
\multicolumn{7}{@{}l}{\emph{(b) Edits done / images with every edit done (\%)}} \\
Vanilla          & 75.3/35.1 & 82.3/59.4 & 80.4/32.0 & 72.7/25.0 & 70.1/49.0 & 85.6/85.6 \\
Step skip        & 76.6/37.4 & 84.4/65.6 & 80.2/30.0 & 75.2/28.0 & 74.1/55.0 & 86.0/86.0 \\
RegionE          & 77.0/37.4 & 87.5/70.3 & 79.2/32.0 & 74.3/27.0 & 70.4/47.0 & 85.0/85.0 \\
TaylorSeer       & 67.2/21.0 & 75.0/54.7 & 71.4/23.0 & 73.5/27.0 & 68.7/54.0 & 74.3/74.3 \\
TeaCache         & 75.6/35.5 & 85.9/65.6 & 77.4/27.0 & 73.1/23.0 & 68.7/42.0 & 86.0/86.0 \\
RAS              & 73.7/33.2 & 84.4/65.6 & 74.8/28.0 & 75.2/31.0 & 69.4/47.0 & 84.5/84.5 \\
SpotEdit         & 75.0/34.0 & 86.5/67.2 & 65.6/13.0 & 74.1/29.0 & 61.6/35.0 & 79.7/79.7 \\
MICE$^\dagger$   & 68.1/29.0 & 83.9/65.6 & 59.2/7.0  & 73.7/24.0 & 71.4/52.0 & n/a \\
MIRAGE$^\ddagger$& 73.4/34.7 & 80.2/57.8 & 77.6/25.0 & 73.7/25.0 & 70.7/43.0 & 81.7/81.7 \\
\textbf{\method-fast}    & 72.1/32.1 & 84.4/62.5 & 81.2/23.0 & 76.8/28.0 & 70.1/48.0 & 85.1/85.1 \\
\textbf{\method-quality} & 74.2/35.5 & 84.9/62.5 & 82.2/26.0 & 76.8/29.0 & 68.4/43.0 & 85.3/85.3 \\
\midrule
\multicolumn{7}{@{}l}{\emph{(c) LPIPS to vanilla / MAE$_\mathrm{B}$ to the input outside the edit masks (0--255)}} \\
Step skip        & 0.145/21.7 & 0.076/16.6 & 0.044/11.0 & 0.028/4.4  & 0.023/5.1  & 0.060 \\
RegionE          & 0.113/22.5 & 0.065/17.7 & 0.045/11.0 & 0.028/4.1  & 0.030/5.3  & 0.047 \\
TaylorSeer       & 0.365/37.3 & 0.303/26.9 & 0.246/23.9 & 0.102/11.3 & 0.134/16.8 & 0.283 \\
TeaCache         & 0.180/23.5 & 0.101/18.8 & 0.061/12.9 & 0.030/4.1  & 0.027/5.3  & 0.076 \\
RAS              & 0.216/23.5 & 0.135/18.7 & 0.098/14.4 & 0.056/7.0  & 0.062/9.4  & 0.109 \\
SpotEdit         & 0.150/20.9 & 0.108/13.3 & 0.090/9.9  & 0.038/2.9  & 0.043/3.4  & 0.077 \\
MICE$^\dagger$   & 0.330/18.6 & 0.199/14.1 & 0.150/10.8 & 0.056/5.4  & 0.064/6.6  & n/a \\
MIRAGE$^\ddagger$& 0.241/14.3 & 0.130/10.4 & 0.078/6.5  & 0.036/2.5  & 0.033/2.7  & 0.082 \\
\textbf{\method-fast}    & 0.222/24.4 & 0.046/19.9 & 0.037/13.7 & 0.053/5.6  & 0.061/7.3  & 0.042 \\
\textbf{\method-quality} & 0.204/26.0 & 0.034/22.4 & 0.034/15.0 & 0.058/8.2  & 0.061/9.8  & 0.041 \\
\bottomrule
\end{tabular*}
\caption{\textbf{Every method on every benchmark} (Klein 4B, 4 steps; VLM judge). Step skip's output is vanilla's at 2
steps, so only its speed is shown in (a). GEdit has no masks, so (c) shows LPIPS only. $^\dagger$Needs masks.
$^\ddagger$Includes VLM time.}
\label{tab:supp_perbench}\label{tab:supp_speed}\label{tab:supp_done}\label{tab:supp_lpips}
\end{table*}

\begin{table}[t]
\centering
\footnotesize
\setlength{\tabcolsep}{4pt}
\begin{tabular*}{\linewidth}{@{\extracolsep{\fill}}lrrrrr@{}}
\toprule
Edits per image & 5 & 6 & 7--8 & 9--12 & 13+ \\
\midrule
Global, 4 steps           & 14.7 & 10.4 & 21.0 & 13.9 & 14.1 \\
\textbf{\method}, 4 steps &  9.6 &  6.8 & 10.1 & 10.6 & 11.9 \\
Global, 8 steps           & 24.4 & 21.0 & 29.6 & 18.5 & 27.1 \\
\textbf{\method}, 8 steps & 12.5 &  9.6 & 12.3 & 12.1 & 16.2 \\
\bottomrule
\end{tabular*}
\caption{\textbf{Drop rate by edit density} (MICE-Bench; \% of vanilla's completed edits lost). Matched speed: 2.09
vs.\ 2.20$\times$ (4 steps), 3.46 vs.\ 3.51$\times$ (8 steps).}
\label{tab:supp_density}
\end{table}

\begin{table}[t]
\centering
\footnotesize
\setlength{\tabcolsep}{4pt}
\begin{tabular*}{\linewidth}{@{\extracolsep{\fill}}lrrrr@{}}
\toprule
Run & Step 1 & Step 2 & Step 3 & Step 4 \\
\midrule
Vanilla               & 4.66\,s & 4.67\,s & 4.68\,s       & 4.68\,s \\
\method-fast          & 4.72\,s & 1.84\,s & 0.57\,s (26\%) & 0.23\,s (8\%) \\
\method-quality       & 4.73\,s & 1.86\,s & 1.10\,s (58\%) & 1.83\,s \\
\bottomrule
\end{tabular*}
\caption{\textbf{Transformer time per step} (MICE-Bench, mean) and share of target tokens computed (100\% if not
shown). \method-quality recomputes the final step.}
\label{tab:supp_steps}
\end{table}

\begin{table}[t]
\centering
\footnotesize
\setlength{\tabcolsep}{4pt}
\begin{tabular*}{\linewidth}{@{\extracolsep{\fill}}lrrrr@{}}
\toprule
Task & Vanilla & Step skip & \method-fast & \method-q. \\
\midrule
background change & 98 & 98  & 95 & 98 \\
color alter       & 98 & 98  & 90 & 90 \\
material alter    & 65 & 60  & 57 & 57 \\
motion change     & 85 & 85  & 82 & 85 \\
ps human          & 71 & 70  & 70 & 76 \\
style change      & 95 & 93  & 95 & 97 \\
subject add       & 98 & 100 & 98 & 98 \\
subject remove    & 91 & 95  & 93 & 95 \\
subject replace   & 88 & 87  & 82 & 80 \\
text change       & 73 & 77  & 72 & 76 \\
tone transfer     & 95 & 95  & 95 & 95 \\
\bottomrule
\end{tabular*}
\caption{\textbf{GEdit-Bench by task} (606 images): edits done (\%).}
\label{tab:supp_gedit}
\end{table}

\begin{table}[t]
\centering
\footnotesize
\setlength{\tabcolsep}{4pt}
\begin{tabular*}{\linewidth}{@{\extracolsep{\fill}}lccc@{}}
\toprule
Run & Spd & PSNR & SSIM \\
\midrule
MICE, seeds 0/1/2  & 2.02/2.02/2.03$\times$ & 17.8/17.0/17.8 & -- \\
LoMOE, seeds 0/1/2 & 2.19/2.16/2.12$\times$ & 19.7/18.9/19.5 & -- \\
MICE, 2 steps$^\S$ & 2.67$\times$ & 19.3 & -- \\
LoMOE, 768$^2$ (16 images)          & 2.22$\times$ & 19.7 & -- \\
\midrule
PIE-Bench++~\cite{paralleledits}, 4 steps & 2.00$\times$ & 16.5 & 0.662 \\
PIE-Bench++, 2 steps                      & 1.42$\times$ & 17.7 & 0.705 \\
\bottomrule
\end{tabular*}
\caption{\textbf{Robustness} (Klein 4B, single-region engine). $^\S$Vs.\ 4-step vanilla. 768$^2$ uses exact chunked attention and a tiled VAE. PIE-Bench++: 700
images, compared with vanilla at the same number of steps.}
\label{tab:supp_robust}\label{tab:supp_piepp}
\end{table}

\begin{table}[t]
\centering
\footnotesize
\setlength{\tabcolsep}{4pt}
\begin{tabular*}{\linewidth}{@{\extracolsep{\fill}}lrrrr@{}}
\toprule
Setting & Spd & PSNR & SSIM & Fallback \\
\midrule
\multicolumn{5}{@{}l}{\emph{22 GEdit images}} \\
\method-fast ($\tau{=}1.0$)          & 2.17$\times$ & 21.02 & 0.806 & 0\% \\
+ full step 2 + fallback             & 1.83$\times$ & 20.60 & 0.816 & 41\% \\
+ full step 3 + fallback (quality)   & 1.81$\times$ & 20.65 & 0.814 & 41\% \\
\multicolumn{5}{@{}l}{\emph{12 MICE images}} \\
+ full step 2 + fallback             & 1.82$\times$ & 17.51 & 0.699 & 50\% \\
+ full step 3 + fallback             & 1.80$\times$ & 17.66 & 0.701 & 50\% \\
\bottomrule
\end{tabular*}
\caption{\textbf{Choosing \method-quality} on the tuning split; used unchanged everywhere else.}
\label{tab:supp_quality_tune}
\end{table}

\begin{table}[t]
\centering
\footnotesize
\setlength{\tabcolsep}{2.5pt}
\begin{tabular*}{\linewidth}{@{\extracolsep{\fill}}llrrrrrr@{}}
\toprule
Rule & $\tau$ & Spd & Done & All & Drop & Q & LPIPS \\
\midrule
Vanilla &  & 1.00$\times$ & 76.4 & 37.4 & -- & 3.44 & 0 \\
\midrule
Global & 0.3  & 2.56$\times$ & 74.3 & 35.1 & 151 & 3.28 & 0.268 \\
       & 0.6  & 3.23$\times$ & 62.8 & 25.6 & 359 & 2.92 & 0.325 \\
       & 0.72 & 3.46$\times$ & 61.0 & 25.2 & 384 & 2.93 & 0.320 \\
       & 1.0  & 3.88$\times$ & 67.1 & 28.2 & 269 & 3.07 & 0.296 \\
       & 1.5  & 4.27$\times$ & 73.4 & 34.4 & 157 & 3.19 & 0.286 \\
\midrule
\textbf{\method} & 0.3 & 2.61$\times$ & 73.9 & 32.4 & 154 & 3.27 & 0.271 \\
       & 0.6  & 3.51$\times$ & 70.9 & 31.7 & 200 & 3.11 & 0.291 \\
       & 1.0  & 4.14$\times$ & 72.6 & 33.6 & 169 & 3.21 & 0.285 \\
       & 1.5  & 4.29$\times$ & 73.3 & 33.6 & 153 & 3.19 & 0.284 \\
\midrule
GT masks & 0.6 & 3.41$\times$ & 69.0 & 27.5 & 226 & 3.09 & 0.299 \\
Argmax & 0.6 & 3.44$\times$ & 70.3 & 31.7 & 206 & 3.11 & 0.292 \\
AB2 update & 0.6 & 3.39$\times$ & 71.2 & 30.9 & 195 & 3.11 & 0.297 \\
\bottomrule
\end{tabular*}
\caption{\textbf{Threshold sweep at 8 steps} (MICE-Bench; Done and All in \%; Q: judge's 1--5 rating).}
\label{tab:supp_sweep8}
\end{table}

\begin{table}[t]
\centering
\footnotesize
\setlength{\tabcolsep}{3pt}
\begin{tabular*}{\linewidth}{@{\extracolsep{\fill}}lrrrrrrr@{}}
\toprule
Trajectories & Imgs & Broken & Spread & $>$2$\times$ & $S_\mathrm{glob}$ & $S_\mathrm{pc}$ & Gain \\
\midrule
MICE, 4 st.  & 262 & 89.3 & 1.57 & 83 & 24.4 & 34.0 & +9.6 \\
MICE, 8 st.  & 262 & 60.3 & 1.49 & 74 & 42.5 & 53.3 & +10.8 \\
LoMOE, 4 st. &  64 & 62.5 & 1.32 & 14 & 22.4 & 29.6 & +7.1 \\
LoMOE, 8 st. &  64 & 29.7 & 1.36 & 13 & 42.3 & 51.9 & +9.6 \\
\midrule
MICE, Base   & 100 & 27.0 & 1.60 & 23 & 56.3 & 63.6 & +7.3 \\
LoMOE, Base  &  64 & 29.7 & 1.37 & 14 & 51.4 & 57.2 & +5.8 \\
\bottomrule
\end{tabular*}
\caption{\textbf{Premise of Theorem~1} on vanilla trajectories (Broken, $S$, Gain in \%; Spread: $\beta$ spread, $\times$; $>$2$\times$: images with spread above 2$\times$). Budget: 5\% of each edit's size (summed distance of
vanilla's final estimate from the source over the edit's tokens); target saving 60\%. Base: Klein Base, first 100 MICE images; with many small steps fewer images break.}
\label{tab:supp_e1}
\end{table}

\begin{table}[t]
\centering
\footnotesize
\setlength{\tabcolsep}{3pt}
\begin{tabular*}{\linewidth}{@{\extracolsep{\fill}}lrrrrrr@{}}
\toprule
 & \multicolumn{3}{c}{MICE} & \multicolumn{3}{c}{LoMOE} \\
\cmidrule(lr){2-4}\cmidrule(lr){5-7}
Variant & Rows & Spd & PSNR & Rows & Spd & PSNR \\
\midrule
Skipping only            & 0.88 & 1.09$\times$ & 21.2 & 0.84 & 1.13$\times$ & 24.4 \\
+ ref.\ cache & 0.58 & 1.48$\times$ & 19.1 & 0.54 & 1.52$\times$ & 21.4 \\
+ text cache             & 0.43 & 1.76$\times$ & 18.2 & 0.39 & 1.87$\times$ & 20.3 \\
+ stored K/V & 0.43 & 2.02$\times$ & 17.8 & 0.39 & 2.19$\times$ & 19.7 \\
\midrule
Refresh 2nd step & 0.53 & 1.72$\times$ & 18.1 & 0.49 & 1.81$\times$ & 19.9 \\
No ref.\ cache & 0.73 & 1.32$\times$ & 18.2 & 0.69 & 1.39$\times$ & 20.1 \\
\bottomrule
\end{tabular*}
\caption{\textbf{Engine ablation} (single-region variant, 4 steps): share of rows computed, speed and PSNR.}
\label{tab:supp_engine}
\end{table}

\begin{table*}[t]
\centering
\footnotesize
\setlength{\tabcolsep}{2.6pt}
\renewcommand{\arraystretch}{0.92}
\begin{tabular*}{\textwidth}{@{\extracolsep{\fill}}llrrrrrrrrrrrrrrr@{}}
\toprule
 & & \multicolumn{5}{c}{Global threshold} & \multicolumn{5}{c}{\method (self-grounded regions)} & \multicolumn{5}{c}{Per-part, GT masks} \\
\cmidrule(lr){3-7}\cmidrule(lr){8-12}\cmidrule(lr){13-17}
Set & $\tau$ & Spd & All & Drop & CLIP & Worst & Spd & All & Drop & CLIP & Worst & Spd & All & Drop & CLIP & Worst \\
\midrule
MICE, 8 st. & 0.3  & 2.56 & 34.0 & 163 & 27.22 & 22.33 & 2.62 & 37.0 & 159 & 27.24 & 22.32 & 2.67 & 34.4 & 158 & 27.24 & 22.33 \\
            & 0.6  & 3.24 & 23.7 & 357 & 25.90 & 21.16 & 3.52 & 32.1 & 235 & 26.75 & 21.94 & 3.42 & 29.4 & 226 & 26.77 & 21.77 \\
            & 0.72 & 3.46 & 24.4 & 367 & 25.91 & 21.17 & -- & -- & -- & -- & -- & -- & -- & -- & -- & -- \\
            & 1.0  & 3.88 & 27.9 & 276 & 26.50 & 21.64 & 4.15 & 32.4 & 215 & 26.96 & 22.06 & 4.07 & 32.1 & 174 & 27.09 & 22.17 \\
            & 1.5  & 4.27 & 33.2 & 187 & 27.00 & 22.15 & 4.29 & 35.9 & 174 & 27.07 & 22.25 & 4.28 & 35.1 & 172 & 27.11 & 22.29 \\
\midrule
LoMOE, 8 st.& 0.3  & 2.58 & 53.1 & 19 & 28.00 & 25.62 & 2.67 & 56.2 & 18 & 28.06 & 25.63 & 2.61 & 54.7 & 19 & 28.04 & 25.63 \\
            & 0.6  & 3.18 & 48.4 & 33 & 27.14 & 24.56 & 3.55 & 59.4 & 15 & 27.93 & 25.61 & 3.43 & 62.5 & 11 & 28.10 & 25.74 \\
            & 1.0  & 3.79 & 50.0 & 23 & 27.20 & 24.82 & 4.05 & 62.5 & 11 & 28.03 & 25.66 & 4.10 & 62.5 &  8 & 28.21 & 25.88 \\
            & 1.5  & 4.09 & 60.9 & 13 & 28.16 & 25.80 & 4.24 & 60.9 & 12 & 28.19 & 25.81 & 4.18 & 60.9 & 11 & 28.26 & 26.03 \\
\midrule
MIRA, 8 st. & 0.6  & 3.17 & 17.0 & 37 & 26.46 & 22.07 & 3.45 & 19.0 & 32 & 26.57 & 22.08 & & & & & \\
            & 1.0  & 3.85 & 15.0 & 39 & 26.37 & 21.88 & 4.14 & 19.0 & 27 & 26.55 & 22.01 & & & & & \\
InfoDet, 8 st. & 0.6 & 2.94 & 0.0 & 92  & 25.89 & 22.30 & 2.97 & 0.0 & 103 & 25.76 & 22.09 & & & & & \\
            & 1.0  & 3.63 & 1.0 & 104 & 25.80 & 21.98 & 3.85 & 1.0 & 91  & 26.04 & 22.13 & & & & & \\
Crello, 8 st. & 0.6 & 3.21 & 6.0 & 64 & 24.99 & 22.89 & 3.32 & 9.0 & 57 & 25.20 & 23.00 & & & & & \\
            & 1.0  & 3.82 & 8.0 & 58 & 24.73 & 22.68 & 4.07 & 7.0 & 60 & 25.25 & 23.13 & & & & & \\
\midrule
MICE, 4 st. & 0.6  & 1.93 & 34.0 & 167 & 27.18 & 22.31 & 1.97 & 33.6 & 155 & 27.20 & 22.28 & 1.98 & 34.7 & 159 & 27.18 & 22.25 \\
            & 1.0  & 2.10 & 29.4 & 233 & 26.65 & 21.82 & 2.20 & 33.6 & 181 & 27.10 & 22.20 & 2.20 & 34.0 & 156 & 27.19 & 22.30 \\
\bottomrule
\end{tabular*}
\caption{\textbf{CLIP check, every rule and benchmark.} Spd: speed-up ($\times$); All: images with every edit applied
(\%); Drop: edits the CLIP check counts as dropped; CLIP / Worst: mean and worst-object local CLIP. Separate runs, so
speeds can differ from \cref{tab:supp_sweep8} by 0.01$\times$. GT masks: upper reference.}
\label{tab:supp_clip}
\end{table*}

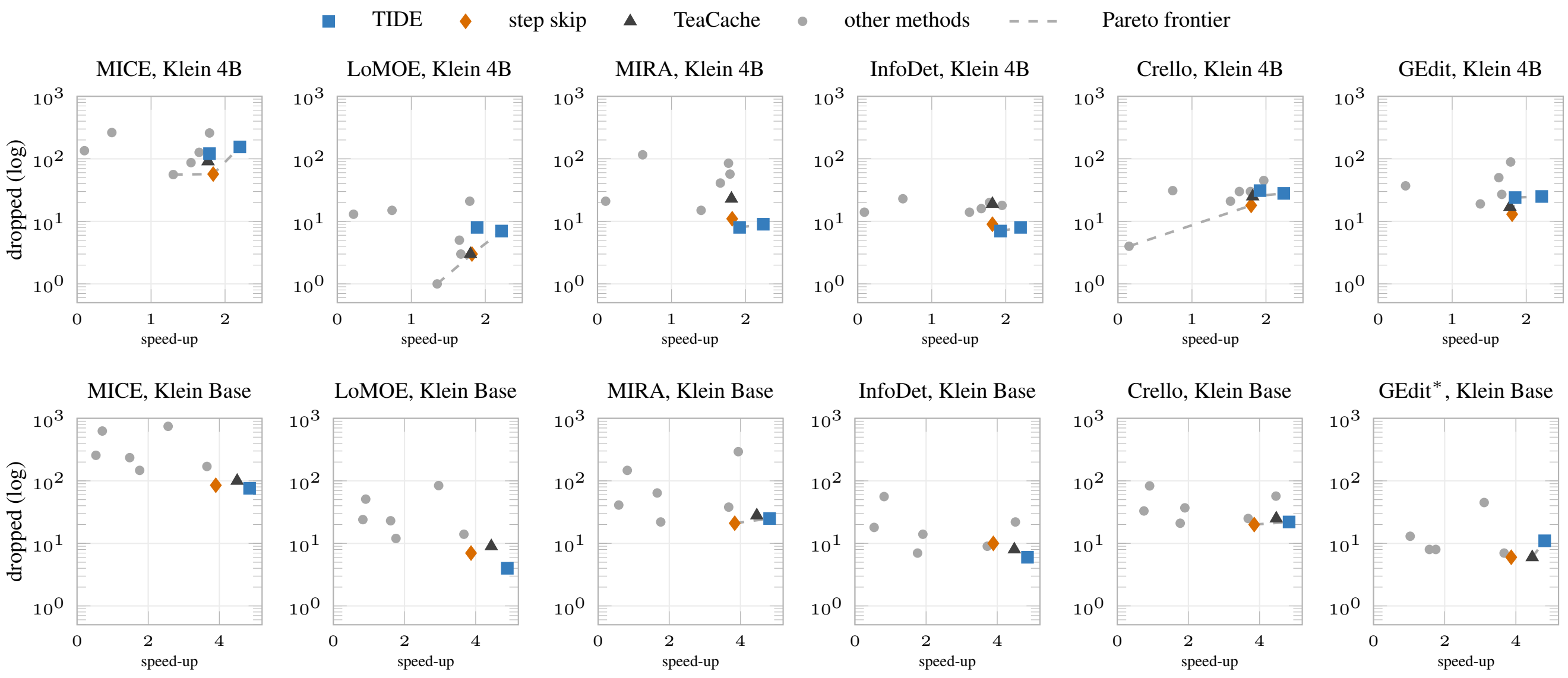
\begin{figure*}[t]
\centering
\pgfplotsset{
  /pgfplots/pfaxis/.style={scale only axis, width=0.118\textwidth, height=2.3cm, ymin=0.5, ymax=1000, xmin=0,
    grid=major, grid style={gray!15}, axis line style={gray!60}, tick style={gray!60},
    tick label style={font=\tiny}, label style={font=\scriptsize}, title style={font=\scriptsize, yshift=-4pt},
    xlabel={speed-up}, xlabel style={font=\tiny, yshift=4pt}, ylabel style={yshift=-5pt}, clip=false},
  pfront/.style={gray!65, dashed, thick},
  pother/.style={only marks, mark=*, black!35, mark size=1.3pt},
  pskip/.style={only marks, mark=diamond*, orange!85!black, mark size=2.2pt},
  ptea/.style={only marks, mark=triangle*, black!75, mark size=2.1pt},
  ptide/.style={only marks, mark=square*, cvprblue, mark size=1.8pt}}
\begin{tikzpicture}
\begin{axis}[hide axis, scale only axis, width=1pt, height=1pt, xmin=0, xmax=1, ymin=0, ymax=1,
  legend columns=5, legend style={font=\scriptsize, draw=none, column sep=10pt, at={(0.5,0.5)}, anchor=center}]
\addlegendimage{ptide}  \addlegendentry{\method}
\addlegendimage{pskip}  \addlegendentry{step skip}
\addlegendimage{ptea}   \addlegendentry{TeaCache}
\addlegendimage{pother} \addlegendentry{other methods}
\addlegendimage{pfront} \addlegendentry{Pareto frontier}
\end{axis}
\end{tikzpicture}\\[1pt]
\begin{tikzpicture}\begin{axis}[pfaxis, ymode=log, title={MICE, Klein 4B}, xmax=2.5, ylabel={dropped (log)}]
\addplot[pfront] coordinates {(1.3,56.0) (1.84,57.0) (2.2,155.0)};
\addplot[pother] coordinates {(1.3,56.0) (1.79,259.0) (1.65,127.0) (1.54,87.0) (0.47,263.0) (0.1,135.0)};
\addplot[pskip] coordinates {(1.84,57.0)};
\addplot[ptea] coordinates {(1.77,91.0)};
\addplot[ptide] coordinates {(2.2,155.0) (1.79,121.0)};
\end{axis}\end{tikzpicture}\hfill
\begin{tikzpicture}\begin{axis}[pfaxis, ymode=log, title={LoMOE, Klein 4B}, xmax=2.5]
\addplot[pfront] coordinates {(1.35,1.0) (1.82,3.0) (2.22,7.0)};
\addplot[pother] coordinates {(1.35,1.0) (1.79,21.0) (1.65,5.0) (1.67,3.0) (0.74,15.0) (0.22,13.0)};
\addplot[pskip] coordinates {(1.82,3.0)};
\addplot[ptea] coordinates {(1.8,3.0)};
\addplot[ptide] coordinates {(2.22,7.0) (1.89,8.0)};
\end{axis}\end{tikzpicture}\hfill
\begin{tikzpicture}\begin{axis}[pfaxis, ymode=log, title={MIRA, Klein 4B}, xmax=2.5]
\addplot[pfront] coordinates {(1.92,8.0) (2.24,9.0)};
\addplot[pother] coordinates {(1.4,15.0) (1.79,57.0) (1.66,41.0) (1.77,85.0) (0.61,116.0) (0.11,21.0)};
\addplot[pskip] coordinates {(1.82,11.0)};
\addplot[ptea] coordinates {(1.81,23.0)};
\addplot[ptide] coordinates {(2.24,9.0) (1.92,8.0)};
\end{axis}\end{tikzpicture}\hfill
\begin{tikzpicture}\begin{axis}[pfaxis, ymode=log, title={InfoDet, Klein 4B}, xmax=2.5]
\addplot[pfront] coordinates {(1.93,7.0) (2.2,8.0)};
\addplot[pother] coordinates {(1.51,14.0) (1.78,20.0) (1.67,16.0) (1.95,18.0) (0.61,23.0) (0.09,14.0)};
\addplot[pskip] coordinates {(1.82,9.0)};
\addplot[ptea] coordinates {(1.82,19.0)};
\addplot[ptide] coordinates {(2.2,8.0) (1.93,7.0)};
\end{axis}\end{tikzpicture}\hfill
\begin{tikzpicture}\begin{axis}[pfaxis, ymode=log, title={Crello, Klein 4B}, xmax=2.5]
\addplot[pfront] coordinates {(0.15,4.0) (1.8,18.0) (1.82,25.0) (2.24,28.0)};
\addplot[pother] coordinates {(1.52,21.0) (1.79,30.0) (1.64,30.0) (1.97,45.0) (0.74,31.0) (0.15,4.0)};
\addplot[pskip] coordinates {(1.8,18.0)};
\addplot[ptea] coordinates {(1.82,25.0)};
\addplot[ptide] coordinates {(2.24,28.0) (1.92,31.0)};
\end{axis}\end{tikzpicture}\hfill
\begin{tikzpicture}\begin{axis}[pfaxis, ymode=log, title={GEdit, Klein 4B}, xmax=2.5]
\addplot[pfront] coordinates {(1.81,13.0) (1.85,24.0) (2.21,25.0)};
\addplot[pother] coordinates {(1.38,19.0) (1.79,89.0) (1.67,27.0) (1.63,50.0) (0.37,37.0)};
\addplot[pskip] coordinates {(1.81,13.0)};
\addplot[ptea] coordinates {(1.78,17.0)};
\addplot[ptide] coordinates {(2.21,25.0) (1.85,24.0)};
\end{axis}\end{tikzpicture}\\[3pt]
\begin{tikzpicture}\begin{axis}[pfaxis, ymode=log, title={MICE, Klein Base}, xmax=5.2, ylabel={dropped (log)}]
\addplot[pfront] coordinates {(4.85,76.0)};
\addplot[pother] coordinates {(1.48,235.0) (3.65,170.0) (1.76,147.0) (2.56,744.0) (0.71,626.0) (0.53,256.0)};
\addplot[pskip] coordinates {(3.9,85.0)};
\addplot[ptea] coordinates {(4.5,100.0)};
\addplot[ptide] coordinates {(4.85,76.0)};
\end{axis}\end{tikzpicture}\hfill
\begin{tikzpicture}\begin{axis}[pfaxis, ymode=log, title={LoMOE, Klein Base}, xmax=5.2]
\addplot[pfront] coordinates {(4.89,4.0)};
\addplot[pother] coordinates {(1.61,23.0) (3.67,14.0) (1.76,12.0) (2.96,84.0) (0.91,51.0) (0.83,24.0)};
\addplot[pskip] coordinates {(3.87,7.0)};
\addplot[ptea] coordinates {(4.44,9.0)};
\addplot[ptide] coordinates {(4.89,4.0)};
\end{axis}\end{tikzpicture}\hfill
\begin{tikzpicture}\begin{axis}[pfaxis, ymode=log, title={MIRA, Klein Base}, xmax=5.2]
\addplot[pfront] coordinates {(3.84,21.0) (4.82,25.0)};
\addplot[pother] coordinates {(1.66,64.0) (3.67,38.0) (1.76,22.0) (3.94,293.0) (0.82,147.0) (0.58,41.0)};
\addplot[pskip] coordinates {(3.84,21.0)};
\addplot[ptea] coordinates {(4.46,28.0)};
\addplot[ptide] coordinates {(4.82,25.0)};
\end{axis}\end{tikzpicture}\hfill
\begin{tikzpicture}\begin{axis}[pfaxis, ymode=log, title={InfoDet, Klein Base}, xmax=5.2]
\addplot[pfront] coordinates {(4.85,6.0)};
\addplot[pother] coordinates {(1.91,14.0) (3.72,9.0) (1.76,7.0) (4.51,22.0) (0.82,56.0) (0.54,18.0)};
\addplot[pskip] coordinates {(3.89,10.0)};
\addplot[ptea] coordinates {(4.48,8.0)};
\addplot[ptide] coordinates {(4.85,6.0)};
\end{axis}\end{tikzpicture}\hfill
\begin{tikzpicture}\begin{axis}[pfaxis, ymode=log, title={Crello, Klein Base}, xmax=5.2]
\addplot[pfront] coordinates {(3.85,20.0) (4.83,22.0)};
\addplot[pother] coordinates {(1.9,37.0) (3.68,25.0) (1.77,21.0) (4.46,57.0) (0.91,83.0) (0.75,33.0)};
\addplot[pskip] coordinates {(3.85,20.0)};
\addplot[ptea] coordinates {(4.47,25.0)};
\addplot[ptide] coordinates {(4.83,22.0)};
\end{axis}\end{tikzpicture}\hfill
\begin{tikzpicture}\begin{axis}[pfaxis, ymode=log, title={GEdit$^*$, Klein Base}, xmax=5.2]
\addplot[pfront] coordinates {(4.46,6.0) (4.81,11.0)};
\addplot[pother] coordinates {(1.57,8.0) (3.67,7.0) (1.75,8.0) (3.11,45.0) (1.03,13.0)};
\addplot[pskip] coordinates {(3.87,6.0)};
\addplot[ptea] coordinates {(4.46,6.0)};
\addplot[ptide] coordinates {(4.81,11.0)};
\end{axis}\end{tikzpicture}
\caption{\textbf{Speed vs.\ dropped edits on every benchmark} (lower right is better). \method on Klein 4B: fast and quality. $^*$First 100 images.}
\label{fig:supp_pareto_all}
\end{figure*}

\paragraph{Frontier.}
A setting is on the frontier (\cref{fig:supp_pareto_all}) if no other setting on the same images, seed, GPU and vanilla
reference is at least as fast and drops no more edits, one of the two strictly. Two settings are \emph{tied} if their
speeds differ by less than 3\% (the spread of repeated timings) and their drops do not differ significantly (exact
McNemar $p>0.05$). On Klein Base (first 100 MICE images), \method on 14 steps (5.23$\times$, 39 drops) and \method +
TeaCache 0.6 (4.94$\times$, 38) tie with TeaCache at its fastest (4.50$\times$, 35; $p{=}0.62$ and $0.75$), all below
the seed noise floor of 53--55, and keep \method's image quality (\cref{tab:supp_frontq}). On Klein 4B, no setting
within the noise floor (96--101) is faster than step skip (1.84$\times$, 57).

\begin{table}[t]
\centering
\footnotesize
\setlength{\tabcolsep}{3.5pt}
\begin{tabular*}{\linewidth}{@{\extracolsep{\fill}}lrrccc@{}}
\toprule
MICE (100 imgs) & Spd & Drop & CLIP-IQA$\uparrow$ & MUSIQ$\uparrow$ & LPIPS$\downarrow$ \\
\midrule
Vanilla (28 steps)       & 1.00$\times$ & -- & 0.718 & 75.5 & 0 \\
Step skip (14 steps)     & 2.00$\times$ & 25 & 0.717 & 75.5 & 0.062 \\
TeaCache                 & 4.50$\times$ & 35 & 0.706 & 74.9 & 0.191 \\
\method                  & 4.85$\times$ & 46 & 0.685 & 74.2 & 0.266 \\
\method + TeaCache 0.6   & 4.94$\times$ & 38 & 0.683 & 74.3 & 0.266 \\
\method on 14 steps      & 5.23$\times$ & 39 & 0.685 & 74.4 & 0.246 \\
\bottomrule
\end{tabular*}
\caption{\textbf{Quality at the frontier points} (speed and Drop as in \cref{tab:supp_stack}; CLIP-IQA and MUSIQ as in
\cref{tab:supp_quality}). Stacking adds speed without lowering \method's quality, which stays below TeaCache's.}
\label{tab:supp_frontq}
\end{table}

\paragraph{Further speed--drop points.}
\Cref{tab:supp_more} lists the runs behind Fig.~3 of the main paper and two further checks. On Klein 4B, step skip has
settings only at 3, 2 and 1 of 4 steps; between 2 steps (1.84$\times$, 57 drops) and 1 step (3.10$\times$, 238) only
\method and the global threshold run, and \method drops fewer edits (155 vs.\ 237 at 2.20 vs.\ 2.09$\times$; 140 vs.\
218 at 2.29 vs.\ 2.24$\times$, $p{<}10^{-11}$). Inside RegionE's own engine, which finishes one edited/unedited split,
the relative rule does not help: per-part and global thresholds tie at matched speed, and neither beats RegionE's
cosine threshold (56 at 1.30$\times$). The gain of per-part budgets thus comes with \method's token-level freezing,
not with the threshold rule alone. On Klein Base, moving the first decision from after step 3 to after step 6 or 9, after the
parts' drift scales separate (\cref{fig:supp_spread}), slows \method to 4.04 and 3.46$\times$ and still does not
separate the two rules ($p\ge0.45$).

\begin{table}[t]
\centering
\footnotesize
\setlength{\tabcolsep}{3.5pt}
\begin{tabular*}{\linewidth}{@{\extracolsep{\fill}}lrrrrr@{}}
\toprule
 & \multicolumn{2}{c}{A} & \multicolumn{2}{c}{B} & \\
\cmidrule(lr){2-3}\cmidrule(lr){4-5}
Setting & Spd & Drop & Spd & Drop & $p$ \\
\midrule
\multicolumn{6}{@{}l}{\emph{(a) Klein 4B; A: step skip, B: TeaCache}} \\
3 of 4 steps; $\delta{=}0.25$ & 1.29$\times$ & 49  & 1.04$\times$ & 23 & \\
2 of 4 steps; $\delta{=}0.4$  & 1.84$\times$ & 57  & 1.29$\times$ & 57 & \\
1 of 4 steps; $\delta{=}0.6$  & 3.10$\times$ & 238 & 1.34$\times$ & 78 & \\
\phantom{1 of 4 steps;} $\delta{=}0.8$ & & & 1.77$\times$ & 91 & \\
\midrule
\multicolumn{6}{@{}l}{\emph{(b) Klein 4B; A: global, B: \method}} \\
$\tau{=}1.5$ & 2.24$\times$ & 218 & 2.29$\times$ & 140 & $5{\times}10^{-12}$ \\
\midrule
\multicolumn{6}{@{}l}{\emph{(c) RegionE engine; A: global, B: per-part}} \\
official (cosine 0.93) & 1.30$\times$ & 56 & & & \\
$\tau{=}0.6$ & 1.33$\times$ & 66 & 1.33$\times$ & 65  & 1.00 \\
$\tau{=}1.0$ & 1.36$\times$ & 79 & 1.39$\times$ & 81  & 0.88 \\
$\tau{=}1.5$ & 1.41$\times$ & 80 & 1.52$\times$ & 109 & $8{\times}10^{-4}$ \\
\midrule
\multicolumn{6}{@{}l}{\emph{(d) Klein Base, 100 images, $\tau{=}1.5$; A: global, B: per-part}} \\
first decision after step 3 & 4.83$\times$ & 44 & 4.85$\times$ & 46 & 0.75 \\
first decision after step 6 & 3.97$\times$ & 39 & 4.04$\times$ & 36 & 0.45 \\
first decision after step 9 & 3.42$\times$ & 45 & 3.46$\times$ & 43 & 0.73 \\
\bottomrule
\end{tabular*}
\caption{\textbf{Further speed--drop points} (MICE-Bench; Klein 4B: 262 images vs.\ 4-step vanilla). (a), (b): drawn in
Fig.~3 of the main paper. (c): RegionE's official code with its cosine threshold replaced by the relative rule (parts
from the masks). $p$: exact McNemar, B vs.\ A.}
\label{tab:supp_more}
\end{table}

\paragraph{Does the rule follow the theory?}
Theorem~1 and the premise check (\cref{tab:supp_e1}) use oracle breaking scores, while Rule (4) only prices each part
by its mean drift. On the 4-step editor this price ranks the parts like the oracle (\cref{tab:supp_rule}; Spearman 0.61
on MICE, positive in 92\% of images), at 8 steps less so (0.28), and on Klein Base not at all (0.06). Neither
freeze-only rule keeps the strict 5\% budgets at a 60\% saving, so the theorems explain \emph{which} edits a global
threshold drops (small ones first, main paper, Tab.~2), not a guarantee for the rule. Klein Base differs because its
parts move alike until the first decision and separate only afterwards (\cref{fig:supp_spread}).

\paragraph{Does the shared-shape assumption of Prop.~D hold?}
Only roughly (\cref{tab:supp_eta}): with each part's drifts divided by their mean and compared with the image's pooled
shape at the first decision, the median $\eta_k$ is 0.17 on MICE and 0.10--0.11 on LoMOE, so Rule (4) can exceed a
part's budget by about a sixth of its own motion. $\eta_k$ is the same at 4 and 8 steps and on Klein Base, so it does
not explain the falling rank correlation; the parts' scales changing after the first decision does.

\begin{table*}[t]
\centering
\footnotesize
\setlength{\tabcolsep}{4pt}
\begin{tabular*}{\textwidth}{@{\extracolsep{\fill}}lrrrrrrrrr@{}}
\toprule
 & \multicolumn{3}{c}{Price $m_k$ vs.\ oracle $\beta_k$} & \multicolumn{2}{c}{Edits over budget} & \multicolumn{4}{c}{Shared drift shape (Prop.~D)} \\
\cmidrule(lr){2-4}\cmidrule(lr){5-6}\cmidrule(lr){7-10}
Trajectories & Imgs & Spearman & $>0$ & Rule (4) & Global & Parts & Median $\eta_k$ & 90th pct. & $\eta_k\le0.1$ \\
\midrule
MICE, Klein 4B, 4 steps     & 260 & 0.61 & 92\% & 100\% & 97\% & 1966 & 0.17 & 0.33 & 20\% \\
LoMOE, Klein 4B, 4 steps    &  37 & 0.71 & 95\% & 100\% & 95\% &  192 & 0.11 & 0.21 & 44\% \\
MICE, Klein 4B, 8 steps     & 260 & 0.28 & 72\% & 100\% & 95\% & 1966 & 0.17 & 0.34 & 18\% \\
LoMOE, Klein 4B, 8 steps    &  37 & 0.29 & 70\% &  99\% & 85\% &  192 & 0.11 & 0.23 & 45\% \\
MICE, Klein Base, 28 steps  &  99 & 0.06 & 56\% &  90\% & 90\% &  702 & 0.16 & 0.32 & 21\% \\
LoMOE, Klein Base, 28 steps &  37 & 0.09 & 57\% &  87\% & 83\% &  192 & 0.10 & 0.21 & 48\% \\
\bottomrule
\end{tabular*}
\caption{\textbf{Does the rule follow the theory?} Vanilla trajectories, ground-truth parts, budgets of
\cref{tab:supp_e1}. Spearman: rank correlation across an image's parts (at least 3 with finite $\beta_k$) between the
price of Rule (4), the mean drift $m_k$ at the first decision, and the oracle breaking score $\beta_k$; $>0$: images
with positive correlation. Over budget: freeze-only rules, $\tau$ set per image for a 60\% saving.
$\eta_k=\sup_x|G_k(x)-G(x)|$ at the first decision, $G$ pooled over the image's edited tokens.}
\label{tab:supp_rule}\label{tab:supp_eta}
\end{table*}

\paragraph{Coarse regions.}
\Cref{tab:supp_regstrat} sorts the edits by how well \method's region matches the part (best IoU with any region), by
whether the part's noun also names another part (two cats), and by whether its mask overlaps another part's. \method
drops more edits when its region is poor (10.7\% in the lowest IoU quarter, 6.6\% in the highest), but fewer than one
global threshold in every IoU quarter (12.2--17.7\%). Parts sharing a noun, where attention cannot tell them apart, are
dropped more often (10.4\% vs.\ 7.4\%), and on overlapping masks the two rules tie (9.4\%). This agrees with Prop.~C:
mis-assigned tokens add error only while they are frozen, so poor regions cost some edits but do not undo the gain of
per-part budgets.
\paragraph{Klein Base: further results.}
\Cref{tab:supp_base_abl,tab:supp_bsteps,tab:supp_brobust} complete the many-step results of the main paper (Sec.~4.4).
\emph{Ablations} (\cref{tab:supp_base_abl}, left): on MICE and LoMOE alike, the caches give most of the speed
(3.16--3.18$\times$ without them), and freezing on top raises caching alone from about 2.5 to 4.9$\times$ for a few
more drops. \emph{Per-part vs.\ global} (right): with checks every 8 steps, per-part thresholds drop fewer edits on MICE
at every $\tau$; on the four other benchmarks the two rules tie, because both revisit every frozen token and fewer
images break a per-edit budget than on the 4-step editor (\cref{tab:supp_e1}). \emph{Fewer steps}
(\cref{tab:supp_bsteps}): \method on 14 steps runs at 5.19--5.23$\times$, faster than \method on 28 steps on every
set, with lower LPIPS on five of seven and fewer drops on five (\eg MIRA: 19 vs.\ 25), but more on InfoDet (9 vs.\ 6) and
Crello (28 vs.\ 22). Step skip stays closest to vanilla at lower speed (7 steps: 3.84--3.91$\times$), and switching
off guidance drops many edits (\eg 140 on MIRA). Speed and LPIPS are stable across seeds and at 768$^2$
(\cref{tab:supp_brobust}).

\begin{table*}[t]
\centering
\footnotesize
\setlength{\tabcolsep}{2.6pt}
\renewcommand{\arraystretch}{0.94}
\begin{minipage}[t]{0.55\textwidth}
\vspace{0pt}
\begin{tabular*}{\linewidth}{@{\extracolsep{\fill}}lrrrrrr@{}}
\toprule
 & \multicolumn{3}{c}{MICE (100)} & \multicolumn{3}{c}{LoMOE (64)} \\
\cmidrule(lr){2-4}\cmidrule(lr){5-7}
Variant & Spd & Done & Drop & Spd & Done & Drop \\
\midrule
\textbf{\method (full)}        & 4.85 & 70.6 & 46  & 4.89 & 80.7 & 4 \\
No wake-up checks              & 7.74 & 20.8 & 360 &      &      &   \\
Checks every 14 steps          & 5.95 & 63.6 & 79  &      &      &   \\
One pass (no re-measure)       & 2.90 & 72.8 & 32  &      &      &   \\
No text / reference cache      & 3.18 & 65.6 & 46  & 3.16 & 71.9 & 12 \\
No text cache                  & 4.14 & 69.8 & 39  & 4.11 & 77.1 & 8 \\
Ref./text refreshed every 2nd  & 4.06 & 69.9 & 44  & 4.02 & 77.6 & 6 \\
Caches only ($\tau{=}0$)       & 2.51 & 72.3 & 27  & 2.48 & 81.2 & 3 \\
Hard (argmax) regions          & 4.86 & 70.9 & 46  &      &      &   \\
Ground-truth masks             & 4.86 & 70.3 & 47  &      &      &   \\
AB2 extrapolation              & 4.86 & 71.7 & 36  &      &      &   \\
Background from source         & 4.85 & 59.3 & 103 &      &      &   \\
\bottomrule
\end{tabular*}
\end{minipage}\hfill
\begin{minipage}[t]{0.42\textwidth}
\vspace{0pt}
\begin{tabular*}{\linewidth}{@{\extracolsep{\fill}}llrrrrrr@{}}
\toprule
 & & \multicolumn{3}{c}{Per-part} & \multicolumn{3}{c}{Global} \\
\cmidrule(lr){3-5}\cmidrule(lr){6-8}
 & $\tau$ & Spd & Drop & LP & Spd & Drop & LP \\
\midrule
MICE    & 0.6 & 4.35 & 70  &       & 4.35 & 116 & \\
(262)   & 1.5 & 4.85 & 76  &       & 4.83 & 120 & \\
        & 2.5 & 4.92 & 74  &       & 4.89 & 125 & \\
        & 3.0 & 4.93 & 76  &       & 4.92 & 121 & \\
\multicolumn{2}{@{}l}{\scriptsize no checks, 1.5$^*$}  & 7.74 & 360 & & 7.73 & 361 & \\
\multicolumn{2}{@{}l}{\scriptsize no checks, 0.6$^*$}  & 7.34 & 353 & & 7.35 & 353 & \\
\multicolumn{2}{@{}l}{\scriptsize every 14, 1.5$^*$}   & 5.95 & 79  & & 5.92 & 79  & \\
\midrule
LoMOE   & 0.6 & 4.34 & 4  & .185 & 4.33 & 4  & .185 \\
        & 1.5 & 4.81 & 4  & .187 & 4.79 & 4  & .187 \\
        & 2.5 & 4.88 & 4  & .187 & 4.86 & 5  & .187 \\
        & 3.0 & 4.89 & 6  & .187 & 4.88 & 6  & .187 \\
MIRA    & 1.5 & 4.82 & 25 & .109 & 4.80 & 22 & .108 \\
        & 2.5 & 4.88 & 24 & .109 & 4.84 & 28 & .109 \\
InfoDet & 1.5 & 4.85 & 6  & .032 & 4.82 & 8  & .038 \\
        & 2.5 & 4.89 & 7  & .038 & 4.88 & 9  & .039 \\
Crello  & 1.5 & 4.83 & 22 & .044 & 4.81 & 23 & .044 \\
        & 2.5 & 4.88 & 22 & .044 & 4.86 & 23 & .044 \\
\bottomrule
\end{tabular*}
\end{minipage}
\caption{\textbf{Klein Base 4B} (28 steps; Spd in $\times$, Done in \%, LP: LPIPS). Left: ablations, one component
changed at a time. Right: per-part vs.\ one global threshold, same engine, regions and checks every 8 steps;
$^*$100 images, decisions held longer.}
\label{tab:supp_base_abl}\label{tab:supp_babl}\label{tab:supp_bclaim}
\end{table*}

\begin{table*}[t]
\centering
\footnotesize
\setlength{\tabcolsep}{2.8pt}
\renewcommand{\arraystretch}{0.94}
\begin{tabular*}{\textwidth}{@{\extracolsep{\fill}}lrrrrrrrrrrrrrrr@{}}
\toprule
 & \multicolumn{3}{c}{Step skip, 14 steps} & \multicolumn{3}{c}{Step skip, 7 steps} & \multicolumn{3}{c}{No guidance}
 & \multicolumn{3}{c}{\textbf{\method}} & \multicolumn{3}{c}{\textbf{\method}, 14 steps} \\
\cmidrule(lr){2-4}\cmidrule(lr){5-7}\cmidrule(lr){8-10}\cmidrule(lr){11-13}\cmidrule(lr){14-16}
Benchmark & Spd & Drop & LP & Spd & Drop & LP & Spd & Drop & LP & Spd & Drop & LP & Spd & Drop & LP \\
\midrule
MICE$^*$  & 2.00 & 25 & .062 & 3.91 & 36 & .143 & 2.01 & 132 & .450 & 4.85 & 46 & .266 & 5.23 & 39 & .246 \\
LoMOE     & 1.98 & 10 & .035 & 3.87 &  7 & .101 & 1.99 &  44 & .347 & 4.89 &  4 & .187 & 5.19 &  3 & .176 \\
MIRA      & 1.98 & 10 & .023 & 3.84 & 21 & .068 & 1.98 & 140 & .208 & 4.82 & 25 & .109 & 5.20 & 19 & .094 \\
InfoDet   & 1.98 &  6 & .014 & 3.89 & 10 & .034 & 2.00 &  26 & .058 & 4.85 &  6 & .032 & 5.21 &  9 & .033 \\
Crello    & 1.98 & 14 & .014 & 3.85 & 20 & .037 & 1.99 &  65 & .069 & 4.83 & 22 & .044 & 5.20 & 28 & .039 \\
GEdit$^*$ & 1.96 &  4 & .029 & 3.87 &  6 & .098 & 1.97 &  23 & .299 & 4.81 & 11 & .114 & 5.19 &  9 & .150 \\
PIE++     & 1.97 &  7 & .064 &      &    &      &      &     &      & 4.83 & 26 & .349 & 5.20 & 20 & .316 \\
\bottomrule
\end{tabular*}
\caption{\textbf{Klein Base: fewer steps, no guidance, and \method on 14 steps} (Spd in $\times$; Drop and LPIPS
against 28-step vanilla on the same images). $^*$First 100 images. PIE++: PIE-Bench++~\cite{paralleledits}, 140
images; blank: not run.}
\label{tab:supp_bsteps}
\end{table*}

\begin{table}[t]
\centering
\footnotesize
\setlength{\tabcolsep}{4pt}
\begin{tabular*}{\linewidth}{@{\extracolsep{\fill}}llrrrr@{}}
\toprule
Set & Run & Spd & Done & Drop & LPIPS \\
\midrule
MICE (100)  & seed 0   & 4.85$\times$ & 70.6 & 46 & 0.266 \\
            & seed 1   & 4.81$\times$ & 71.7 & 45 & 0.272 \\
            & seed 2   & 4.81$\times$ & 70.1 & 39 & 0.265 \\
LoMOE (64)  & seed 0   & 4.89$\times$ & 80.7 &  4 & 0.187 \\
            & seed 1   & 4.81$\times$ & 76.0 &  9 & 0.167 \\
            & seed 2   & 4.82$\times$ & 76.6 &  7 & 0.164 \\
LoMOE (16)  & 768$^2$  & 4.57$\times$ & 87.0 &  0 & 0.211 \\
\bottomrule
\end{tabular*}
\caption{\textbf{Robustness on Klein Base} (\method). Drop against vanilla with the same seed; 768$^2$ with exact
chunked attention and a tiled VAE.}
\label{tab:supp_brobust}
\end{table}

\paragraph{Observations.}
(1) One global threshold loses edits of
small objects far more often (main paper, Tab.~2); per-part budgets lose fewer at every object size, about half as many
at 8 steps. (2) Per-part budgets drop fewer edits at
every edit density (\cref{tab:supp_density}). (3) Fixed costs of about 2\,s per image (text encoder, VAE) cap the
end-to-end speed-up of every method; \method's transformer speed-up is 2.5$\times$. (4) The CLIP check agrees with
the judge on MICE, LoMOE and MIRA; on the text sets (InfoDet, Crello) the rules are close and either can lead
(\cref{tab:supp_clip}). (5) On GEdit, \method-quality matches vanilla on most tasks; color, material and subject
replacement lose about 8 points (\cref{tab:supp_gedit}).

\subsection{Timing on modern hardware and measured compute}
\label{sec:supp_hw}
All main timings use a GTX 1080 Ti. To check that the speed-ups are not an artifact of an older GPU, we re-time every
method on an NVIDIA A100 (40\,GB, all models on one GPU) with the same code, settings and images
(\cref{tab:supp_a100}), and count the transformer's floating-point operations directly (\cref{tab:supp_flops}).

\paragraph{Klein Base.} The A100 speed-ups match the GTX 1080 Ti: \method runs at 4.69$\times$ end to end
(5.22$\times$ for the transformer alone) and remains the fastest method.

\paragraph{Klein 4B.} The whole edit takes only 0.68\,s on the A100, so \method's fixed per-step work (regions, token
gathering) weighs more. \method-fast falls from 2.20 to 1.74$\times$, level with step skip (1.73$\times$), and
\method-quality to 1.27$\times$, since it also computes the final step in full.

\paragraph{Compute.} Independent of the GPU, \method needs 2.6$\times$ fewer TFLOPs than vanilla on Klein 4B and
5.3$\times$ fewer on Klein Base, the largest saving of any compared method. On Klein 4B, step skip and TeaCache save
2.0$\times$ and keep more edits.

\paragraph{Output consistency.} The same image and seed on the two GPUs differ by LPIPS 0.17--0.28 (20 images), less
than a change of seed (0.29--0.39), so the GTX 1080 Ti quality results carry over.

\begin{table}[t]
\centering
\footnotesize
\setlength{\tabcolsep}{2.2pt}
\renewcommand{\arraystretch}{0.94}
\begin{tabular*}{\linewidth}{@{\extracolsep{\fill}}lrrrrrrr@{}}
\toprule
 & \multicolumn{2}{c}{Time (s)} & \multicolumn{3}{c}{A100 speed-up} & & 1080 Ti \\
\cmidrule(lr){2-3}\cmidrule(lr){4-6}
Method & TF & E2E & TF & E2E & Med.\,3 & Peak & E2E \\
\midrule
\multicolumn{8}{@{}l}{\emph{Klein 4B, 4 steps}} \\
Vanilla              & 0.55 & 0.68 & 1.00 & 1.00 & --   & 15.5 & 1.00 \\
Step skip (2)        & 0.27 & 0.40 & 2.05 & 1.73 & 1.71 & 15.5 & 1.84 \\
TeaCache             & 0.29 & 0.42 & 1.94 & 1.63 & 1.69 & 15.6 & 1.77 \\
RegionE              & 0.43 & 0.57 & 1.28 & 1.20 & 1.23 & 16.0 & 1.30 \\
RAS$^\dagger$        & 0.31 & 0.44 & 1.73 & 1.52 & 1.52 & 16.0 & 1.65 \\
TaylorSeer$^\dagger$ & 0.29 & 0.42 & 1.85 & 1.61 & 1.61 & 16.4 & 1.79 \\
SpotEdit$^\dagger$   & 0.38 & 0.53 & 1.42 & 1.28 & 1.28 & 18.7 & 1.54 \\
\method-quality      & 0.40 & 0.54 & 1.39 & 1.27 & 1.29 & 16.0 & 1.79 \\
\method-fast         & 0.26 & 0.39 & 2.15 & 1.74 & 1.74 & 16.0 & 2.20 \\
\midrule
\multicolumn{8}{@{}l}{\emph{Klein Base, 28 steps}} \\
Vanilla              & 8.32 & 8.53 & 1.00 & 1.00 & --   & 15.6 & 1.00 \\
Step skip (7)        & 2.04 & 2.23 & 4.07 & 3.83 & 3.83 & 15.6 & 3.90 \\
TeaCache             & 1.74 & 1.94 & 4.79 & 4.39 & 4.26 & 15.6 & 4.50 \\
RAS$^\dagger$        & 4.44 & 4.66 & 1.78 & 1.74 & 1.74 & 16.7 & 1.76 \\
TaylorSeer$^\dagger$ & 2.46 & 2.66 & 3.26 & 3.10 & 3.10 & 17.3 & 3.65 \\
SpotEdit$^\dagger$   & 3.19 & 3.46 & 2.50 & 2.36 & 2.36 & 18.7 & 2.56 \\
\method              & 1.60 & 1.82 & 5.22 & 4.69 & 4.53 & 16.7 & 4.85 \\
\method, 14 steps    & 1.44 & 1.64 & 5.73 & 5.18 & --   & 16.7 & 5.23 \\
\bottomrule
\end{tabular*}
\caption{\textbf{Timing on an A100} (MICE-Bench, all 262 images; 14 steps: first 100). TF: transformer; E2E: end to
end; speed-ups in $\times$ against vanilla on the same GPU. Med.\,3: median of three repeats on the first 30 / 20
images (repeats within 3\%); $^\dagger$timed on that subset only. Peak: GPU memory (GB).}
\label{tab:supp_a100}
\end{table}

\begin{table}[t]
\centering
\footnotesize
\setlength{\tabcolsep}{4pt}
\begin{tabular*}{\linewidth}{@{\extracolsep{\fill}}lrr|lrr@{}}
\toprule
\multicolumn{3}{@{}l|}{\emph{Klein 4B, 4 steps}} & \multicolumn{3}{l@{}}{\emph{Klein Base, 28 steps}} \\
Method & TFLOPs & Saving & Method & TFLOPs & Saving \\
\midrule
Vanilla          & 71.0 & 1.00 & Vanilla           & 993.7 & 1.00 \\
Step skip (2)    & 35.5 & 2.00 & Step skip (14)    & 496.8 & 2.00 \\
TeaCache         & 35.5 & 2.00 & Step skip (7)     & 248.4 & 4.00 \\
RegionE          & 53.3 & 1.33 & TeaCache 0.4      & 355.1 & 2.80 \\
\method-quality  & 46.7 & 1.52 & TeaCache 0.6      & 248.6 & 4.00 \\
\method-fast     & 27.1 & 2.62 & TeaCache 0.8      & 213.1 & 4.66 \\
\method, 2 steps & 25.1 & 2.83 & \method           & 186.4 & 5.33 \\
                 &      &      & + TeaCache 0.6    & 186.7 & 5.32 \\
                 &      &      & + TeaCache 0.8    & 186.2 & 5.34 \\
                 &      &      & \method, 14 steps & 172.2 & 5.77 \\
\bottomrule
\end{tabular*}
\caption{\textbf{Measured transformer compute} per image (PyTorch FLOP counter, first 5 MICE images; saving in
$\times$ over vanilla).}
\label{tab:supp_flops}
\end{table}

\subsection{Robustness of the Drop results}
\label{sec:supp_robust_drop}

The per-part vs.\ global gap is the central result, so we check that it does not depend on the seed, the judge or
how the instruction is split.

\paragraph{Seeds.} Across three seeds, each against vanilla with the same seed, one global threshold drops
266\,$\pm$\,27 edits and \method-fast 169\,$\pm$\,14, significant at every seed (\cref{tab:supp_seeds}). On the first
100 Klein Base images, step skip and TeaCache drop slightly fewer edits than \method at every seed; on all 262 images
\method drops fewer than TeaCache (76 vs.\ 100, \cref{tab:supp_stats}).

\paragraph{Judges.} The gap also holds under a second judge from another model family (\cref{tab:supp_judge2}) and
for human raters on the discordant edits (\cref{tab:supp_pairs}).

\paragraph{Instruction parts.} \method does not rely on the benchmark's own split (\cref{tab:supp_split}). A
rule-based split, an LLM split, merging two parts or splitting one in two never loses significantly more edits; the
coarser splits even lose slightly fewer.

\paragraph{Scale of LPIPS.} Re-running vanilla with another seed moves its output by LPIPS 0.29 (Klein 4B, MICE), 0.39
(Klein Base, MICE) and 0.26 (Klein Base, LoMOE), more than \method-fast's distance to vanilla on Klein 4B (0.22).

\begin{table}[t]
\centering
\footnotesize
\setlength{\tabcolsep}{4pt}
\begin{tabular*}{\linewidth}{@{\extracolsep{\fill}}lcccc@{}}
\toprule
Drop & Seed 0 & Seed 1 & Seed 2 & Mean $\pm$ sd \\
\midrule
\multicolumn{5}{@{}l}{\emph{Klein 4B, MICE (262)}} \\
Step skip (2 steps) &  57 &  66 &  69 & 64.0 $\pm$ 6.2 \\
Global threshold    & 237 & 273 & 289 & 266.3 $\pm$ 26.6 \\
\method-fast        & 155 & 183 & 170 & 169.3 $\pm$ 14.0 \\
$p$, global vs.\ \method & $<10^{-12}$ & $<10^{-13}$ & $<10^{-18}$ & \\
\midrule
\multicolumn{5}{@{}l}{\emph{Klein Base, MICE (100)}} \\
Step skip (7 steps) & 36 & 39 & 34 & 36.3 $\pm$ 2.5 \\
TeaCache            & 35 & 37 & 32 & 34.7 $\pm$ 2.5 \\
\method             & 46 & 45 & 39 & 43.3 $\pm$ 3.8 \\
\bottomrule
\end{tabular*}
\caption{\textbf{Dropped edits over three seeds}, each against vanilla with the same seed ($p$: exact McNemar).}
\label{tab:supp_seeds}
\end{table}

\begin{table}[t]
\centering
\footnotesize
\setlength{\tabcolsep}{4pt}
\begin{tabular*}{\linewidth}{@{\extracolsep{\fill}}lccl@{}}
\toprule
Parts given to \method & Spd & Drop & vs.\ benchmark parts \\
\midrule
Benchmark parts (main)  & 2.20$\times$ & 155 & -- \\
Rule-based split        & 2.17$\times$ & 140 & $-15$ [$-31$, 0], $p{=}0.036$ \\
LLM split (Qwen3-VL-8B) & 2.22$\times$ & 152 & $-3$ [$-16$, +9], $p{=}0.71$ \\
Two parts merged        & 2.19$\times$ & 139 & $-16$ [$-32$, $-2$], $p{=}0.026$ \\
One part split in two   & 2.19$\times$ & 159 & +4 [$-6$, +14], $p{=}0.57$ \\
\bottomrule
\end{tabular*}
\caption{\textbf{How the instruction is split} (Klein 4B, $\tau{=}1.0$, MICE, 262 images). Only the parts given to
\method change; the judge always asks about the benchmark's parts.}
\label{tab:supp_split}
\end{table}

\subsection{Combining \method with TeaCache and fewer steps}
\label{sec:supp_stack}

\method is orthogonal to step-level acceleration, so we combine it with TeaCache and with a 14-step schedule on Klein
Base (\cref{tab:supp_stack}). TeaCache's official step-skipping rule runs on top of \method but never on its dense,
check or last steps; skipped steps reuse the last velocity of active tokens and the exact endpoint-freeze velocity of
frozen ones.

\begin{figure}[t]
\centering
\begin{tikzpicture}
\begin{axis}[width=\linewidth, height=4.6cm, xmin=0.8, xmax=5.6, ymin=-2, ymax=55,
  grid=major, grid style={gray!15}, axis line style={gray!60}, tick style={gray!60},
  xlabel={speed-up over vanilla (A100)}, ylabel={edits dropped $\downarrow$},
  tick label style={font=\scriptsize}, label style={font=\scriptsize}, clip=false]
\addplot[gray!60, dashed, thick] coordinates {(1.00,0) (1.00,25) (1.96,25) (1.96,26) (2.66,26) (2.66,27) (3.69,27) (3.69,35) (4.23,35) (4.23,38) (4.38,38) (4.38,39) (4.86,39)};
\addplot[orange!85!black, thick, mark=diamond*, mark size=2.2pt] coordinates {(1.00,0) (1.96,25) (3.77,36)};
\addplot[black!70, thick, mark=triangle*, mark size=2pt] coordinates {(2.66,26) (3.69,27) (4.23,35)};
\addplot[only marks, black!70, mark=triangle, mark size=2pt] coordinates {(2.62,35)};
\addplot[only marks, cvprblue, mark=square*, mark size=2pt] coordinates {(4.38,46) (4.76,39) (4.86,39)};
\addplot[only marks, cvprblue, mark=square, mark size=2pt, thick] coordinates {(4.29,39) (4.38,38) (4.41,47)};
\node[font=\tiny, text=orange!70!black, anchor=north west, inner sep=1.5pt] at (axis cs:1.96,25) {step skip, 14};
\node[font=\tiny, text=orange!70!black, anchor=south, inner sep=2pt] at (axis cs:3.77,36.5) {step skip, 7};
\node[font=\tiny, text=black!70, anchor=north, inner sep=2pt] at (axis cs:3.0,25.5) {TeaCache};
\node[font=\tiny, text=black!70, anchor=east, inner sep=3pt] at (axis cs:2.62,35) {TeaCache, 14 steps};
\node[font=\tiny, text=cvprblue, anchor=east, inner sep=2pt] at (axis cs:4.33,46) {\textbf{\method}};
\node[font=\tiny, text=cvprblue, anchor=north west, inner sep=2pt] at (axis cs:4.42,37.5) {+ TeaCache};
\node[font=\tiny, text=cvprblue, anchor=south west, inner sep=1.5pt] at (axis cs:4.90,40) {\textbf{\method}, 14 steps};
\end{axis}
\end{tikzpicture}
\caption{\textbf{Speed vs.\ dropped edits on an A100} (Klein Base, first 100 MICE images; dashed: Pareto frontier; open
squares: \method + TeaCache). As on the GTX 1080 Ti, \method on 14 steps extends the frontier beyond TeaCache's
fastest setting (4.76--4.86$\times$ at 39 drops vs.\ 4.23$\times$ at 35, $p\ge0.57$).}
\label{fig:supp_pareto_a100}
\end{figure}
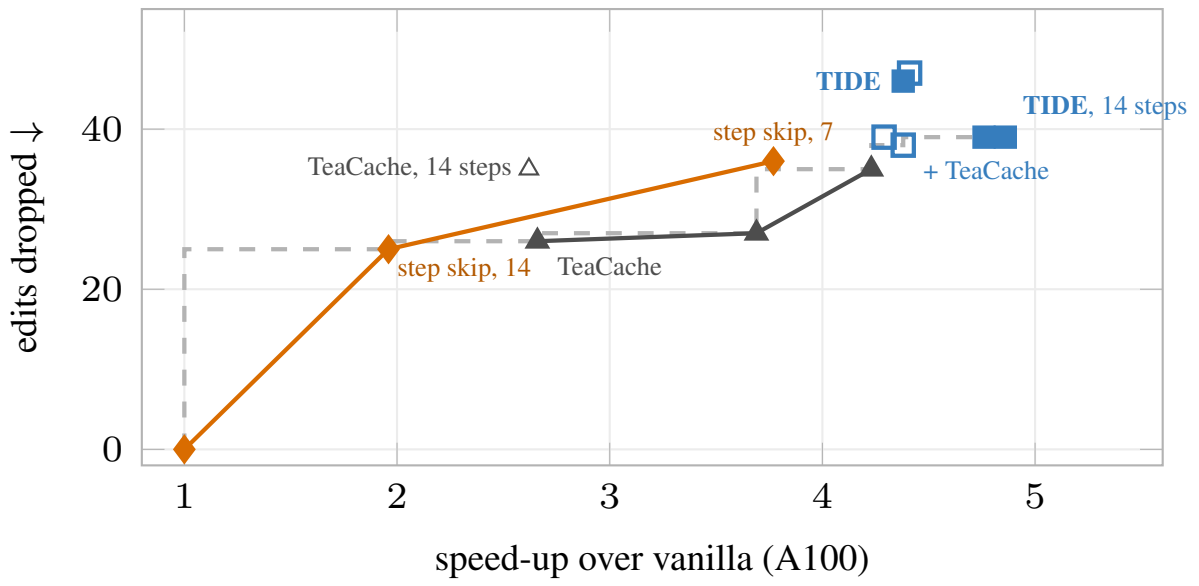

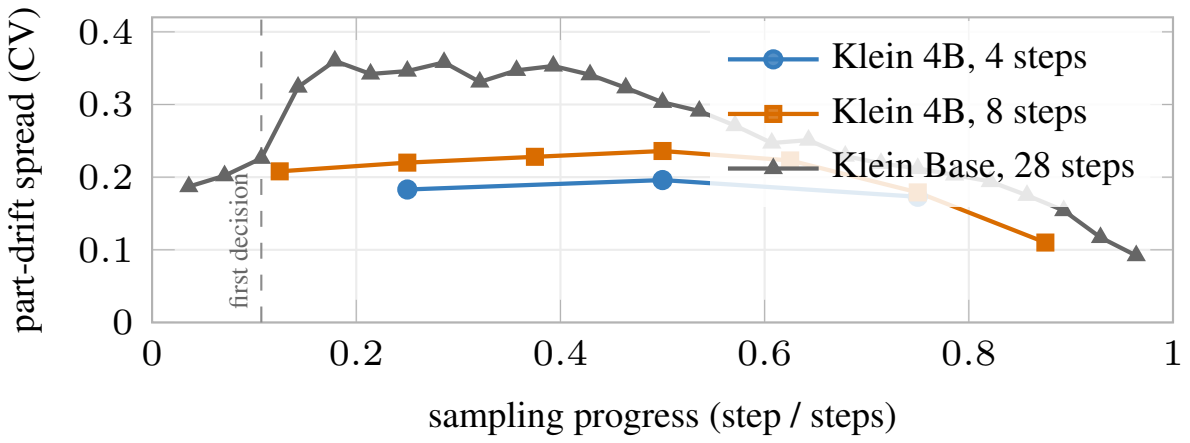
\begin{figure}[t]
\centering
\begin{tikzpicture}
\begin{axis}[width=\linewidth, height=3.6cm, xmin=0, xmax=1, ymin=0, ymax=0.42,
  grid=major, grid style={gray!15}, axis line style={gray!60}, tick style={gray!60},
  xlabel={sampling progress (step / steps)}, ylabel={part-drift spread (CV)},
  tick label style={font=\scriptsize}, label style={font=\scriptsize},
  legend style={font=\scriptsize, draw=none, fill=white, fill opacity=0.85, text opacity=1, at={(0.99,0.99)}, anchor=north east},
  legend cell align=left]
\addplot[cvprblue, thick, mark=*, mark size=1.5pt] coordinates {(0.250,0.183) (0.500,0.196) (0.750,0.173)};
\addlegendentry{Klein 4B, 4 steps}
\addplot[orange!85!black, thick, mark=square*, mark size=1.3pt] coordinates {(0.125,0.208) (0.250,0.220) (0.375,0.228) (0.500,0.236) (0.625,0.223) (0.750,0.179) (0.875,0.110)};
\addlegendentry{Klein 4B, 8 steps}
\addplot[black!60, thick, mark=triangle*, mark size=1.3pt] coordinates {(0.036,0.187) (0.071,0.202) (0.107,0.226) (0.143,0.324) (0.179,0.360) (0.214,0.342) (0.250,0.346) (0.286,0.358) (0.321,0.331) (0.357,0.347) (0.393,0.353) (0.429,0.341) (0.464,0.323) (0.500,0.303) (0.536,0.291) (0.571,0.271) (0.607,0.247) (0.643,0.251) (0.679,0.229) (0.714,0.219) (0.750,0.212) (0.786,0.203) (0.821,0.194) (0.857,0.175) (0.893,0.154) (0.929,0.117) (0.964,0.092)};
\addlegendentry{Klein Base, 28 steps}
\draw[black!45, dashed] (axis cs:0.107,0) -- (axis cs:0.107,0.42);
\node[font=\tiny, text=black!60, anchor=south west, rotate=90, inner sep=1pt] at (axis cs:0.100,0.01) {first decision};
\end{axis}
\end{tikzpicture}
\caption{\textbf{When do the parts' drift scales separate?} Coefficient of variation of the parts' mean drift within an
image (vanilla, MICE, mean over images). On Klein Base the parts move alike until the first decision (after step 3, dashed)
and separate only afterwards.}
\label{fig:supp_spread}
\end{figure}

\paragraph{Results.} \method on 14 steps reaches 5.23$\times$ at 39 drops, about twice as fast as TeaCache on 14
steps (2.68$\times$, 35, $p{=}0.57$). On the A100 it extends the speed--drop frontier beyond TeaCache's fastest setting
(\cref{fig:supp_pareto_a100}). Adding TeaCache to \method on 28 steps adds little speed (4.82--4.94$\times$).

\paragraph{Timing.} This A100 column is from a second A100 session; vanilla ran 7\% faster there, so speed-ups are 2--8\% lower than in \cref{tab:supp_a100}. Drop uses the GTX 1080 Ti outputs.

\begin{table}[t]
\centering
\footnotesize
\setlength{\tabcolsep}{3.5pt}
\begin{tabular*}{\linewidth}{@{\extracolsep{\fill}}lcccc@{}}
\toprule
Klein Base, MICE (100) & 1080 Ti & A100 & Drop & $p$ \\
\midrule
Step skip (7 steps)      & 3.91$\times$ & 3.77$\times$ & 36 & 0.2 \\
TeaCache 0.8             & 4.50$\times$ & 4.23$\times$ & 35 & 0.099 \\
TeaCache 0.6             & 3.86$\times$ & 3.68$\times$ & 27 & 0.0066 \\
TeaCache 0.4             & 2.76$\times$ & 2.66$\times$ & 26 & 0.0045 \\
\textbf{\method}         & 4.85$\times$ & 4.38$\times$ & 46 & -- \\
\method + TeaCache 0.8   & 4.93$\times$ & 4.41$\times$ & 47 & 1 \\
\method + TeaCache 0.6   & 4.94$\times$ & 4.38$\times$ & 38 & 0.077 \\
\method + TeaCache 0.4   & 4.82$\times$ & 4.29$\times$ & 39 & 0.12 \\
\midrule
Vanilla, 14 steps          & 2.00$\times$ & 1.96$\times$ & 25 & 0.15 \\
TeaCache, 14 steps         & 2.68$\times$ & 2.62$\times$ & 35 & -- \\
\textbf{\method}, 14 steps & 5.23$\times$ & 4.76$\times$ & 39 & 0.57 \\
\quad $\tau{=}1.0$         & 5.04$\times$ & 4.64$\times$ & 42 & 0.31 \\
\quad $\tau{=}2.5$         & 5.35$\times$ & 4.86$\times$ & 39 & 0.60 \\
\bottomrule
\end{tabular*}
\caption{\textbf{\method with TeaCache and 14 steps} (speed and Drop vs.\ 28-step vanilla). $p$ (McNemar): top, vs.\
\method; bottom, vs.\ TeaCache on 14 steps.}
\label{tab:supp_stack}
\end{table}

\subsection{Where the speed and the losses come from}
\label{sec:supp_attrib}

\method combines three savings: cached text and reference keys and values after step 0, reuse of the empty-prompt
branch on Klein Base, and frozen tokens. \Cref{tab:supp_attrib} separates them by running \method at $\tau{=}0$,
which keeps every saving except freezing.

\paragraph{Klein 4B.} The caches alone give 1.71$\times$ and already lose 120 edits; freezing raises the speed to
2.20$\times$ for 35 more drops. Step skip (1.84$\times$, 57 drops) keeps more edits than either, so on this 4-step editor
most of \method-fast's losses come from the cached keys and values, not from the per-instruction freezing itself.
Refreshing the frozen tokens' keys and values brings \method-fast back to 121 drops ($p{=}1$ vs.\ caches alone) but also
back to 1.71$\times$; extrapolating them keeps 2.15$\times$ but loses 231 edits.

\paragraph{Allocation vs.\ an oracle.} An oracle that knows vanilla's final image and spreads the same number of
frozen tokens over the parts by their true error loses as many edits as \method-fast (155 and 158 vs.\ 155,
$p{\ge}0.81$). At this speed, the per-part rule allocates the budget as well as the oracle.

\paragraph{Klein Base.} Stale keys and values cost nothing here (45 drops refreshed or extrapolated vs.\ 46, $p{=}1$).
The caches alone give 2.51$\times$ at 27 drops, and freezing raises this to 4.85$\times$. Giving TeaCache the same
caches does not speed it up (4.43--4.51$\times$ vs.\ 4.50$\times$; 38--39 drops vs.\ 35), because it must then also
compute \method's dense, check and last steps (10 of 28 steps instead of 6--7). On LoMOE and MIRA
(\cref{tab:supp_lomoe_tc}), \method is faster than every TeaCache setting, with fewer drops on LoMOE and as many as
TeaCache 0.6--0.8 on MIRA.

\begin{table}[t]
\centering
\footnotesize
\setlength{\tabcolsep}{2.6pt}
\renewcommand{\arraystretch}{0.94}
\begin{tabular*}{\linewidth}{@{\extracolsep{\fill}}lrrrrl@{}}
\toprule
Setting & Spd & A100 & Drop & LPIPS & $p$ \\
\midrule
\multicolumn{6}{@{}l}{\emph{Klein 4B, MICE (262), vs.\ 4-step vanilla}} \\
Step skip (2 of 4 steps)        & 1.84 & --   & 57  & 0.145 & \\
Caches only ($\tau{=}0$)        & 1.71 & --   & 120 & 0.203 & 0.0012 \\
\textbf{\method-fast}           & 2.20 & --   & 155 & 0.222 & -- \\
\quad frozen K/V refreshed      & 1.71 & --   & 121 & 0.206 & $1.5{\times}10^{-4}$ \\
\quad frozen K/V extrapolated   & 2.15 & --   & 231 & 0.257 & $4{\times}10^{-11}$ \\
\quad oracle split, drift order & 2.16 & --   & 155 & 0.221 & 1 \\
\quad oracle split, true error  & 2.19 & --   & 158 & 0.218 & 0.81 \\
\midrule
\multicolumn{6}{@{}l}{\emph{Klein Base, MICE (100), vs.\ 28-step vanilla}} \\
TeaCache 0.8                    & 4.50 & 4.23 & 35  & 0.191 & \\
\quad + \method's caches        & 4.51 & 4.20 & 39  & 0.254 & 0.64$^\dagger$ \\
TeaCache 0.6 + \method's caches & 4.43 & 4.08 & 38  & 0.254 & 0.73$^\dagger$ \\
Caches only ($\tau{=}0$)        & 2.51 & --   & 27  & 0.250 & 0.0034 \\
\textbf{\method}                & 4.85 & 4.38 & 46  & 0.266 & -- \\
\quad frozen K/V refreshed      & 4.34 & --   & 45  & 0.266 & 1 \\
\quad frozen K/V extrapolated   & 4.80 & --   & 45  & 0.268 & 1 \\
\method on 14 steps             & 5.23 & 4.76 & 39  & 0.246 & \\
\bottomrule
\end{tabular*}
\caption{\textbf{Where the speed and the losses come from} (Spd in $\times$; A100: session of
\cref{fig:supp_pareto_a100}). $p$: exact McNemar vs.\ \method, $^\dagger$vs.\ TeaCache 0.8.}
\label{tab:supp_attrib}
\end{table}

\paragraph{Variants in \cref{tab:supp_attrib}.}
\emph{Caches only} runs \method with $\tau{=}0$: text and reference keys and values are cached after step 0, and on
Klein Base the empty-prompt branch is reused and the check steps run. \emph{K/V refreshed} computes frozen tokens at
every sparse step only to refresh their keys and values; their velocity is discarded and they still move by
endpoint-freeze. \emph{K/V extrapolated} extrapolates their keys and values linearly from their last two computed steps.
\emph{Oracle split}: at every step \method's rule proposes the frozen set and the background keeps it, but the same
number of object tokens is spread over the parts by their true freezing error (distance of the token's endpoint to
vanilla's final latent), so that the largest error per unit of demand is as small as possible; within a part, tokens go
by drift or by true error. Further paired tests: on Klein 4B, caches only vs.\ step skip $p{=}4{\times}10^{-10}$ and
K/V refreshed vs.\ caches only $p{=}1$; on Klein Base, TeaCache with \method's caches vs.\ \method on 14 steps $p{=}1$.

\begin{table}[t]
\centering
\footnotesize
\setlength{\tabcolsep}{3pt}
\begin{tabular*}{\linewidth}{@{\extracolsep{\fill}}lrrrrrr@{}}
\toprule
 & \multicolumn{3}{c}{LoMOE (64)} & \multicolumn{3}{c}{MIRA (100)} \\
\cmidrule(lr){2-4}\cmidrule(lr){5-7}
Klein Base & Spd & Drop & LPIPS & Spd & Drop & LPIPS \\
\midrule
TeaCache 0.4           & 2.74 & 6 & 0.061 & 2.74 & 10 & 0.037 \\
TeaCache 0.6           & 3.85 & 9 & 0.103 & 3.83 & 22 & 0.066 \\
TeaCache 0.8           & 4.44 & 9 & 0.124 & 4.46 & 28 & 0.087 \\
\textbf{\method}       & 4.89 & 4 & 0.187 & 4.82 & 25 & 0.109 \\
\method + TeaCache 0.6 & 4.86 & 4 & 0.186 & 4.90 & 23 & 0.108 \\
\bottomrule
\end{tabular*}
\caption{\textbf{TeaCache curve on LoMOE and MIRA} (Klein Base; Spd in $\times$; Drop and LPIPS vs.\ 28-step
vanilla).}
\label{tab:supp_lomoe_tc}
\end{table}

\paragraph{TeaCache on LoMOE and MIRA.} \method is faster than every TeaCache setting on both benchmarks
(\cref{tab:supp_lomoe_tc}). On LoMOE it drops fewer edits than TeaCache 0.8 (4 vs.\ 9, $p{=}0.23$); on MIRA it drops as
many as TeaCache 0.6--0.8 (25 vs.\ 22--28, $p{\ge}0.63$), while TeaCache 0.4 keeps more edits at 2.74$\times$ (10,
$p{=}0.0015$). Adding TeaCache to \method gains little.

\subsection{Image quality and evaluation robustness}
\label{sec:supp_further}

\paragraph{Seed noise floor.}
Re-running vanilla with another seed already loses some of the edits that seed-0 vanilla completes
(\cref{tab:supp_floor}). On Klein Base, \method drops fewer edits than this floor on five of six benchmarks (GEdit:
11 vs.\ 7). On Klein 4B, \method-fast stays within it on MIRA and InfoDet, and step skip on every benchmark.

\begin{table}[t]
\centering
\footnotesize
\setlength{\tabcolsep}{3pt}
\begin{tabular*}{\linewidth}{@{\extracolsep{\fill}}lrrrrrr@{}}
\toprule
 & MICE & LoMOE & MIRA & InfoDet & Crello & GEdit \\
\midrule
\multicolumn{7}{@{}l}{\emph{Klein 4B, 4 steps}} \\
Completed (seed 0)  & 1577 & 158 & 402 & 360 & 206 & 519 \\
Floor (other seed)  & 96--101 & 3--5 & 22 & 25 & 22 & 17 \\
Step skip (2 steps) & 57  & 3 & 11 & 9  & 18 & 13 \\
TeaCache            & 91  & 3 & 23 & 19 & 25 & 17 \\
\method-fast        & 155 & 7 & 9  & 8  & 28 & 25 \\
\midrule
\multicolumn{7}{@{}l}{\emph{Klein Base, 28 steps (MICE, GEdit: first 100)}} \\
Completed (seed 0)  & 501 & 144 & 376 & 343 & 198 & 86 \\
Floor (other seed)  & 53--55 & 8--11 & 34 & 14 & 30 & 7 \\
Step skip (7 steps) & 36 & 7 & 21 & 10 & 20 & 6 \\
TeaCache            & 35 & 9 & 28 & 8  & 25 & 6 \\
\method             & 46 & 4 & 25 & 6  & 22 & 11 \\
\bottomrule
\end{tabular*}
\caption{\textbf{Seed noise floor.} Floor: edits that seed-0 vanilla completes and vanilla with seed 1 (MICE, LoMOE:
seeds 1 and 2) loses. Method rows: Drop against seed-0 vanilla.}
\label{tab:supp_floor}
\end{table}

\paragraph{Second judge and VIEScore.}
Idefics3-8B~\cite{idefics3}, built on Llama~3 rather than Qwen, asks the same questions with the same threshold
(\cref{tab:supp_judge2}). It is more lenient (88.6\% vs.\ 75.3\% of vanilla's parts done; 76.8\% agreement) and so
counts fewer drops, but it ranks the methods in the same order and confirms the gap between the rules (lost only by the
global threshold: 34, only by \method: 17; $p{=}0.024$). VIEScore uses its official prompts, scored by Qwen3-VL-8B.

\begin{table}[t]
\centering
\footnotesize
\setlength{\tabcolsep}{3pt}
\begin{tabular*}{\linewidth}{@{\extracolsep{\fill}}lrrrrr@{}}
\toprule
 & \multicolumn{2}{c}{Drop} & \multicolumn{3}{c}{VIEScore} \\
\cmidrule(lr){2-3}\cmidrule(lr){4-6}
Method & Judge & 2nd & SC & PQ & O \\
\midrule
\multicolumn{6}{@{}l}{\emph{Klein 4B, MICE (262)}} \\
Vanilla (4 steps)    & --  & -- & 8.33 & 8.29 & 8.24 \\
Step skip (2 steps)  & 57  & 33 & 8.31 & 8.24 & 8.22 \\
TeaCache             & 91  & 60 & 8.24 & 8.06 & 8.08 \\
Global threshold     & 237 & 96 & 7.77 & 7.45 & 7.52 \\
\method-fast         & 155 & 79 & 8.01 & 7.75 & 7.79 \\
\midrule
\multicolumn{6}{@{}l}{\emph{Klein Base, MICE (100)}} \\
Vanilla (28 steps)   & --  & -- & 7.66 & 8.29 & 7.83 \\
Step skip (7 steps)  & 36  & 21 & 7.62 & 8.18 & 7.80 \\
TeaCache             & 35  & 20 & 7.64 & 8.20 & 7.82 \\
\method              & 46  & 29 & 7.80 & 8.01 & 7.73 \\
\bottomrule
\end{tabular*}
\caption{\textbf{Second judge and VIEScore}~\cite{viescore}. 2nd: Idefics3-8B. SC: semantic consistency; PQ:
perceptual quality; O $=\sqrt{\mathrm{SC}\cdot\mathrm{PQ}}$; 0--10, higher is better.}
\label{tab:supp_judge2}
\end{table}

\paragraph{Background, image quality and text.}
\Cref{tab:supp_quality} reports background preservation (output vs.\ input outside all edit masks), no-reference
quality (CLIP-IQA~\cite{clipiqa}, MUSIQ~\cite{musiq} on KonIQ~\cite{koniq}, via pyiqa~\cite{pyiqa}) and text accuracy
(EasyOCR~\cite{easyocr} on each text part's box, compared with the target text). On Klein Base, \method keeps the
background as well as vanilla, and its quality and text accuracy are close to the other accelerators. On Klein 4B,
\method-fast scores lower on all three, like TaylorSeer and RAS, because frozen tokens keep an estimate from an
earlier step.

\begin{table*}[t]
\centering
\footnotesize
\setlength{\tabcolsep}{4pt}
\renewcommand{\arraystretch}{0.94}
\begin{tabular*}{\textwidth}{@{\extracolsep{\fill}}lrrrrrrrrr@{}}
\toprule
 & \multicolumn{3}{c}{Background (5 sets)} & \multicolumn{2}{c}{No-reference (6 sets)} & \multicolumn{2}{c}{Text, InfoDet} & \multicolumn{2}{c}{Text, Crello} \\
\cmidrule(lr){2-4}\cmidrule(lr){5-6}\cmidrule(lr){7-8}\cmidrule(lr){9-10}
Method & PSNR$\uparrow$ & SSIM$\uparrow$ & LPIPS$\downarrow$ & CLIP-IQA$\uparrow$ & MUSIQ$\uparrow$ & sim.$\uparrow$ & exact$\uparrow$ & sim.$\uparrow$ & exact$\uparrow$ \\
\midrule
\multicolumn{10}{@{}l}{\emph{Klein 4B, 4 steps}} \\
Vanilla           & 22.96 & 0.876 & 0.127 & 0.692 & 71.9 & 0.522 & 11.1 & 0.455 & 11.7 \\
Step skip (2)     & 23.08 & 0.883 & 0.122 & 0.677 & 71.5 & 0.483 &  9.3 & 0.410 & 10.4 \\
TeaCache          & 22.84 & 0.881 & 0.125 & 0.677 & 71.7 & 0.492 & 10.5 & 0.404 &  9.3 \\
TaylorSeer        & 18.00 & 0.699 & 0.247 & 0.557 & 67.9 & 0.351 &  4.7 & 0.247 &  3.9 \\
RAS               & 21.05 & 0.849 & 0.144 & 0.643 & 70.9 & 0.406 &  8.3 & 0.304 &  6.8 \\
RegionE           & 23.05 & 0.877 & 0.126 & 0.683 & 71.6 & 0.512 & 11.3 & 0.401 & 10.0 \\
SpotEdit          & 24.54 & 0.878 & 0.119 & 0.657 & 70.8 & 0.491 & 12.2 & 0.354 &  6.7 \\
MICE$^*$          & 23.89 & 0.819 & 0.096 & 0.688 & 71.8 & 0.477 &  8.2 & 0.500 & 10.1 \\
MIRAGE            & 27.09 & 0.919 & 0.085 & 0.681 & 71.8 & 0.522 & 11.4 & 0.486 & 15.3 \\
\method-quality   & 20.59 & 0.842 & 0.151 & 0.640 & 70.8 & 0.362 &  6.2 & 0.314 &  8.9 \\
\method-fast      & 21.46 & 0.860 & 0.143 & 0.629 & 70.1 & 0.390 &  7.8 & 0.285 &  6.0 \\
\midrule
\multicolumn{10}{@{}l}{\emph{Klein Base, 28 steps}} \\
Vanilla           & 21.04 & 0.816 & 0.179 & 0.709 & 72.0 & 0.499 & 10.9 & 0.405 &  8.4 \\
Step skip (7)     & 21.44 & 0.829 & 0.164 & 0.695 & 71.8 & 0.477 & 10.1 & 0.369 &  6.4 \\
TeaCache          & 21.90 & 0.814 & 0.168 & 0.704 & 71.8 & 0.464 & 10.1 & 0.371 &  6.7 \\
TaylorSeer        & 19.87 & 0.806 & 0.198 & 0.694 & 71.8 & 0.463 &  9.1 & 0.326 &  4.4 \\
RAS               & 21.03 & 0.823 & 0.175 & 0.692 & 71.7 & 0.463 &  9.8 & 0.339 &  6.0 \\
RegionE           & 21.51 & 0.816 & 0.184 & 0.653 & 69.8 & 0.466 & 10.1 & 0.321 &  6.1 \\
SpotEdit          & 24.83 & 0.801 & 0.213 & 0.624 & 68.1 & 0.447 & 11.9 & 0.282 &  4.5 \\
MICE              & 16.76 & 0.610 & 0.299 & 0.682 & 71.9 & 0.295 &  4.3 & 0.299 &  3.5 \\
MIRAGE            & 25.81 & 0.899 & 0.098 & 0.688 & 71.7 & 0.470 &  9.1 & 0.421 &  9.8 \\
\method           & 21.48 & 0.836 & 0.162 & 0.681 & 71.3 & 0.451 &  9.9 & 0.339 &  7.8 \\
\bottomrule
\end{tabular*}
\caption{\textbf{Background, image quality and text accuracy}, mean over benchmarks. Text sim.: one minus the
normalized edit distance of the best-matching OCR line; exact: \% of parts read exactly. $^*$No-reference scores over
5 sets.}
\label{tab:supp_quality}\label{tab:supp_ocr}
\end{table*}

\paragraph{Resolution 1024$^2$.}
On an A100 at 1024$^2$ (\cref{tab:supp_1024}), per-part thresholds drop far fewer edits than one global threshold at the
same $\tau$, though the global threshold also runs faster. At matched speed, step skip keeps more edits than \method
(15 vs.\ 38 at 1.76--1.77$\times$).

\begin{table}[t]
\centering
\footnotesize
\setlength{\tabcolsep}{4pt}
\begin{tabular*}{\linewidth}{@{\extracolsep{\fill}}lrrl@{}}
\toprule
Klein 4B, 1024$^2$, MICE (100) & Spd & Drop & $p$ vs.\ global \\
\midrule
Step skip (2 steps)            & 1.76$\times$ & 15  & \\
TeaCache                       & 1.28$\times$ & 13  & \\
Global, $\tau{=}1.0$           & 1.88$\times$ & 284 & \\
\textbf{\method}, $\tau{=}1.0$ & 1.66$\times$ & 90  & $<10^{-51}$ \\
Global, $\tau{=}1.5$           & 2.01$\times$ & 189 & \\
\textbf{\method}, $\tau{=}1.5$ & 1.77$\times$ & 38  & $<10^{-45}$ \\
\bottomrule
\end{tabular*}
\caption{\textbf{1024$^2$} (A100, 4 steps; vanilla 2.6\,s per image; outputs judged at 512$^2$).}
\label{tab:supp_1024}
\end{table}

\paragraph{GEdit-Bench scores.}
With Qwen3-VL-8B in place of GEdit-Bench's proprietary scorer (\cref{tab:supp_gedit_vie}), all accelerators on Klein Base stay within 0.07 of vanilla's overall score
(\method: 0.04 below); on Klein 4B, \method-fast is 0.14 below.

\begin{table}[t]
\centering
\footnotesize
\setlength{\tabcolsep}{4pt}
\begin{tabular*}{\linewidth}{@{\extracolsep{\fill}}lrrr|lrrr@{}}
\toprule
\multicolumn{4}{@{}l|}{\emph{Klein 4B, 4 steps}} & \multicolumn{4}{l@{}}{\emph{Klein Base, 28 steps}} \\
Method & SC & PQ & O & Method & SC & PQ & O \\
\midrule
Vanilla       & 8.34 & 8.17 & 8.07 & Vanilla       & 7.98 & 7.84 & 7.64 \\
Step skip (2) & 8.44 & 8.09 & 8.12 & Step skip (7) & 7.92 & 7.79 & 7.57 \\
TeaCache      & 8.53 & 8.13 & 8.21 & TeaCache      & 7.92 & 7.98 & 7.63 \\
\method-fast  & 8.30 & 7.86 & 7.93 & \method       & 7.94 & 7.78 & 7.60 \\
\bottomrule
\end{tabular*}
\caption{\textbf{GEdit-Bench scores} (first 100 images; VIEScore, 0--10).}
\label{tab:supp_gedit_vie}
\end{table}
\begin{table*}[t]
\centering
\footnotesize
\setlength{\tabcolsep}{3pt}
\renewcommand{\arraystretch}{0.94}
\begin{tabular*}{\textwidth}{@{\extracolsep{\fill}}llrrl@{}}
\toprule
Attempt & Setting & Spd & Drop & Reference (Drop, $p$) \\
\midrule
Fewer steps (Klein 4B) & \method on 2 steps                 & 2.43$\times$ & 130 & 1 step: 238 ($<10^{-18}$); 2 steps: 57 ($<10^{-11}$) \\
                               & Global threshold on 2 steps        & 2.45$\times$ & 130 & identical outputs \\
                               & \method on 3 steps                 & 2.26$\times$ & 130 & global on 3 steps: 139 at 2.19$\times$ (0.27) \\
\midrule
Text parts never frozen        & Crello / InfoDet                   & 2.16 / 2.17$\times$ & 39 / 14 & \method-fast 28 / 8 (0.035 / 0.29); OCR 0.27 / 0.38 vs.\ 0.28 / 0.39 \\
\midrule
Demand as price (Base) & $\tau{=}1.5$                       & 4.87$\times$ & 48 & mean drift: 46 (0.75) \\
                               & $\tau{=}2.5$                       & 4.89$\times$ & 48 & mean drift: 41 (0.039) \\
                               & $\tau{=}3.0$                       & 5.00$\times$ & 46 & mean drift: 45 (1) \\
\midrule
Ledger rule                    & Klein Base, $b{=}0.05$, $\tau{=}1.5$ & 3.67$\times$ & 48  & Rule (4): 46 at 4.86$\times$ \\
                               & Klein Base, $b{=}0.02$, $\tau{=}3.0$ & 3.66$\times$ & 52  & Rule (4): 45 at 4.94$\times$ \\
                               & Klein 4B 8 st., $b{=}0.05$, $\tau{=}0.6$ & 2.84$\times$ & 351 & Rule (4): 200 at 3.51$\times$ \\
                               & Klein 4B 8 st., $b{=}0.05$, $\tau{=}1.0$ & 2.90$\times$ & 509 & Rule (4): 169 at 4.14$\times$ \\
\midrule
SpotEdit, relative thr. & per part / global, $\tau{=}0.6$    & 1.46 / 1.52$\times$ & 71 / 69   & official: 87 at 1.54$\times$ (0.87) \\
                               & per part / global, $\tau{=}1.0$    & 1.53 / 1.55$\times$ & 80 / 77   & official: 87 at 1.54$\times$ (0.79) \\
                               & per part / global, $\tau{=}1.5$    & 1.66 / 1.63$\times$ & 154 / 133 & official: 87 at 1.54$\times$ (0.04) \\
\bottomrule
\end{tabular*}
\caption{\textbf{Alternative design choices} (MICE-Bench; Klein 4B: 262 images vs.\ 4-step vanilla; Klein Base: first
100 images). Each row changes one component of \method; the last column compares it with the setting used in the paper.
$p$: exact McNemar; for SpotEdit, per part vs.\ global.}
\label{tab:supp_tried}
\end{table*}

\begin{table}[t]
\centering
\footnotesize
\setlength{\tabcolsep}{3.5pt}
\begin{tabular*}{\linewidth}{@{\extracolsep{\fill}}lrrrr@{}}
\toprule
Edits (Klein 4B, MICE) & Edits & IoU & \method & Global \\
\midrule
IoU, lowest quarter   & 384  & 0.04 & 41 (10.7)  & 47 (12.2) \\
IoU, 2nd quarter      & 378  & 0.14 & 51 (13.5)  & 67 (17.7) \\
IoU, 3rd quarter      & 381  & 0.25 & 31 (8.1)   & 53 (13.9) \\
IoU, highest quarter  & 381  & 0.47 & 25 (6.6)   & 59 (15.5) \\
\midrule
Noun shared           & 1187 & 0.20 & 123 (10.4) & 177 (14.9) \\
Unique noun           & 337  & 0.29 & 25 (7.4)   & 49 (14.5) \\
\midrule
Masks overlap $>$10\% & 127  & 0.16 & 12 (9.4)   & 12 (9.4) \\
No overlap            & 1397 & 0.23 & 136 (9.7)  & 214 (15.3) \\
\midrule
All                   & 1524 & 0.22 & 148 (9.7)  & 226 (14.8) \\
\bottomrule
\end{tabular*}
\caption{\textbf{Dropped edits by region quality} (vanilla's completed edits; drops with \% in brackets). \method-fast
at 2.20$\times$, global threshold at 2.09$\times$, same engine and regions. IoU: best IoU of the part's mask with any
self-grounded region. Masks are matched to the judged parts by their text; 9 parts with empty masks at token resolution
and the 44 parts of 8 images whose masks cannot be matched are left out.}
\label{tab:supp_regstrat}
\end{table}

\paragraph{Quality at matched speed.}
At matched speed, per-part budgets keep more edits than one global threshold without costing image quality
(\cref{fig:supp_quality_curves}). \Cref{fig:supp_speed_quality} places every method by speed and no-reference quality,
and \cref{tab:supp_inout} shows where the distance to vanilla comes from and how sharp the outputs are.

\paragraph{SpotEdit on Klein Base.}
SpotEdit recomputes on average 75\% of the image
tokens per step on Klein Base (59\% on Klein 4B), so its losses do not come from skipping too much. They concentrate on
the many-object sets (MICE, MIRA) and are small on the text sets, consistent with a single per-token reuse decision
tuned for one edit at a time.
\pgfplotsset{qc/.style={width=0.53\linewidth,height=3.2cm,scale only axis=false,grid=major,grid style={gray!18},tick label style={font=\tiny},label style={font=\scriptsize},title style={font=\scriptsize,yshift=-4pt},xlabel={Speed-up},mark size=1.5pt}}
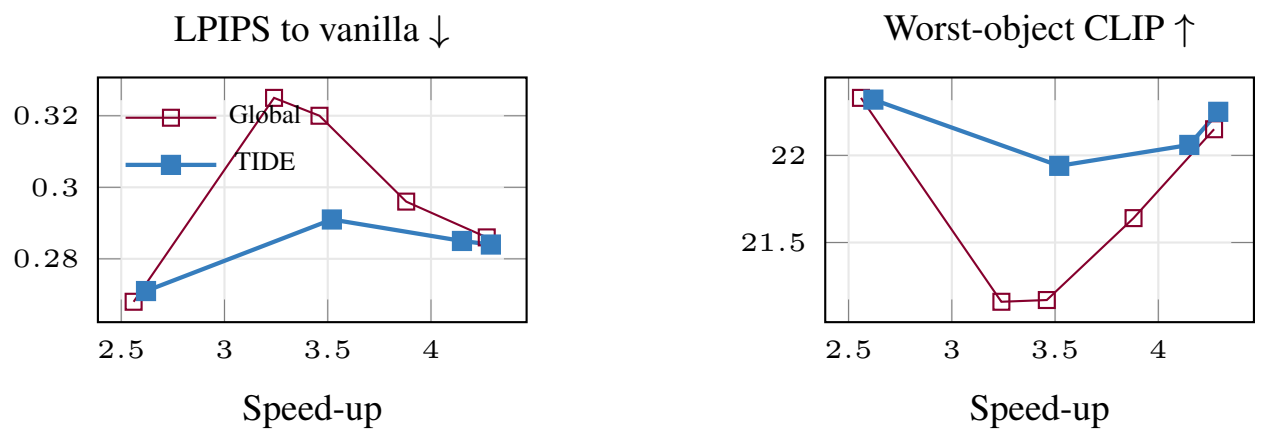
\begin{figure}[t]
\centering
\begin{tikzpicture}\begin{axis}[qc,title={LPIPS to vanilla $\downarrow$},legend style={font=\tiny,draw=none,fill=none,at={(0.02,0.98)},anchor=north west}]
\addplot[purple!70!black,mark=square] coordinates {(2.56,0.268) (3.24,0.325) (3.46,0.320) (3.88,0.296) (4.27,0.286)}; \addlegendentry{Global}
\addplot[cvprblue,thick,mark=square*] coordinates {(2.62,0.271) (3.52,0.291) (4.15,0.285) (4.29,0.284)}; \addlegendentry{\method}
\end{axis}\end{tikzpicture}\hfill
\begin{tikzpicture}\begin{axis}[qc,title={Worst-object CLIP $\uparrow$}]
\addplot[purple!70!black,mark=square] coordinates {(2.56,22.330) (3.24,21.160) (3.46,21.170) (3.88,21.640) (4.27,22.150)};
\addplot[cvprblue,thick,mark=square*] coordinates {(2.62,22.320) (3.52,21.940) (4.15,22.060) (4.29,22.250)};
\end{axis}\end{tikzpicture}
\caption{\textbf{Keeping edits does not cost image quality} (per-part vs.\ one global threshold, Klein 4B, 8 steps, MICE-Bench, one point per $\tau$; same engine and regions, only the rule differs). At medium speed \method is both closer to vanilla and better on its worst-edited object; at the two ends the rules coincide (data: \cref{tab:supp_clip} and the speed--LPIPS sweep).}
\label{fig:supp_quality_curves}
\end{figure}

\begin{figure*}[t]
\centering
\begin{tikzpicture}\begin{axis}[qc,width=0.42\textwidth,title={Klein 4B (4 steps)},ylabel={CLIP-IQA$\uparrow$},ymin=0.43,ymax=0.70,clip=false]
\addplot[only marks,mark=*,black!50] coordinates {(1.00,0.661) (1.84,0.638) (1.77,0.617) (1.30,0.650) (1.79,0.462) (1.65,0.574) (1.52,0.618) (2.08,0.518)};
\addplot[only marks,mark=square*,cvprblue] coordinates {(1.78,0.584) (2.20,0.556)};
\node[font=\tiny,anchor=south,inner sep=1.5pt,text=black!70] at (axis cs:1.00,0.661) {Vanilla};
\node[font=\tiny,anchor=south west,inner sep=1.5pt,text=black!70] at (axis cs:1.84,0.638) {Step skip};
\node[font=\tiny,anchor=east,inner sep=1.5pt,text=black!70] at (axis cs:1.77,0.617) {TeaCache};
\node[font=\tiny,anchor=south,inner sep=1.5pt,text=black!70] at (axis cs:1.30,0.650) {RegionE};
\node[font=\tiny,anchor=north,inner sep=1.5pt,text=black!70] at (axis cs:1.79,0.462) {TaylorSeer};
\node[font=\tiny,anchor=north,inner sep=1.5pt,text=black!70] at (axis cs:1.65,0.574) {RAS};
\node[font=\tiny,anchor=south,inner sep=1.5pt,text=black!70] at (axis cs:1.52,0.618) {SpotEdit};
\node[font=\tiny,anchor=north,inner sep=1.5pt,text=black!70] at (axis cs:2.08,0.518) {Global thr.};
\node[font=\tiny,anchor=west,inner sep=1.5pt,text=cvprblue] at (axis cs:1.78,0.584) {\method-q};
\node[font=\tiny,anchor=south,inner sep=1.5pt,text=cvprblue] at (axis cs:2.20,0.556) {\method-fast};
\end{axis}\end{tikzpicture}\hfill
\begin{tikzpicture}\begin{axis}[qc,width=0.42\textwidth,title={Klein Base (28 steps)},ylabel={CLIP-IQA$\uparrow$},ymin=0.54,ymax=0.76,clip=false]
\addplot[only marks,mark=*,black!50] coordinates {(1.00,0.717) (3.90,0.702) (4.50,0.704) (1.48,0.649) (3.65,0.703) (1.76,0.698) (2.56,0.568)};
\addplot[only marks,mark=square*,cvprblue] coordinates {(4.85,0.679)};
\node[font=\tiny,anchor=south,inner sep=1.5pt,text=black!70] at (axis cs:1.00,0.717) {Vanilla};
\node[font=\tiny,anchor=north west,inner sep=1.5pt,text=black!70] at (axis cs:3.90,0.702) {Step skip};
\node[font=\tiny,anchor=south,inner sep=1.5pt,text=black!70] at (axis cs:4.50,0.704) {TeaCache};
\node[font=\tiny,anchor=north,inner sep=1.5pt,text=black!70] at (axis cs:1.48,0.649) {RegionE};
\node[font=\tiny,anchor=south east,inner sep=1.5pt,text=black!70] at (axis cs:3.65,0.703) {TaylorSeer};
\node[font=\tiny,anchor=south,inner sep=1.5pt,text=black!70] at (axis cs:1.76,0.698) {RAS};
\node[font=\tiny,anchor=south,inner sep=1.5pt,text=black!70] at (axis cs:2.56,0.568) {SpotEdit};
\node[font=\tiny,anchor=north,inner sep=1.5pt,text=cvprblue] at (axis cs:4.85,0.679) {\method};
\end{axis}\end{tikzpicture}
\caption{\textbf{Speed vs.\ absolute image quality} (MICE-Bench, CLIP-IQA, no reference; higher is better). On the many-step editor \method is the fastest method at a quality close to the other accelerators; on the 4-step editor it trades quality for speed, yet stays above one global threshold at similar speed. MICE and MIRAGE ($<$1$\times$) are left out.}
\label{fig:supp_speed_quality}
\end{figure*}
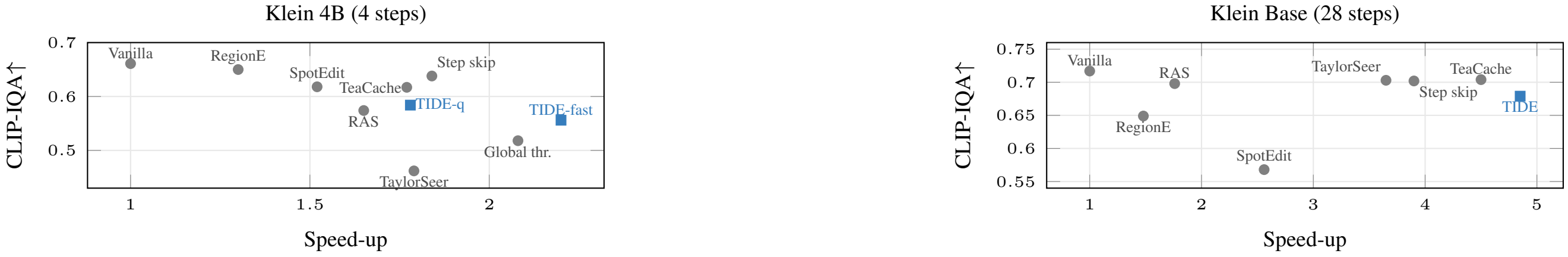

\begin{table}[t]
\centering
\tabsize
\resizebox{\linewidth}{!}{%
\begin{tabular}{@{}lcccc@{}}
\toprule
 & \multicolumn{2}{c}{LPIPS to vanilla$\downarrow$} & \multicolumn{2}{c}{Sharpness vs.\ vanilla} \\ \cmidrule(lr){2-3}\cmidrule(lr){4-5}
Method & in edits & outside & in edits & outside \\
\midrule
\multicolumn{5}{@{}l}{\emph{Klein 4B (4 steps), MICE-Bench}} \\
Step skip (2 steps) & 0.217 & 0.118 & 0.76 & 0.83 \\
TeaCache & 0.275 & 0.144 & 1.11 & 1.12 \\
RegionE & 0.169 & 0.095 & 0.90 & 0.99 \\
One global threshold & 0.377 & 0.192 & 0.38 & 0.66 \\
\method-quality & 0.295 & 0.169 & 0.66 & 0.88 \\
\method-fast & 0.337 & 0.179 & 0.60 & 0.79 \\
\multicolumn{5}{@{}l}{\emph{Klein Base (28 steps), MICE-Bench}} \\
Step skip (7 steps) & 0.205 & 0.135 & 0.87 & 0.92 \\
TeaCache & 0.273 & 0.175 & 1.14 & 1.11 \\
\method & 0.381 & 0.242 & 0.73 & 0.80 \\
\bottomrule
\end{tabular}}
\caption{\textbf{Distance to vanilla and sharpness} (MICE-Bench, 262 images). LPIPS (spatial) to vanilla's output,
inside and outside the edit masks; sharpness: variance of the Laplacian relative to vanilla (median; 1 = as sharp,
$<$1 = smoother). Most of the distance lies inside the edits. \method is smoother than vanilla, the cost of freezing
tokens on an earlier estimate; one global threshold is smoother still.}
\label{tab:supp_inout}
\end{table}

\begin{table}[t]
\centering
\tabsize
\resizebox{\linewidth}{!}{\begin{tabular}{@{}lrrrr@{}}
\toprule
Variant & Spd$\uparrow$ & Done\%$\uparrow$ & Drop$\downarrow$ & LPIPS$\downarrow$ \\
\midrule
\multicolumn{5}{@{}l}{\emph{(a) Regions and scheduling rule (8 steps, $\tau{=}0.6$)}} \\
One global threshold ($\tau{=}0.72$) & 3.46$\times$ & 61.0 & 384 & 0.320 \\
Per-part, ground-truth masks & 3.41$\times$ & 69.0 & 226 & 0.299 \\
Per-part, argmax regions & 3.44$\times$ & 70.3 & 206 & 0.292 \\
\textbf{Per-part, OT regions (ours)} & 3.51$\times$ & 70.9 & 200 & 0.291 \\
\midrule
\multicolumn{5}{@{}l}{\emph{(b) When to decide (4 steps, $\tau{=}1.0$)}} \\
Decide once after step 1 (strict one-pass) & 2.10$\times$ & 72.8 & 146 & 0.226 \\
Background frozen to the source & 2.18$\times$ & 70.4 & 182 & 0.240 \\
\textbf{Re-check every step, freeze-only (ours)} & 2.20$\times$ & 72.1 & 155 & 0.222 \\
\bottomrule
\end{tabular}}
\caption{\textbf{Ablations} (main paper, Sec.~4.5) on MICE-Bench (262 images; VLM judge, Drop relative to vanilla with the same number of steps).}
\label{tab:supp_ablation_copy}
\end{table}

\subsection{Robustness on Qwen-Image-Edit-2511}
\label{sec:supp_qwen}
We repeat the robustness checks of Klein Base on the second model family, Qwen-Image-Edit-2511 (50 steps, true CFG 4.0, 512$^2$,
\method with the Klein Base setting unchanged; one A100, bf16). \Cref{tab:supp_qwen_seeds,tab:supp_qwen_seeds3} repeat
\method over three seeds, \cref{tab:supp_qwen_floor} gives the seed noise floor, \cref{tab:supp_qwen_time} the timing over three repeats,
\cref{tab:supp_qwen_stats} the paired tests, \cref{tab:supp_qwen_judge2} a second judge and VIEScore, and \cref{tab:supp_qwen_pairs} a
paired human check of the edits where \method and TeaCache disagree. Every Drop is counted against vanilla with the same seed and images.

\begin{table*}[t]
\centering
\tabsize
\begin{tabular}{@{}lrrrrrr@{}}
\toprule
Comparison (A vs.\ B) & Drop A & Drop B & A$-$B & 95\% CI & Only A / only B & McNemar $p$ \\
\midrule
\method vs.\ TeaCache, MICE & 59 & 78 & -19 & [-38, -1] & 16/35 & 0.011 \\
\method vs.\ TeaCache, LoMOE & 2 & 5 & -3 & [-7, +1] & 1/4 & 0.37 \\
\method vs.\ TeaCache, MIRA & 4 & 14 & -10 & [-17, -3] & 2/12 & 0.013 \\
\method vs.\ TeaCache, InfoDet & 8 & 9 & -1 & [-5, +3] & 3/4 & 1.00 \\
\method vs.\ TeaCache, Crello & 16 & 17 & -1 & [-8, +6] & 7/8 & 1.00 \\
\method vs.\ TeaCache, GEdit & 3 & 5 & -2 & [-5, +1] & 1/3 & 0.62 \\
\method vs.\ step skip, MICE & 59 & 63 & -4 & [-18, +10] & 20/24 & 0.64 \\
\method vs.\ step skip, LoMOE & 2 & 3 & -1 & [-4, +2] & 1/2 & 1.00 \\
\method vs.\ step skip, MIRA & 4 & 5 & -1 & [-6, +4] & 3/4 & 1.00 \\
\method vs.\ step skip, InfoDet & 8 & 16 & -8 & [-15, -1] & 3/11 & 0.057 \\
\method vs.\ step skip, Crello & 16 & 18 & -2 & [-9, +5] & 7/9 & 0.80 \\
\method vs.\ step skip, GEdit & 3 & 4 & -1 & [-5, +3] & 2/3 & 1.00 \\
\bottomrule
\end{tabular}
\caption{\textbf{Significance of the Drop differences on Qwen-Image-Edit-2511} (50 steps; MICE: 262 images; GEdit: first 100). Paired
bootstrap 95\% CI over images (10,000 resamples) and exact two-sided McNemar test, as in \cref{tab:supp_stats}.}
\label{tab:supp_qwen_stats}
\end{table*}

\paragraph{How \method is mapped onto Qwen-Image-Edit.}
Qwen-Image-Edit-2511 is, like Klein, an in-context flow transformer: its 60 double-stream blocks attend jointly over the text tokens of
its Qwen2.5-VL encoder and one image stream that holds the target latents followed by the reference latents (the reference with its own
zero timestep). \method uses the same pieces as on Klein. \emph{Parts and words}: the instruction is located inside the encoder's chat
template and each part's content words are mapped to their text-stream tokens. \emph{Regions}: the attention of every target query on
these tokens is accumulated over all heads of all 60 blocks during the dense steps, and unbalanced OT with the constants of Klein assigns
the 32$\times$32 target tokens of a 512$^2$ image to parts. \emph{Freezing and engine}: rule (4) with $\tau{=}1.5$, $s{=}4$ dense steps,
a check step every $R{=}8$ steps and a dense last step; sparse steps compute only the active target rows, with the text and reference
keys and values cached after step 0 and refreshed at check steps. \emph{True CFG}: the pipeline calls the transformer for the prompt
and the empty prompt; as on Klein Base, the empty-prompt branch is computed every third sparse step and its difference reused in
between. The reference is encoded at 512$^2$ like the target (the pipeline's default is about one megapixel). A pass with every row
computed reproduces the official forward. No setting was tuned on Qwen.

\begin{table}[t]
\centering
\tabsize
\begin{tabular}{@{}lrrrr@{}}
\toprule
Run & Spd & Done\% & Drop & LPIPS \\
\midrule
MICE, seed 0 (100 images) & 3.48$\times$ & 79.2 & 23 & 0.151 \\
MICE, seed 1 (100 images) & 3.47$\times$ & 80.3 & 21 & 0.148 \\
MICE, seed 2 (100 images) & 3.48$\times$ & 78.2 & 24 & 0.155 \\
LoMOE, seed 0 (64 images) & 3.46$\times$ & 91.7 & 2 & 0.118 \\
LoMOE, seed 1 (64 images) & 3.45$\times$ & 90.4 & 4 & 0.123 \\
LoMOE, seed 2 (64 images) & 3.46$\times$ & 91.0 & 3 & 0.115 \\
\bottomrule
\end{tabular}
\caption{\textbf{Robustness on Qwen-Image-Edit-2511} (\method, Klein Base setting): seeds, with Drop counted against vanilla run with
the same seed. Same format as \cref{tab:supp_brobust}.}
\label{tab:supp_qwen_seeds}
\end{table}

\begin{table}[H]
\centering
\tabsize
\begin{tabular}{@{}lcccc@{}}
\toprule
Qwen, MICE-Bench (100 images) & Seed 0 & Seed 1 & Seed 2 & Mean $\pm$ s.d. \\
\midrule
Step skip & 25 & 27 & 24 & 25.3 $\pm$ 1.5 \\
TeaCache & 31 & 34 & 29 & 31.3 $\pm$ 2.5 \\
\method & 23 & 21 & 24 & 22.7 $\pm$ 1.5 \\
\bottomrule
\end{tabular}
\caption{\textbf{Dropped edits over three seeds on Qwen-Image-Edit-2511}, each against vanilla with the same seed. Same format as
\cref{tab:supp_seeds}.}
\label{tab:supp_qwen_seeds3}
\end{table}

\begin{table}[H]
\centering
\tabsize
\resizebox{\linewidth}{!}{%
\begin{tabular}{@{}lcccccc@{}}
\toprule
 & MICE & LoMOE & MIRA & InfoDet & Crello & GEdit \\
\midrule
Vanilla, another seed (floor) & 28--32 / 624 & 4--6 / 178 & 9 / 412 & 12 / 358 & 19 / 230 & 4 / 90 \\
Step skip & 25 & 3 & 5 & 16 & 18 & 4 \\
TeaCache & 31 & 5 & 14 & 9 & 17 & 5 \\
\method & 23 & 2 & 4 & 8 & 16 & 3 \\
\bottomrule
\end{tabular}}
\caption{\textbf{Seed noise floor on Qwen-Image-Edit-2511} (50 steps; MICE and GEdit: first 100 images). Floor: edits that vanilla with
seed 0 completes and vanilla with another seed loses, out of those completed. Same format as \cref{tab:supp_floor}.}
\label{tab:supp_qwen_floor}
\end{table}

\begin{table}[H]
\centering
\tabsize
\resizebox{\linewidth}{!}{%
\begin{tabular}{@{}lrrrrrr@{}}
\toprule
Method & TF (s) & E2E (s) & Spd TF & Spd E2E & Spd, med.\ of 3 & Peak (GB) \\
\midrule
Vanilla (50 steps) & 26.45 & 28.10 & 1.00$\times$ & 1.00$\times$ & -- & 48.6 \\
Step skip & 7.13 & 8.78 & 3.71$\times$ & 3.20$\times$ & 3.20$\times$ & 48.6 \\
TeaCache & 6.76 & 8.41 & 3.91$\times$ & 3.34$\times$ & 3.34$\times$ & 49.1 \\
\method & 6.42 & 8.07 & 4.12$\times$ & 3.48$\times$ & 3.48$\times$ & 49.7 \\
\bottomrule
\end{tabular}}
\caption{\textbf{Timing on Qwen-Image-Edit-2511} (MICE-Bench, one A100, nothing else on the GPU). Med.\ of 3: median over three repeats
on the first 20 images, vanilla re-timed in each repeat. Same format as \cref{tab:supp_a100}.}
\label{tab:supp_qwen_time}
\end{table}

\begin{table}[H]
\centering
\tabsize
\resizebox{\linewidth}{!}{%
\begin{tabular}{@{}lccccc@{}}
\toprule
Method & Drop, judge & Drop, 2nd judge & VIEScore SC & PQ & O \\
\midrule
Vanilla (50 steps) & -- & -- & 7.92 & 8.35 & 8.13 \\
Step skip & 25 & 16 & 7.84 & 8.28 & 8.06 \\
TeaCache & 31 & 19 & 7.78 & 8.31 & 8.04 \\
\method & 23 & 14 & 7.90 & 8.26 & 8.08 \\
\bottomrule
\end{tabular}}
\caption{\textbf{A second judge and VIEScore on Qwen-Image-Edit-2511} (MICE-Bench, first 100 images). 2nd judge: Idefics3-8B, same
questions and threshold as in \cref{tab:supp_judge2}.}
\label{tab:supp_qwen_judge2}
\end{table}

\begin{table}[H]
\centering
\tabsize
\resizebox{\linewidth}{!}{%
\begin{tabular}{@{}lrcccc@{}}
\toprule
 & & \multicolumn{2}{c}{TeaCache} & \multicolumn{2}{c}{\method} \\
\cmidrule(lr){3-4}\cmidrule(lr){5-6}
Judge says & Edits & Done & Not done & Done & Not done \\
\midrule
Lost only by TeaCache & 35 & 19 & 16 & 30 & 5 \\
Lost only by \method & 16 & 12 & 4 & 11 & 5 \\
Kept by both (control) & 15 & 13 & 2 & 14 & 1 \\
Lost by both (control) & 10 & 2 & 8 & 3 & 7 \\
\midrule
\multicolumn{6}{@{}l}{\emph{Human-verified Drop on the 51 discordant edits}} \\
Lost (rater majority) & & \multicolumn{2}{c}{20} & \multicolumn{2}{c}{10} \\
Lost by this rule only & & \multicolumn{2}{c}{18} & \multicolumn{2}{c}{8} \\
Fleiss' $\kappa$ between raters & & \multicolumn{2}{c}{0.71} & \multicolumn{2}{c}{0.65} \\
\bottomrule
\end{tabular}}
\caption{\textbf{Paired human check on Qwen-Image-Edit-2511} (MICE-Bench, 51 discordant edits and 25 controls; 33 raters, 16--17 votes per edit). Same protocol and format
as \cref{tab:supp_pairs}.}
\label{tab:supp_qwen_pairs}
\end{table}

\section{Additional qualitative results}
\label{sec:qual}
\Cref{fig:supp_global,fig:supp_baselines,fig:supp_regions,fig:supp_failures} show more examples. Each caption states its selection rule.

\newcommand{\qimg}[2]{\includegraphics[width=#1\textwidth]{fig/#2.jpg}}
\newcommand{\qhead}[1]{{\scriptsize #1}}
\newcommand{\qnote}[2]{\multicolumn{#1}{@{}l@{}}{\scriptsize #2}}

\begin{figure*}[t]
\centering
\setlength{\tabcolsep}{0.6pt}
\renewcommand{\arraystretch}{0.4}
\begin{tabular}{@{}ccccc@{\hspace{5pt}}ccccc@{}}
\qhead{Input} & \qhead{Vanilla} & \qhead{Global} & \qhead{\textbf{\method}} & \qhead{Regions} &
\qhead{Input} & \qhead{Vanilla} & \qhead{Global} & \qhead{\textbf{\method}} & \qhead{Regions} \\[2pt]
\qimg{0.0965}{} & \qimg{0.0965}{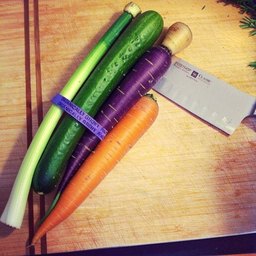} & \qimg{0.0965}{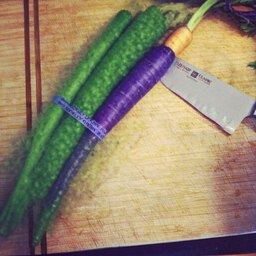} & \qimg{0.0965}{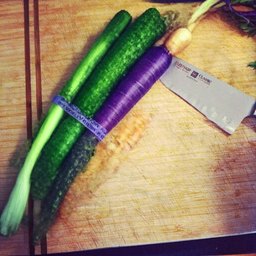} & \qimg{0.0965}{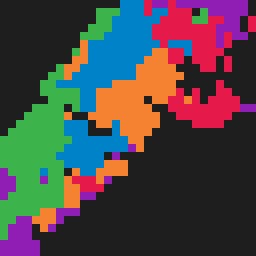} &
\qimg{0.0965}{} & \qimg{0.0965}{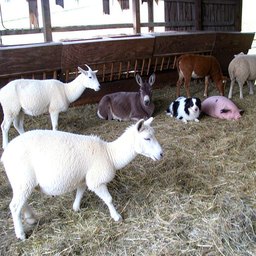} & \qimg{0.0965}{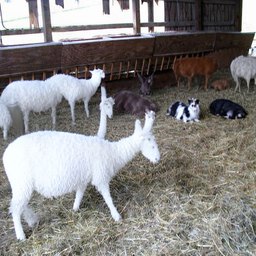} & \qimg{0.0965}{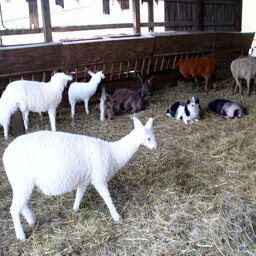} & \qimg{0.0965}{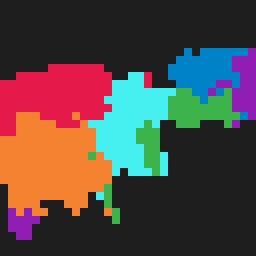} \\
\qnote{5}{\#171 lost: carrot$\to$leek, carrot$\to$parsnip, knife$\to$cleaver} &
\qnote{5}{\#157 lost: sheep$\to$donkey, sheep$\to$sleeping pig} \\[4pt]
\qimg{0.0965}{} & \qimg{0.0965}{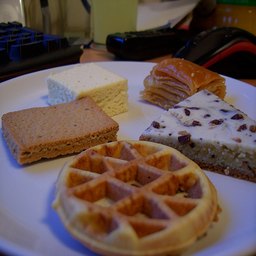} & \qimg{0.0965}{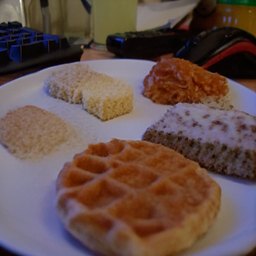} & \qimg{0.0965}{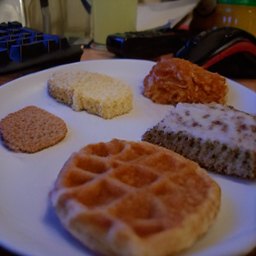} & \qimg{0.0965}{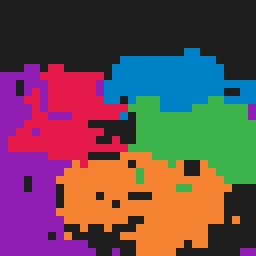} &
\qimg{0.0965}{} & \qimg{0.0965}{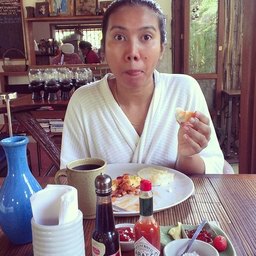} & \qimg{0.0965}{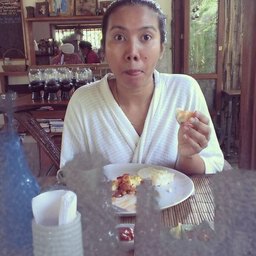} & \qimg{0.0965}{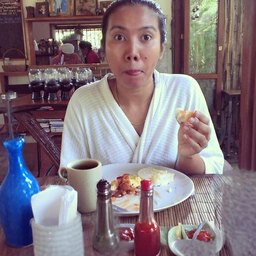} & \qimg{0.0965}{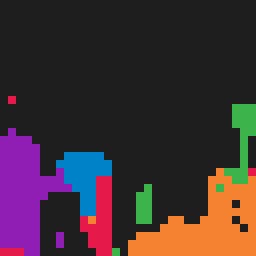} \\
\qnote{5}{\#286 lost: pastry$\to$graham cracker, pastry$\to$baklava} &
\qnote{5}{\#35 lost: glass$\to$coffee mug, napkin holder$\to$blue vase} \\[4pt]
\qimg{0.0965}{} & \qimg{0.0965}{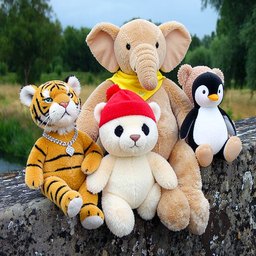} & \qimg{0.0965}{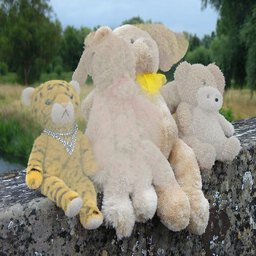} & \qimg{0.0965}{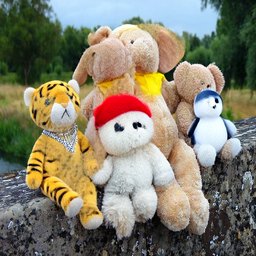} & \qimg{0.0965}{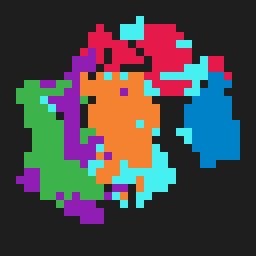} &
\qimg{0.0965}{} & \qimg{0.0965}{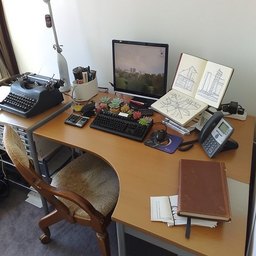} & \qimg{0.0965}{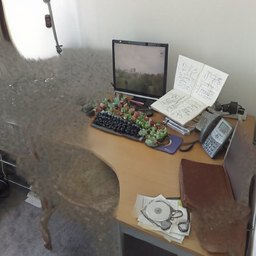} & \qimg{0.0965}{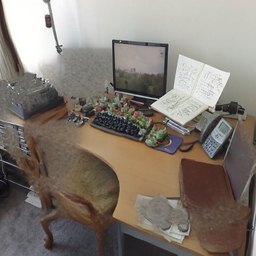} & \qimg{0.0965}{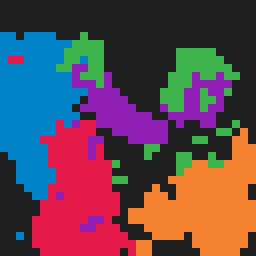} \\
\qnote{5}{\#69 lost: teddy$\to$panda with red hat, teddy$\to$penguin} &
\qnote{5}{\#82 lost: office chair$\to$armchair, laptop$\to$typewriter} \\
\end{tabular}
\caption{\textbf{One global threshold (2.09$\times$) vs.\ per-instruction budgets (\method, 2.20$\times$)} (MICE-Bench,
Klein 4B, 4 steps). Under each image: the edits the global threshold loses and \method keeps. Selection: images with at
most 6 edits where the global threshold loses an edit vanilla made and \method keeps all of them, ordered by the number
of lost edits.}
\label{fig:supp_global}
\end{figure*}

\begin{figure*}[t]
\centering
\setlength{\tabcolsep}{0.6pt}
\renewcommand{\arraystretch}{0.4}
\begin{tabular}{@{}ccccccccc@{}}
\qhead{Input} & \qhead{Vanilla} & \qhead{Step skip} & \qhead{RegionE} & \qhead{TeaCache} & \qhead{RAS} & \qhead{SpotEdit} &
\qhead{TaylorSeer} & \qhead{\textbf{\method-fast}} \\[2pt]
\qimg{0.109}{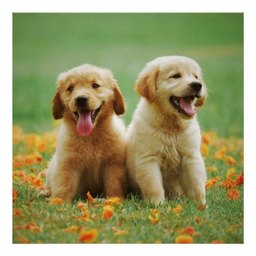} & \qimg{0.109}{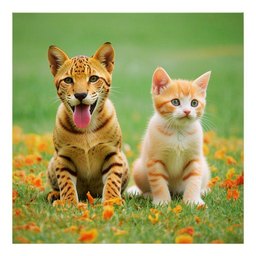} & \qimg{0.109}{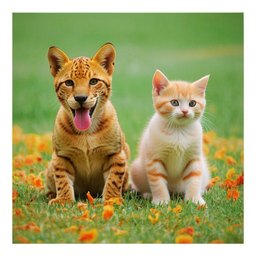} & \qimg{0.109}{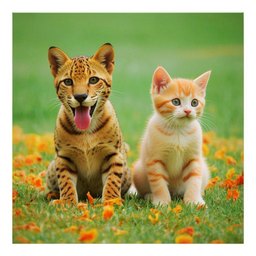} & \qimg{0.109}{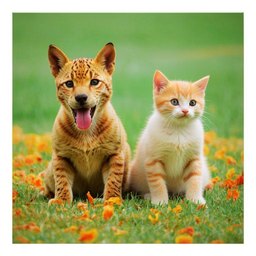} & \qimg{0.109}{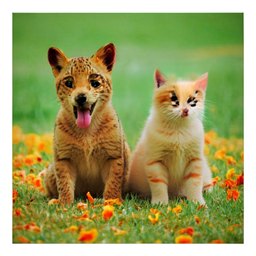} & \qimg{0.109}{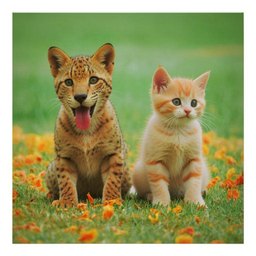} & \qimg{0.109}{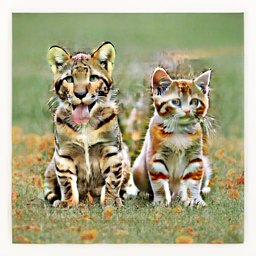} & \qimg{0.109}{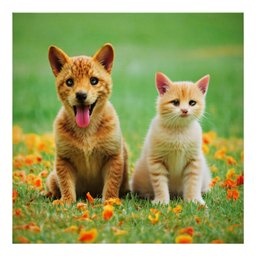} \\
\qnote{9}{LoMOE \#00 (2 edits): right dog$\to$kitten, left dog$\to$baby leopard} \\[4pt]
\qimg{0.109}{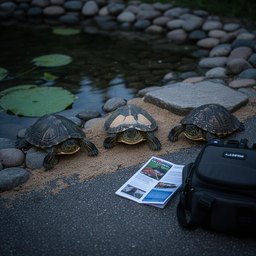} & \qimg{0.109}{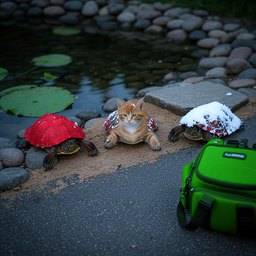} & \qimg{0.109}{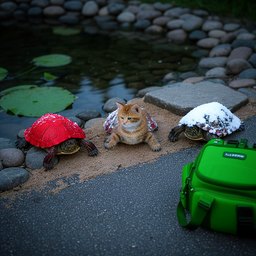} & \qimg{0.109}{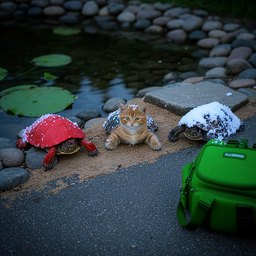} & \qimg{0.109}{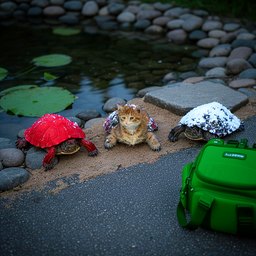} & \qimg{0.109}{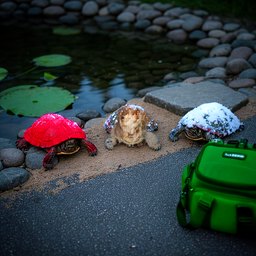} & \qimg{0.109}{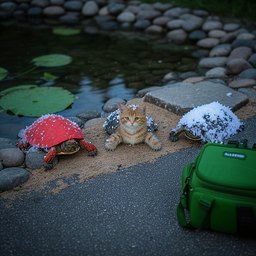} & \qimg{0.109}{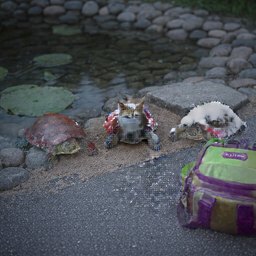} & \qimg{0.109}{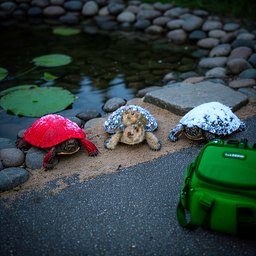} \\
\qnote{9}{MIRA \#0 (5 edits): left shell$\to$red, middle turtle$\to$kitten, snow on right shell, remove newspaper, recolor} \\[4pt]
\qimg{0.109}{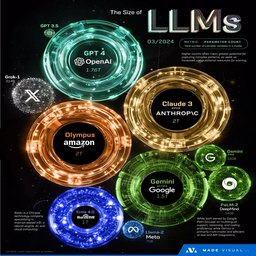} & \qimg{0.109}{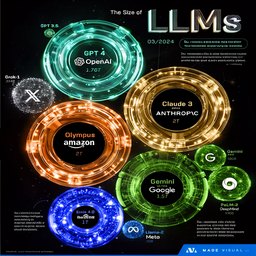} & \qimg{0.109}{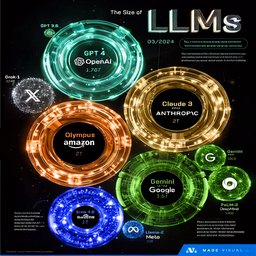} & \qimg{0.109}{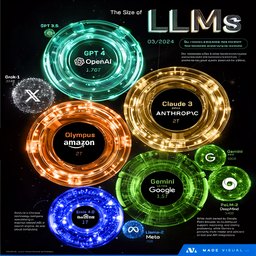} & \qimg{0.109}{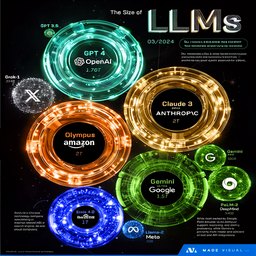} & \qimg{0.109}{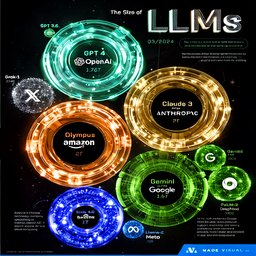} & \qimg{0.109}{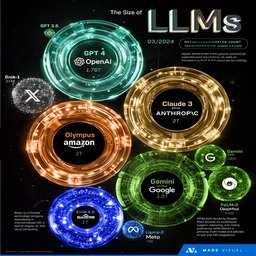} & \qimg{0.109}{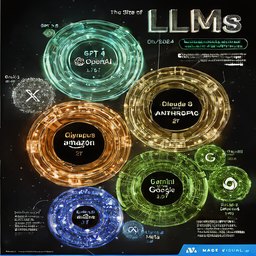} & \qimg{0.109}{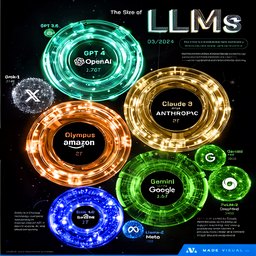} \\
\qnote{9}{InfoDet \#0: English text boxes translated to French} \\[4pt]
\qimg{0.109}{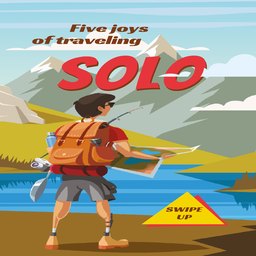} & \qimg{0.109}{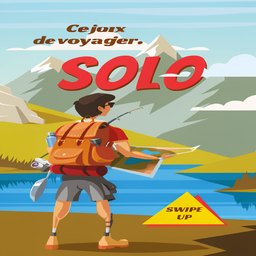} & \qimg{0.109}{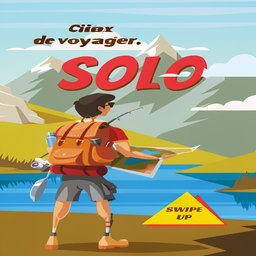} & \qimg{0.109}{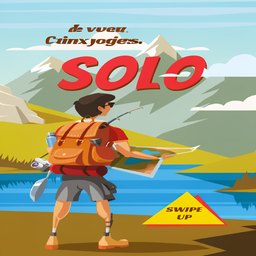} & \qimg{0.109}{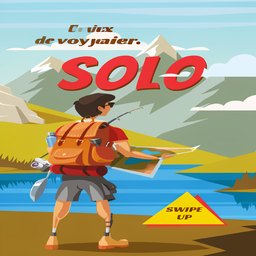} & \qimg{0.109}{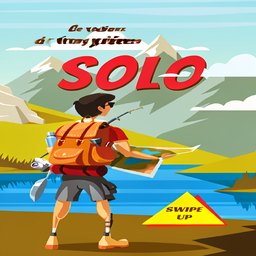} & \qimg{0.109}{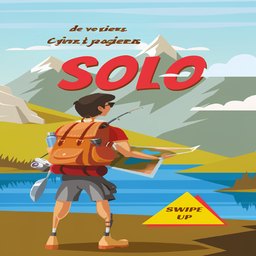} & \qimg{0.109}{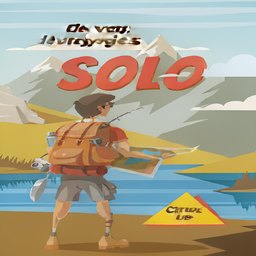} & \qimg{0.109}{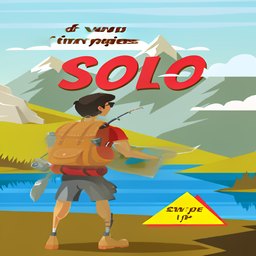} \\
\qnote{9}{Crello \#0: ``of traveling''$\to$``de voyager.'', ``Five joys''$\to$``Cinq joies.''} \\
\end{tabular}
\caption{\textbf{All accelerators on four more benchmarks} (Klein 4B, 4 steps). Selection: the first image of each
benchmark, in dataset order, that every method finished. Speed-ups are in \cref{tab:supp_speed}.}
\label{fig:supp_baselines}
\end{figure*}

\begin{figure*}[t]
\centering
\setlength{\tabcolsep}{0.6pt}
\renewcommand{\arraystretch}{0.4}
\begin{tabular}{@{}ccc@{\hspace{5pt}}ccc@{\hspace{5pt}}ccc@{}}
\qhead{Input} & \qhead{GT masks} & \qhead{\textbf{\method}} &
\qhead{Input} & \qhead{GT masks} & \qhead{\textbf{\method}} &
\qhead{Input} & \qhead{GT masks} & \qhead{\textbf{\method}} \\[2pt]
\qimg{0.107}{} & \qimg{0.107}{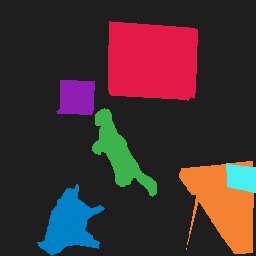} & \qimg{0.107}{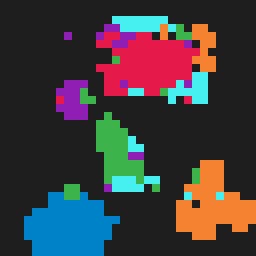} &
\qimg{0.107}{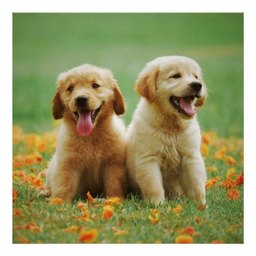} & \qimg{0.107}{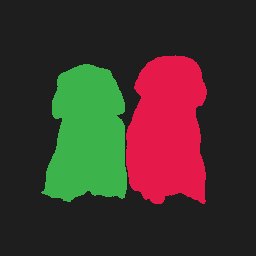} & \qimg{0.107}{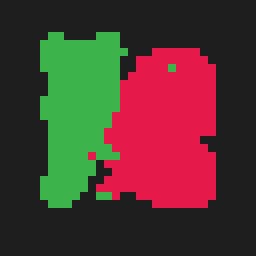} &
\qimg{0.107}{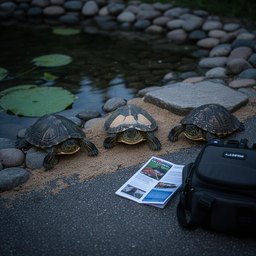} & \qimg{0.107}{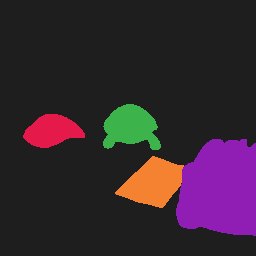} & \qimg{0.107}{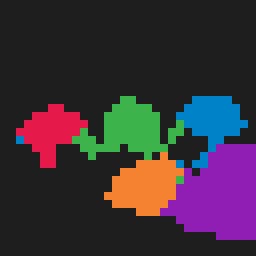} \\
\qimg{0.107}{} & \qimg{0.107}{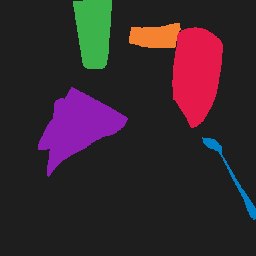} & \qimg{0.107}{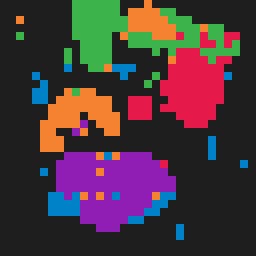} &
\qimg{0.107}{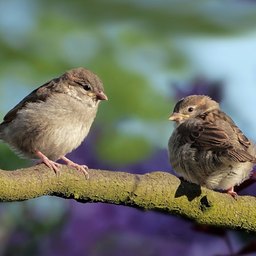} & \qimg{0.107}{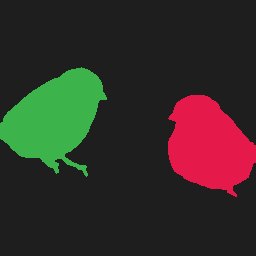} & \qimg{0.107}{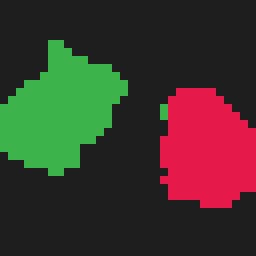} &
\qimg{0.107}{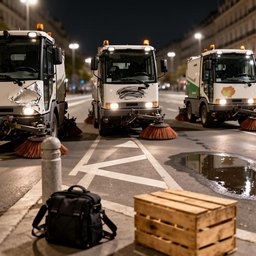} & \qimg{0.107}{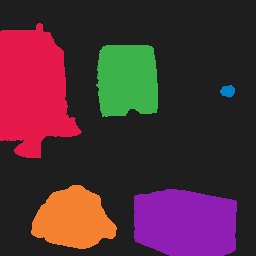} & \qimg{0.107}{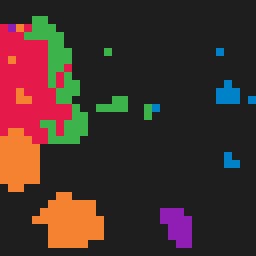} \\[2pt]
\multicolumn{3}{@{}c@{}}{\scriptsize MICE \#24, \#01 (mean IoU 0.33)} &
\multicolumn{3}{c}{\scriptsize LoMOE \#00, \#01 (mean IoU 0.61)} &
\multicolumn{3}{@{}c@{}}{\scriptsize MIRA \#0, \#1 (mean IoU 0.47)} \\
\end{tabular}
\caption{\textbf{Self-grounded regions vs.\ ground-truth masks} (first two images of each benchmark; colors are
arbitrary and do not match between the two maps). \method's regions come from the editor's attention and flow only;
they are coarse but place each instruction part on the right area, which is what the per-part budgets need.}
\label{fig:supp_regions}
\end{figure*}

\begin{figure*}[t]
\centering
\setlength{\tabcolsep}{0.6pt}
\renewcommand{\arraystretch}{0.4}
\begin{tabular}{@{}ccc@{\hspace{5pt}}ccc@{\hspace{10pt}}ccc@{\hspace{5pt}}ccc@{}}
\qhead{Input} & \qhead{Vanilla} & \qhead{\textbf{\method}} & \qhead{Input} & \qhead{Vanilla} & \qhead{\textbf{\method}} &
\qhead{Input} & \qhead{Vanilla} & \qhead{\textbf{\method}} & \qhead{Input} & \qhead{Vanilla} & \qhead{\textbf{\method}} \\[2pt]
\qimg{0.0782}{} & \qimg{0.0782}{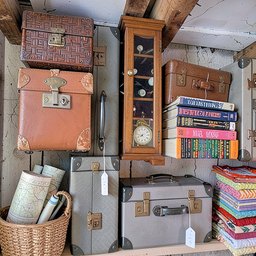} & \qimg{0.0782}{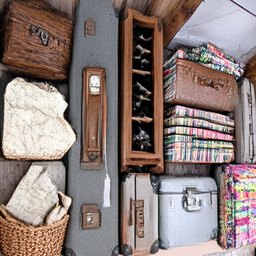} &
\qimg{0.0782}{} & \qimg{0.0782}{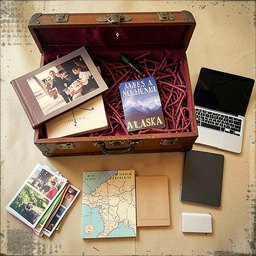} & \qimg{0.0782}{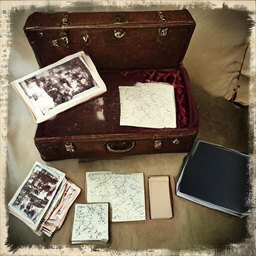} &
\qimg{0.0782}{} & \qimg{0.0782}{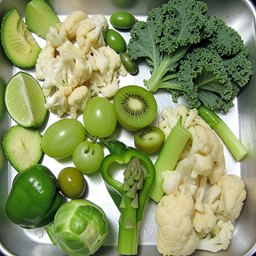} & \qimg{0.0782}{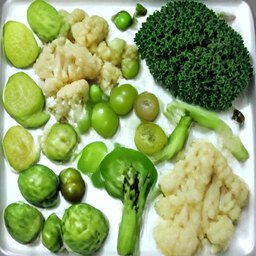} &
\qimg{0.0782}{} & \qimg{0.0782}{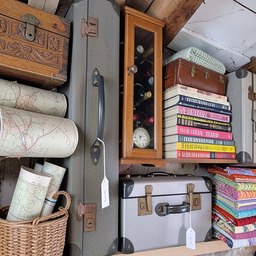} & \qimg{0.0782}{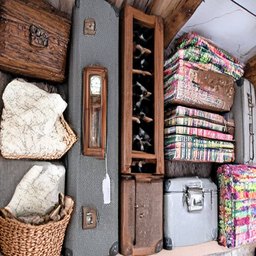} \\[2pt]
\qnote{3}{8 steps, \#38: LPIPS 0.58} & \qnote{3}{8 steps, \#190: LPIPS 0.53} &
\qnote{3}{4 steps, \#67: LPIPS 0.44} & \qnote{3}{4 steps, \#38: LPIPS 0.42} \\
\end{tabular}
\caption{\textbf{Failure cases} (MICE-Bench; 8 steps: $\tau{=}0.6$; 4 steps: \method-fast): for each setting, the two images where \method is furthest from vanilla.
Frozen tokens keep an early, blurry estimate when parts that look settled after the first drift still change later.
At 8 steps this washes out regions; at 4 steps it is much rarer (mean LPIPS per benchmark in \cref{tab:supp_lpips}).}
\label{fig:supp_failures}
\end{figure*}

\section{Limitations, broader impact and reproducibility}
\label{sec:impact}

\paragraph{Limitations.}
\emph{Theory.} The guarantees hold for a linearized error model with per-part budgets: they explain which edits a
threshold drops, not image quality as a whole.
\emph{Regions.} The regions are coarse. The budgets tolerate this (Prop.~C, \cref{tab:supp_regstrat}), but parts with
near-identical words or overlapping objects remain hard.
\emph{Evaluation.} The VLM judge checks whether each edit is present and overcounts losses: raters confirm about a third
of its ``not done'' answers, though they confirm the gap between the rules (\cref{tab:supp_pairs}). It does not penalize
artifacts, which we measure separately (\cref{tab:supp_quality,tab:supp_judge2}).
\emph{Image quality.} On the 4-step editor, \method-fast trades image quality and text accuracy for speed, since frozen
tokens can keep an early, blurry estimate (\cref{fig:supp_failures}); on the many-step editor it is close to the other
accelerators.
\emph{Few-step editors and fast GPUs.} On a 4-step distilled editor, plain step reduction keeps more edits than token
skipping at similar speed, and on an A100 \method's fixed per-step work shrinks its speed-up to that of step skip
(\cref{tab:supp_a100}). On the many-step editor, \method combines with step reduction (\cref{tab:supp_stack}).
\emph{Scope.} We test three editors of two families (FLUX.2 Klein, Qwen-Image-Edit); larger editors are left for
future work.

\paragraph{Broader impact.}
Faster editing lowers the cost of legitimate creative and design work and of evaluating editing systems. It equally
lowers the cost of producing manipulated images. \method adds no editing capability beyond the underlying editor, whose
safeguards and watermarking apply unchanged.

\paragraph{Reproducibility.}
The code for \method, the baselines, the judge and the statistics, with the exact settings of the main runs, will be made
publicly available.